%% file: neurips_2026.tex
\documentclass{article}

\PassOptionsToPackage{numbers,sort&compress}{natbib}
\usepackage[preprint]{neurips_2026}

\usepackage[utf8]{inputenc}
\usepackage[T1]{fontenc}
\usepackage{hyperref}
\usepackage{url}
\usepackage{booktabs}
\usepackage{amsfonts}
\usepackage{amsmath}
\usepackage{nicefrac}
\usepackage{microtype}
\usepackage{xcolor}
\usepackage{graphicx}
\usepackage{multirow}
\usepackage{subcaption}
\usepackage{enumitem}
\usepackage{pifont}
\usepackage{xspace}
\usepackage{booktabs}
\usepackage{siunitx}
\usepackage{textcomp}
\usepackage{listings}
\usepackage{tabularx}
\usepackage{array}
\usepackage{placeins}
\usepackage{wrapfig}
\usepackage{makecell}
\usepackage{longtable}

\hypersetup{hidelinks}

\title{XPlainVerse: A Million-Scale Benchmark for Explainable Deepfake Detection}

\author{%
\textbf{Abhijeet Narang\textsuperscript{1} \quad
Kartik Kuckreja\textsuperscript{2} \quad
Shreya Ghosh\textsuperscript{3}}
\\
\textbf{Muhammad Haris Khan\textsuperscript{2} \quad
Jianfei Cai\textsuperscript{1} \quad
Abhinav Dhall\textsuperscript{1}}
\\
{\normalfont\small
\textsuperscript{1}Monash University \quad
\textsuperscript{2}MBZUAI \quad
\textsuperscript{3}The University of Queensland}
\\
{\normalfont\scriptsize
\texttt{\{abhijeet.narang1,jianfei.cai,abhinav.dhall\}@monash.edu}}
\\
{\normalfont\scriptsize
\texttt{shreya.ghosh@uq.edu.au}
\quad
\texttt{\{kartik.kuckreja,muhammad.haris\}@mbzuai.ac.ae}
}
}

\newcommand{\dataset}{XPlainVerse\xspace}

\newcommand{\xmark}{\ding{55}}

\newcolumntype{Y}{>{\raggedright\arraybackslash}X}
\lstdefinestyle{xvpromptstyle}{%
  basicstyle=\ttfamily\scriptsize,
  breaklines=true,
  breakatwhitespace=true,
  columns=fullflexible,
  keepspaces=true,
  showstringspaces=false,
  upquote=true,
  frame=single,
  framerule=0.35pt,
  framesep=4pt,
  rulecolor=\color{black!45},
  backgroundcolor=\color{black!2},
  aboveskip=0pt,
  belowskip=0pt,
  literate={–}{{\textendash}}1 {’}{{\textquoteright}}1 {“}{{\textquotedblleft}}1 {”}{{\textquotedblright}}1 {•}{{\textbullet}}1 {…}{{\textellipsis}}1 {→}{{$\rightarrow$}}1%
}
\lstnewenvironment{xvpromptbox}[1]{%
  \quote\noindent\textbf{#1}\\[0.1em]\lstset{style=xvpromptstyle}%
}{%
  \endquote%
}

\newcommand{\xvPlaceholderImage}[2][0.16\textwidth]{%
  \IfFileExists{#2}{%
    \includegraphics[width=#1]{#2}%
  }{%
    \fbox{\begin{minipage}[c][0.75in][c]{#1}\centering\scriptsize\textcolor{gray}{Insert image}\end{minipage}}%
  }%
}

\newcolumntype{L}[1]{>{\raggedright\arraybackslash}m{#1}}
\newcolumntype{C}[1]{>{\centering\arraybackslash}m{#1}}

\begin{document}

\maketitle

\input{sections/abstract}

\input{sections/introduction}
\input{sections/related_works}

\input{sections/method}

\input{sections/results_and_analysis}
\input{sections/conclusion}
\bibliographystyle{unsrtnat}
\bibliography{xplainverse_references}

\input{sections/supplementary}


\end{document}

%% file: sections/abstract.tex
\begin{figure}[!htbp]
    \centering
    \includegraphics[width=\linewidth]{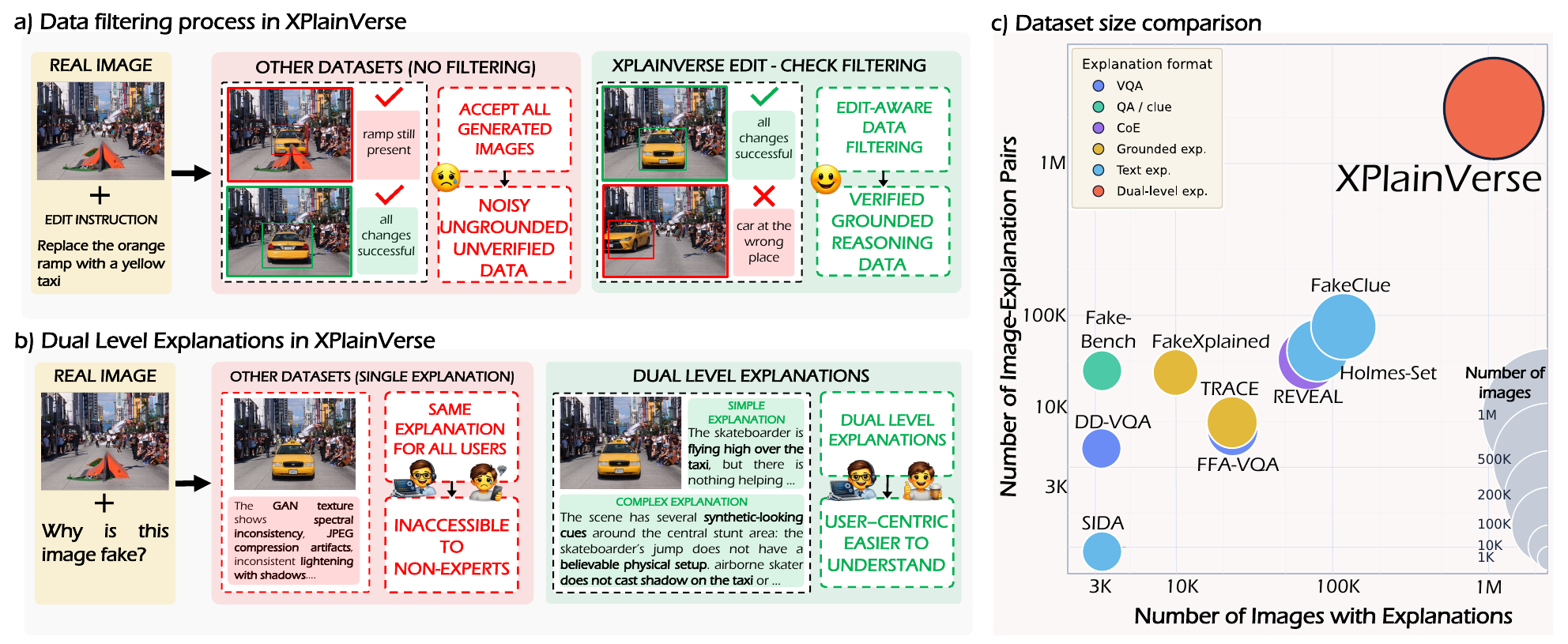}
    \caption{
    The proposed large-scale deepfake explainability dataset \dataset created with (a) Edit-Aware data filtering process containing (b) Dual-level (complex and simple) deepfake explanations in \dataset{}. (c) Size comparison of \dataset with other reasoning-centric deepfake benchmarks.
    }
    \label{fig:teaser}
    \vspace{-0.6em}
\end{figure}

\begin{abstract}
As deepfake detection models increasingly produce natural language explanations, their reasoning often remains weakly grounded in visual artifacts, limiting reliability and user trust. Existing benchmarks mainly evaluate classification accuracy, overlooking whether explanations reflect the actual manipulations. This gap hinders progress toward deployable, explainable deepfake detection systems. To this end, we introduce \textit{XPlainVerse}, a large-scale benchmark designed for joint deepfake detection and human-centered explanation. XPlainVerse comprises \textit{one million} real and manipulated images, pairing authentic images from five established sources with forgeries generated by twelve off-the-shelf image editing and synthesis models. We further propose a multi-stage filtering pipeline,\textit{ Edit-Check}, to verify if manipulations satisfy their intended edits, enabling reliable reasoning supervision at scale. Beyond dataset scale, XPlainVerse provides two complementary explanation styles: technical explanations for expert analysis and simplified explanations optimized for non-technical users. To evaluate explanation quality beyond surface similarity, we propose novel metrics, \textit{EntityScore} and \textit{EvidenceScore}, that measure reasoning fidelity by checking whether explanations correctly identify manipulated entities and visual evidence. Human annotations on 2,000 explanation pairs validate our dataset quality against human judgment. We believe XPlainVerse will establish grounded explanation quality as a measurable dimension of deepfake detection and support scalable research on trustworthy, interpretable models.

\end{abstract}

%% file: sections/introduction.tex
\section{Introduction}
Manipulated images no longer rely on obvious visual artifacts to deceive. Off-the-shelf generative models can seamlessly alter faces, bodies, or scenes while preserving visual realism, making human inspection increasingly unreliable~\cite{hertz2022prompttoprompt,brooks2023instructpix2pix,wu2025qwenimage}. As a result, automated deepfake detectors have become essential across journalism, forensics, and content moderation where detection accuracy alone is no longer sufficient~\cite{rossler2019faceforensicspp,dolhansky2020dfdc,huang2025sida}. When a system flags an image as manipulated, users inevitably ask \textit{why}. Although many detectors now offer natural-language explanations, these often resemble plausible narratives rather than visual evidence, limiting trust and obscuring failure modes. This growing gap between confident explanations and the visual cues that truly support them underscores a key challenge: developing deepfake detection systems that not only decide correctly, but also explain their decisions in grounded, interpretable, and user-meaningful ways. 


Existing benchmarks only partially address this need. Most focus primarily on binary real versus fake classification and provide little or no support for explanation generation~\cite{rossler2019faceforensicspp,dolhansky2020dfdc,zhu2023genimage,wang2023dire}. Even when explanations are considered, current generation models are typically trained on small or synthetic datasets that lack visual diversity and grounding, leading to explanations that are fluent but weakly tied to actual image artifacts. 
A core reason explanation-oriented datasets remain limited in scale~\cite{zhang2024commonsense,huang2024ffaa,li2024fakebench,huang2025sida} is that scaling reasoning annotation automatically introduces substantial label noise, as naively prompting generative models to explain manipulations yields inconsistent and often hallucinated rationales. Our pipeline directly addresses this by combining edit-aware quality filtering with structured generation, producing grounded explanations at scale without sacrificing annotation fidelity. In addition, prevailing evaluation metrics largely emphasize textual similarity or classification accuracy, overlooking whether explanations correctly identify the manipulated entities or supporting visual evidence. Current benchmarks also under-emphasize robustness under realistic post-processing, distribution shift, and adversarial perturbations~\cite{huang2025xtransfer,jia2025foa}.

\begin{table*}[t]
\centering
\caption{\small
Comparison of deepfake explanation-oriented datasets.
QA: Image-conditioned question answering, ME: Model Ensemble, QC: Quality Check, HF: Human Feedback, Human Ann.: Human Annotation
}
\vspace{-2mm}
\label{tab:dataset_comparison}
\scriptsize
\renewcommand{\arraystretch}{1.12}
\setlength{\tabcolsep}{3.0pt}
\begin{tabular*}{\textwidth}{@{\extracolsep{\fill}} l c >{\centering\arraybackslash}p{2.2cm} >{\centering\arraybackslash}p{2.6cm} r r c @{}}
\toprule
\textbf{Dataset} &
\textbf{Year} &
\textbf{Annotation Source} &
\textbf{Annotation Style} &
\textbf{\makecell{Images with\\Explanation}} &
\textbf{\makecell{Total Image-\\Explanation Pairs}} &
\textbf{\makecell{Simple\\Explanation}} \\
\midrule
DD-VQA~\cite{zhang2024commonsense}       & 2024 & Human Ann.         & QA                      & 2.97K         & 14.8K         & \xmark \\
FFA-VQA~\cite{huang2024ffaa}             & 2024 & Model + Human QC   & QA                      & 20.1K         & 20.1K         & \xmark \\
FakeBench~\cite{li2024fakebench}         & 2024 & Model + Human QC   & QA + Text Explanations  & 3K            & 48K           & \xmark \\
REVEAL-Bench~\cite{cao2025reveal}        & 2025 & ME                 & Chain-of-Evidence       & 60K           & 60K           & \xmark \\
FakeXplained~\cite{ji2025fakexplained}   & 2025 & Human Ann.         & Grounded Explanations   & 8.77K         & $\approx$47.5K & \xmark \\
SIDA~\cite{huang2025sida}               & 2025 & Model + Human QC   & Text Explanations       & 3K            & 3K            & \xmark \\
Holmes-Set~\cite{zhou2025aigiholmes}     & 2025 & ME + HF            & Text Explanations       & 69K           & 69K           & \xmark \\
FakeClue~\cite{wen2025spot}             & 2025 & ME                 & Text Explanations       & 100K          & 100K          & \xmark \\
TRACE~\cite{ji2026trace}               & 2026 & ME + QC            & Grounded Explanations   & 20K           & 20K           & \xmark \\
\midrule
\textbf{XPlainVerse} & \textbf{2026} &
\textbf{\makecell{ME + Human Ann.\\ + Human QC}} &
\textbf{Dual-Level Explanations} &
\textbf{1.0M} & \textbf{1.53M} & \textbf{530K} \\
\bottomrule
\end{tabular*}
\end{table*}

User needs further compound the problem. Research in human-centered XAI shows that different users require qualitatively different explanations~\cite{miller2019explanation,ribera2019better,liao2020questioning,ehsan2020human,linder2021level}: a technical analyst wants a precise enumeration of visual artifacts. At the same time, a non-expert moderator needs a single legible reason to act on. LayLens~\cite{narang2025laylens} demonstrates this in the deepfake domain specifically, showing that simplified explanations reduce cognitive load and improve user confidence. Robustness is also largely absent from evaluation: real deployments involve recompressed, blurred, or adversarially perturbed images, yet most benchmarks test on pristine data~\cite{huang2025xtransfer,jia2025foa}.

We introduce \textbf{XPlainVerse}, a \textbf{1M image benchmark} for joint deepfake detection and reasoning. XPlainVerse addresses each of the gaps as discussed above. Our contributions are as follows.

\textbf{Quality-controlled fake image filtering.} We introduce \textit{Edit-Check}, a novel data filtering pipeline to generate reasoning data at scale. Every generated image pair passes through a multi-stage filtering pipeline that verifies the visible changes match the intended edit and discards samples with prominent off-target modifications. Further, to evaluate real-world performance, we add perturbations spanning adversarial attacks and natural corruptions like recompression, denoising, etc.
%

\textbf{Multi-level explanations for diverse users.} \dataset provides complex explanations targeting technical users, which describe multiple concrete visual artifacts, and simple explanations targeting non-expert users, which communicate a single accessible reason. For real images, authenticity explanations highlight natural details that support a genuine label.
%

\textbf{Grounded evaluation metrics for explanation quality.} We introduce novel evaluation metrics, \textit{EntityScore} and \textit{EvidenceScore} that test whether a predicted explanation identifies the correct manipulated entity and the visual evidence supporting it. For simple explanations, separate metrics explicitly measure clarity and conciseness, operationalizing the user-oriented goals.
%

%% file: sections/related_works.tex
\vspace{-2mm}
\section{Related Work}

\noindent\textbf{Detection datasets and explanation benchmarks.}
Deepfake and synthetic-image detection have been shaped by a sequence of progressively larger benchmarks. Early datasets, including FaceForensics++~\cite{rossler2019faceforensicspp}, DFDC~\cite{dolhansky2020dfdc}, and DFFD~\cite{dang2020dffd} focused on facial manipulations, while more recent resources such as MultiFakeVerse~\cite{gupta2025multifakeverse} and SemiTruths~\cite{pal2024semitruths} extend to object- and scene-level edits. These datasets are designed for classification, localization, or robustness evaluation, not for natural-language explanation supervision. A separate line of explanation-oriented benchmarks, including DD-VQA~\cite{zhang2024commonsense}, FFA-VQA~\cite{huang2024ffaa}, FakeBench~\cite{li2024fakebench}, and SIDA~\cite{huang2025sida}, introduces QA-style and textual explanations for manipulated images. However, these datasets are small relative to modern detection benchmarks, and their explanations skew toward technical forensic analysis rather than non-expert communication.

\noindent\textbf{Detection methods and explainability.}
Detection methods have evolved from handcrafted frequency and artifact cues~\cite{liu2021spsl,li2020facexray,li2018eye} to learning-based approaches using convolutional, recurrent, and transformer architectures~\cite{afchar2018mesonet,guera2018deepfake,zhao2021multiattentional,zhang2022patchdiffusion, wang2022m2tr}. Some recent work introduces localization supervision to identify manipulated regions~\cite{dang2020dffd,xu2025fakeshield}, and vision-language approaches such as SIDA~\cite{huang2025sida}, BusterX++~\cite{wen2025busterxpp}, and FakeShield~\cite{xu2025fakeshield} augment classifiers with textual explanations. Yet surface-level fluency does not guarantee that an explanation correctly identifies the manipulated entity or cites evidence present in the image. Further, some recent attempts have tried to evaluate deepfake reasoning through images directly\cite{kuckreja2026pixels}. Human-centered XAI research~\cite{rong2024human,liao2022human} further argues that explanations should be tailored to user expertise, a distinction that existing forensic benchmarks rarely make. Concurrently, advances in diffusion-based image editing~\cite{hertz2022prompttoprompt,brooks2023instructpix2pix,wu2025qwenimage} have made manipulations harder to detect via low-level artifacts alone, reinforcing the need for edit-aware supervision and grounded natural-language explanations. As such, these gaps motivate XPlainVerse.

%% file: sections/method.tex
\begin{figure}[t]
    \centering
    \includegraphics[width=1.0\linewidth]{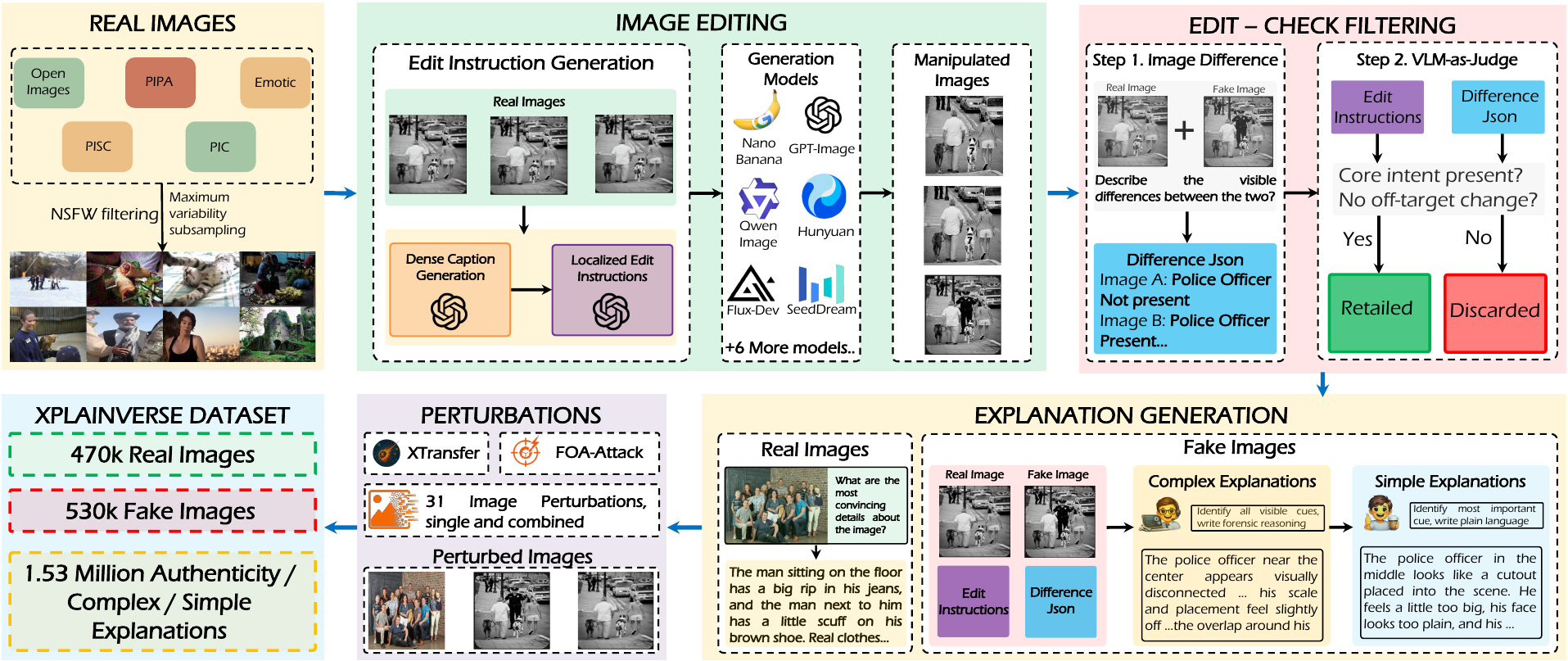}
    \caption{Overview of the XPlainVerse data generation pipeline. For each real image, we generate multiple edit instructions, followed by synthetic image generation. These synthetic images are then passed through our multi-stage filtering pipeline. Finally, we add a range of low-level perturbations and adversarial attacks to both the real and fake sets. }
    \label{fig:overview}
\end{figure}

\section{XPlainVerse Dataset}
\label{sec:dataset}

We start our data generation pipeline by first collecting publicly available real-image datasets spanning diverse real-world scenes, objects, people, activities, and social interactions, including OpenImages~\cite{kuznetsova2020openimages}, EMOTIC~\cite{kosti2020emotic}, PIPA~\cite{oh2015person}, PIC 2.0~\cite{liu2022humancentric}, and PISC~\cite{li2020visualsocial}.
This is followed by a comprehensive data filtering and explanation generation process, as detailed in Figure \ref{fig:overview}. We use an ensemble of multiple non-overlapping LLM / VLM families during each step of data generation to avoid potential biases, with higher sampling for stronger models. Our data generation process is as follows.\footnote{Full prompts for all data filtering and generation steps are provided in Appendix~\ref{app:prompts}.}



\subsection{Data Generation Pipeline}


\noindent \textbf{Step 1: Caption-guided edit-instruction generation.} We begin by leveraging the extensive visual understanding capabilities of foundational vision–language models (VLMs) (particularly GPT-4o-mini~\cite{openai2024gpt4omini}) to obtain concise, information-dense captions to describe primary and secondary subjects, appearance details, spatial layout, lighting, and object interactions without speculation. Further, GPT‑4o‑mini~\cite{openai2024gpt4omini} uses these captions to produce structured edit instructions comprising 2-4 concrete, semantically targeted manipulations. The edits span \textit{person‑centric}, \textit{object‑centric}, \textit{geometry‑centric}, and \textit{scene‑centric} transformations, including attribute and emotion changes, object insertion or removal, copy‑move duplication, local planar warping, and shape deformation. 

\noindent \textbf{Step 2: Candidate manipulated-image construction.} We pair the instructions with the corresponding real images and generate manipulated images using a diverse set of off-the-shelf image‑editing models such as FLUX.2-dev~\cite{blackforestlabs2026flux2dev}, HunyuanImage~\cite{hunyuan2025hunyuanimage}, LongCat-Image-Edit~\cite{meituan2025longcatimage}, Qwen-Image-Edit-2511~\cite{wu2025qwenimage}, GPT-Image-1.5~\cite{openai2025gptimage15}, Seedream-4.5~\cite{bytedance2026seedream45}, Wan 2.6~\cite{alibaba2026wan26}, Nano Banana 2~\cite{google2026nanobanana2}, and Nano Banana Pro~\cite{google2025nanobananapro}. To further broaden generator coverage, we also use manipulated candidates from MultiFakeVerse~\cite{gupta2025multifakeverse}, which includes images generated using three additional editing models: Gemini 2.0 Flash~\cite{kampf2025gemini20flash}, GPT-Image-1~\cite{openai2025gptimage1}, and ICEdit~\cite{zhang2025icedit}. Together, these sources yield an initial pool of approximately 609K manipulated candidates spanning twelve generation models.



\noindent \textbf{Step 3: Edit-Check Pipeline.} To retain only high-quality image-instruction pairs, we apply a two-stage filtering pipeline. First, given a real image
$I_{\text{real}}$ and its generated counterpart $I_{\text{fake}}$, we randomly sample one VLM from a pool of four models (Gemini-3-Flash~\cite{google2025gemini3flash}, GPT-5-mini~\cite{openai2026gpt5mini}, GLM-4.6V~\cite{zai2025glm46v}, and Qwen3.5-27B~\cite{qwen2026qwen35}) to extract a structured JSON object enumerating all verifiable visual differences between the two images. Each identified change is annotated with a salience score ranging from 1 to 5, along with a precise description of the changed region or entity.
In the second stage, a text-only judge sampled from either DeepSeek-V3.2~\cite{deepseek2025v32} or GPT-OSS-120B~\cite{openai2025gptoss} determines whether the extracted differences align with the original edit instruction. Each judge outputs a \texttt{KEEP} or \texttt{DISCARD} decision based on two criteria: whether the core edit intent is present, and whether prominent unintended changes were introduced. Samples where off-target modifications dominate are hard-discarded regardless of partial edit-intent compliance.

The filtering pipeline reduces the initial pool of approximately 609K manipulated candidates to about 265K retained clean manipulated images, corresponding to a discard rate of approximately 56.5\%. This high discard rate highlights the failure rate of generators: without filtering, many explanations would be generated for images whose intended edits are absent or dominated by unintended scene changes, resulting in weakly grounded explanation supervision. Qualitative examples of accepted and discarded candidates are shown in Figure~\ref{fig:teaser}, with additional filtering examples in Appendix~\ref{app:filtering}.


\noindent\textbf{Step 4: Introducing Perturbation.}
To assess robustness under real‑world conditions such as post‑processing, recompression, and adversarial manipulation, each image is paired with a perturbed counterpart. Perturbations are applied cumulatively across splits, i.e., validation includes all training perturbation families, and test includes all validation perturbation families while adding harder corruptions. Training perturbations include an xTransfer-based adversarial variant~\cite{huang2025xtransfer} and standard corruptions spanning noise, blur, compression, and photometric changes. Validation additionally introduces a targeted FOA-style attack~\cite{jia2025foa} and additional perturbations, while the test set further adds harder geometric and frequency-domain corruptions, including affine transformations and ringing halo effect. Across all splits, most perturbed counterparts contain a single perturbation; 10\% combine two perturbations, and another 10\% use heavier mixtures (ranging from 6-17 perturbation). The full perturbation list and parameter ranges are provided in Appendix~\ref{app:perturbations}.

\begin{figure*}[t]
  \centering
  \includegraphics[width=1\textwidth]{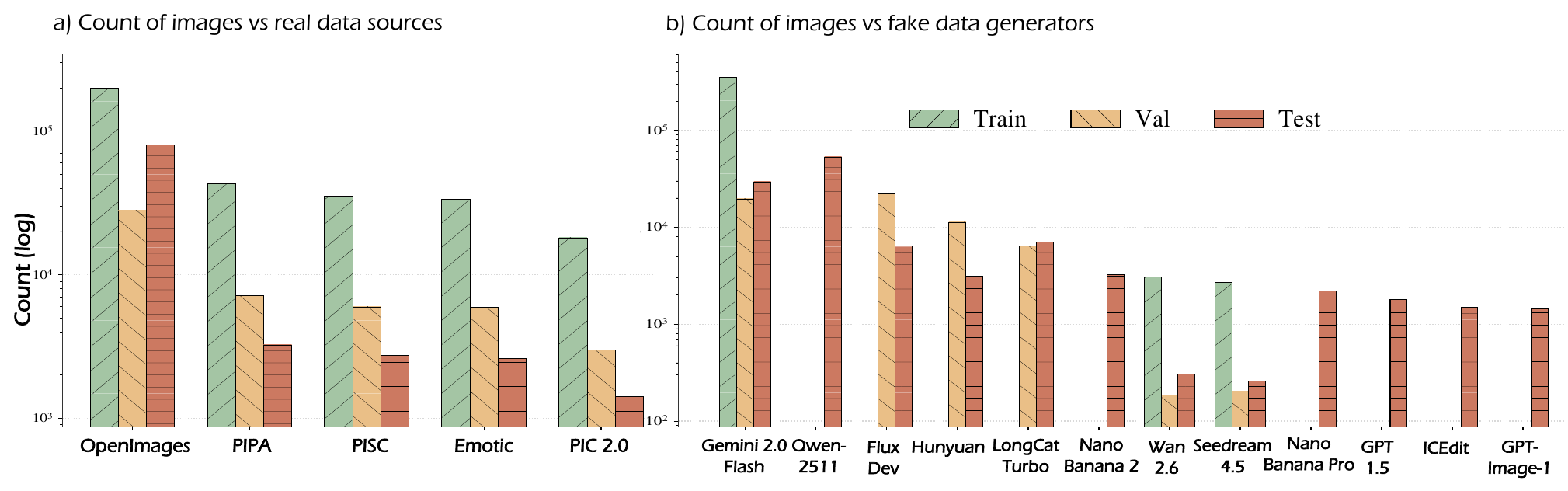}
  \caption{
  a) Distribution for the real data sources subset across the Train, Validation, and Test splits. b) Distribution of the fake data generators set across the Train, Validation, and Test splits.
  }
  \vspace{-5mm}
  \label{fig:split_composition}
\end{figure*}

\noindent \textbf{Step 5: Explanation Generation}
For each image, we generate natural‑language explanations that justify either its manipulation (for fake images) or its authenticity (for real images), with different levels of detail depending on the explanation type.

For manipulated images, we generate two types of explanations. A complex explanation describes the visual evidence of manipulation in enough detail for technical or forensic analysis. To produce it, one model randomly sampled from a pool of four (Gemini-3-Flash \cite{google2025gemini3flash}, GPT-5-mini \cite{openai2026gpt5mini}, GLM-4.6V \cite{zai2025glm46v}, and Qwen3.5-27B~\cite{qwen2026qwen35}) receives the fake image alongside three auxiliary references: the paired real image, the edit-instruction, and the visual-difference produced during filtering. These references help the model identify informative regions and manipulation-relevant cues. The resulting explanations describe cues such as geometric inconsistencies, implausible text or boundaries, inconsistent lighting or reflections, and localized detail failures. A simple explanation is then generated for each fake image by conditioning on the corresponding complex explanation and the same image-level context. Written in plain, layman's terms and focused on a single salient reason why the image appears manipulated, these explanations are designed to be accessible to users of diverse backgrounds.

For real images, each sample is paired with an authenticity explanation that highlights visible properties supporting its genuineness, such as natural imperfections, physically plausible interactions, or realistic material and texture details. These explanations are generated using the same four models above, with the addition of Gemini-2.5-Pro \cite{google2025gemini25pro}.

\subsection{Human Annotations}
To obtain a \textit{gold-standard reference}, we collect human-written explanations from 12 annotators on a subset of 2,000 manipulated images. Annotators are shown each fake image together with its paired real image for context and are instructed to describe all visible cues that indicate manipulation, focusing on evidence observable in the fake image itself and using the real image only as contextual support. To assess consistency, we additionally create a small overlap set of 10 examples annotated independently by 5 annotators, and report inter-annotator agreement using \textit{EntityScore} and \textit{EvidenceScore}, yielding scores of 0.5 and 0.44, respectively. This overlap study provides a consistency check on whether annotators identify similar manipulation cues despite natural variation in phrasing. All annotations follow strict university IRB guidelines (see Appendix \ref{app:ethics}). These human annotations help in validating whether our automatic explanation pipeline produces outputs with similar grounding, completeness, and usefulness (Table \ref{tab:human_eval_pairwise_separate_overall}) to those written by humans.
\subsection{Dataset Quality Analysis and Validation}
\label{sec:quality}

\begin{wraptable}{r}{0.52\linewidth}
\vspace{-1.2em}
\centering
\caption{
Image-quality comparisons.
}
\vspace{-0.8em}
\footnotesize
\setlength{\tabcolsep}{3pt}
\renewcommand{\arraystretch}{0.86}
\begin{tabular}{llcccc}
\toprule
Pair type
& Subset
& PSNR
& SSIM
& LPIPS
& CLIP-I \\
\midrule
Real-fake
& ID discarded  & 17.82 & 0.563 & 0.272 & 0.900 \\
& ID kept       & \textbf{18.80} & \textbf{0.613} & \textbf{0.204} & \textbf{0.920} \\
& OOD discarded & 16.83 & 0.572 & 0.321 & 0.928 \\
& OOD kept      & \textbf{17.87} & \textbf{0.667} & \textbf{0.210} & \textbf{0.930} \\
\midrule
Clean-pert.
& Train & 37.83 & 0.945 & 0.053 & 0.985 \\
& Val   & 36.15 & 0.920 & 0.090 & 0.981 \\
& Test  & 35.58 & 0.890 & 0.098 & 0.976 \\
\bottomrule
\end{tabular}
\vspace{-1.2em}
\label{tab:image_fidelity}
\end{wraptable}
 We analyze XPlainVerse along four complementary dimensions: image fidelity and edit preservation, semantic and explanation diversity, distribution coverage, and human-validated explanation quality. These analyses are intended to verify that the benchmark is not only large but also visually realistic, semantically diverse, and suitable for grounded explanation supervision.

\noindent \textbf{1. Image Fidelity and Explanation Diversity.} To assess curation quality in real-fake pairs, we examine whether retained generated images more faithfully reproduce their paired real images compared to discarded candidates. We compute PSNR, SSIM, LPIPS, and CLIP-I across 20K candidate pairs (10K retained and 10K discarded), balanced across ID and OOD samples (Table \ref{tab:image_fidelity}). Retained pairs consistently exhibit higher PSNR and SSIM, lower LPIPS, and stronger CLIP-I similarity in both settings, supporting that curation successfully filters for structural and semantic fidelity. Because these metrics operate over entire images rather than localized edits, we treat them as a preservation evaluation. For clean-perturbed pairs, we compute the same metrics on 10K pairs per split. The train split remains closest to the clean images, while validation and test splits reflect progressively stronger degradation, consistent with the intended perturbations.

\begin{wraptable}{r}{0.48\linewidth}
\vspace{-1.2em}
\centering
\caption{
Explanation-diversity comparison.
}
\footnotesize
\setlength{\tabcolsep}{3.5pt}
\renewcommand{\arraystretch}{0.88}
\begin{tabular}{lcccc}
\toprule
Dataset / subset
& $n$
& SB$\downarrow$
& D-2$\uparrow$
& Ent.$\uparrow$ \\
\midrule
SIDA                         & 3K & 0.497 & 0.153 & 93.2\% \\
FakeClue                     & 3K & 0.689 & 0.088 & 99.2\% \\
\textsc{XPlainVerse} complex & 3K & \textbf{0.355} & \textbf{0.243} & \textbf{99.6}\% \\
\textsc{XPlainVerse} simple  & 3K & 0.394 & 0.219 & 99.5\% \\
\bottomrule
\end{tabular}
\label{tab:explanation_diversity}
\end{wraptable}

 We also evaluate explanation diversity using 3K subsets from \dataset{}, SIDA, and FakeClue. Self-BLEU measures repetition, Distinct-2 measures lexical diversity, and entity mention rate measures the proportion of explanations that reference at least one concrete region, object, or subject (Table \ref{tab:explanation_diversity}). \dataset{} complex achieves the lowest Self-BLEU and highest Distinct-2, indicating less repetitive wording than SIDA and FakeClue. \dataset{} simple remains more lexically diverse than both external datasets, and over 99\% of explanations in both \dataset{} types mention at least one entity.

\begin{figure*}[htbp]
  \centering
  \includegraphics[width=1.0\textwidth]{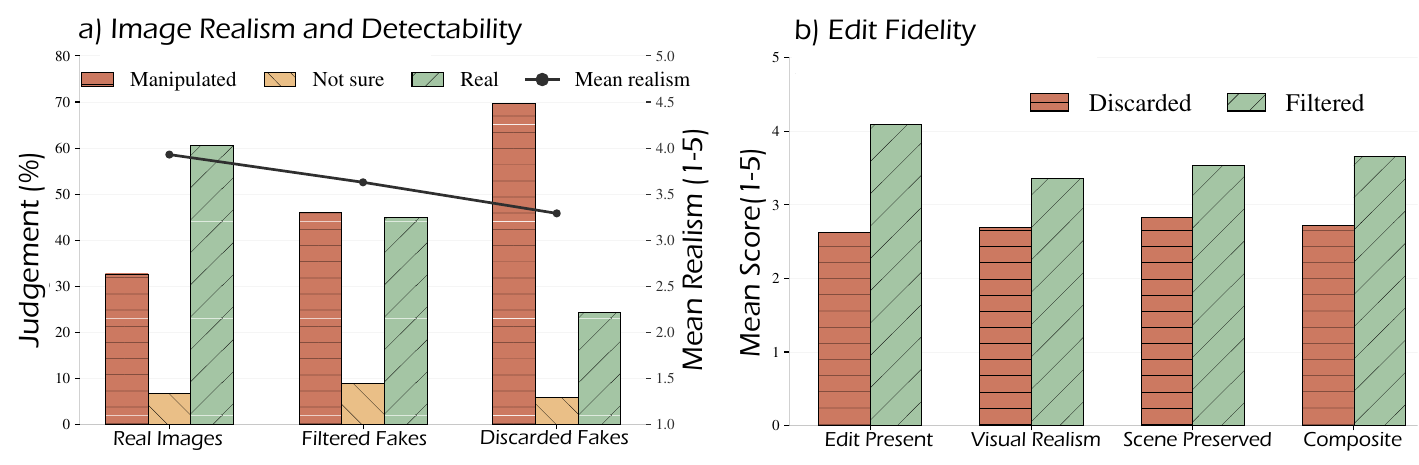}
\caption{
    (a)~Realism comparison between real, filtered, and discarded fakes.
    (b)~Edit-fidelity scores.\\
    Filtered fakes are rated higher in both image realism and edit fidelity than discarded fakes.
   }
    \label{fig:human_evaluation_results1}
\end{figure*}

\begin{figure*}[htbp]
  \centering
  \includegraphics[width=1.0\textwidth]{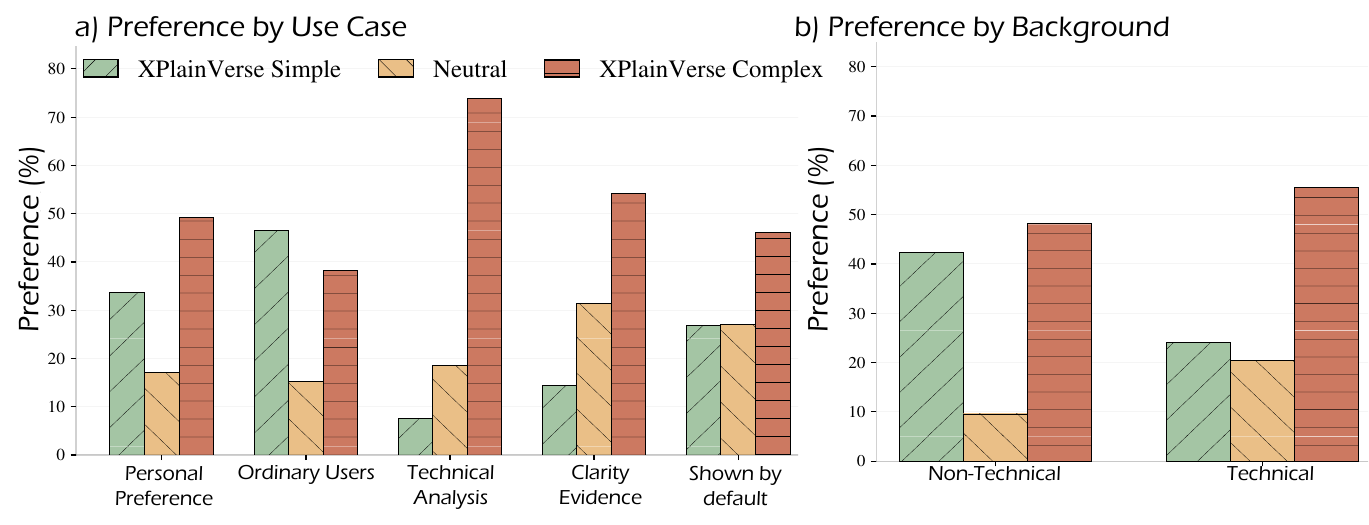}
\caption{
    (a)~Use-case preference comparison between complex and simple explanations.
    (b)~Preference by background comparison between complex and simple explanations.}
    \label{fig:human_evaluation_results2}
\end{figure*}

\noindent \textbf{2. Human Evaluation.}\footnote{Details about the user study and ethics are available in Appendices \ref{app:human_study} and \ref{app:ethics} } We conduct a seven-stage human evaluation covering image realism, edit correctness, explanation quality (grounding, specificity, completeness, unsupported claims), pairwise style comparisons, and readability. Figures~\ref{fig:human_evaluation_results1} and \ref{fig:human_evaluation_results2} summarize the main findings.

\textit{Image and edit quality.} Filtered fake images were difficult to distinguish from real photographs, with annotators labeling them \emph{Real} 44.9\% of the time versus \emph{Manipulated} 46.1\%, closely mirroring the real-image distribution (Figure~\ref{fig:human_evaluation_results1}a). Discarded generations were identified as manipulated 69.7\% of the time, with lower mean realism scores (3.29 vs. 3.63), confirming the filtering pipeline removes conspicuous artifacts and off-target changes. Filtered edits also outperform discarded edits on all four fidelity criteria, with a composite score of 3.66 vs. 2.71 (Figure~\ref{fig:human_evaluation_results1}b). The largest gap appears in the edit presence ($\approx$4.0 vs. 2.6), confirming that the alignment judge filters genuine failures.

\textit{Preference and Explanation quality.} Preferences split clearly by use case (Figure \ref{fig:human_evaluation_results2}(a,b)): Simple is favored for general users (46.5\% vs. 38.3\% for Complex), while Complex dominates for technical users (73.9\% vs. 7.5\%). No single explanation style serves both audiences well, and the gap between the two is large enough that choosing one over the other would leave a significant fraction of users with a suboptimal experience, underscoring the need for user-specific explanations. Further, complex explanations achieved a composite score of 3.79/5 and were preferred over the SIDA-style baseline by 64.8\% of annotators and over human-written explanations by 61.8\% (Table~\ref{tab:human_eval_pairwise_separate_overall}), with the largest margins on specificity (+42.4\%) and completeness (+37.8\%). XPlainVerse Simple scored 3.94/5, with ease-of-understanding at 4.11 and memorability at 4.13 (Appendix Table~\ref{tab:app_human_study_results})
\begin{table*}[t]
\centering
\vspace{-2mm}
\caption{
Comparison of Human preference of SIDA and Human explanations vs. \dataset  (\%).}
\vspace{-2mm}
\scriptsize
\setlength{\tabcolsep}{5.5pt}
\renewcommand{\arraystretch}{1.15}
\begin{tabular}{llcccccccc}
\toprule
& & \multicolumn{4}{c}{\textbf{ \ SIDA}} & \multicolumn{4}{c}{\textbf{\ Human explanation}} \\
\cmidrule(lr){3-6} \cmidrule(lr){7-10}
\textbf{Section} & \textbf{Criterion}
& \textbf{Complex}
& \textbf{Tie}
& \textbf{SIDA}
& \textbf{$\Delta$ vs.\ SIDA}
& \textbf{Complex}
& \textbf{Tie}
& \textbf{Human}
& \textbf{$\Delta$ vs.\ Human} \\
\midrule
\multirow{4}{*}{\textbf{Quality}}
& Grounding
& 64.7 & 12.5 & 22.7 & +42.0
& 60.3 & 7.1 & 32.5 & +27.8 \\
& Specificity
& 60.8 & 14.3 & 25.0 & +35.8
& 67.3 & 7.8 & 24.9 & +42.4 \\
& Completeness
& 58.2 & 16.9 & 24.9 & +33.3
& 64.3 & 9.2 & 26.5 & +37.8 \\
& Unsupported claims
& 46.5 & 30.5 & 23.1 & +23.4
& 47.3 & 25.5 & 27.2 & +20.1 \\
\midrule
\textbf{Overall} & Preference
& 64.8 & 13.1 & 22.2 & +42.6
& 61.8 & 8.2 & 30.0 & +31.8 \\
\bottomrule
\end{tabular}

\label{tab:human_eval_pairwise_separate_overall}
\end{table*}

%% file: sections/results_and_analysis.tex
\section{Evaluation Metrics}
\label{sec:metrics}

A central contribution of XPlainVerse is an \emph{intent-aware} evaluation metric for complex explanations. Standard metrics like BERTScore~\cite{zhang2019bertscore} measure semantic similarity but fail to capture whether a prediction identifies the same manipulated regions or supports them with the same visual evidence. We address this with two tailored metrics: \emph{EntityScore} and \emph{EvidenceScore}.

Given a predicted explanation and a reference explanation, we first use Qwen3.5-4B~\cite{qwen2026qwen35} to extract from each explanation a set of \emph{diagnostic entities} and a set of \emph{evidence claims}. Diagnostic entities correspond to the manipulated regions, subjects, or semantically meaningful image parts mentioned in the predicted and reference explanations. Evidence claims correspond to the concrete visual
cues used to justify why the image appears fake.

We then compute coverage in two directional passes using the same Qwen3.5-4B evaluator~\cite{qwen2026qwen35}. In the first pass, Qwen3.5-4B is given the extracted entities and evidence claims from the \emph{predicted} explanation, together with the full \emph{reference} explanation, and judges whether each predicted item is supported by the reference text. This yields the precision-side scores: \emph{Entity-Precision} and \emph{Evidence-Precision}. In the second pass, it is given the extracted entities and evidence claims from the \emph{reference} explanation, together with the full \emph{predicted} explanation, and judges whether each reference item is supported by the prediction. This yields the recall-side scores: \emph{Entity-Recall} and \emph{Evidence-Recall}. We compute \textit{EntityScore} and \textit{EvidenceScore} as harmonic means of their corresponding precision and recall terms. Full metric definitions are provided in Appendix~\ref{app:metric_details}. Alongside these intent-aware metrics, we report BERTScore-F1 as a complementary measure of overall semantic similarity.


For simple explanations, the goal is to preserve the main reason while remaining accessible to non-experts. We therefore report BERTScore-F1 for semantic fidelity and the SLE simplicity score~\cite{cripwell2023sle} for readability, where higher SLE indicates simpler explanations.

\begin{table*}[!t]
\centering \vspace{-2mm}
\caption{In-distribution and out-of-distribution results on \dataset. Best scores across ID and OOD are shown in \textbf{bold}. Entity: EntityScore, Evidence: EvidenceScore, BERT: BERTScore}
\label{tab:id_ood_fakeonly}
\begingroup
\setlength{\tabcolsep}{3.2pt}
\renewcommand{\arraystretch}{0.92}
\footnotesize
\sisetup{detect-weight=true, detect-inline-weight=math}
\begin{tabular}{@{}>{\centering\arraybackslash}m{0.9cm} >{\centering\arraybackslash}m{1.25cm} l S[table-format=2.2] S[table-format=1.3] S[table-format=1.3] S[table-format=1.3] S[table-format=2.2] S[table-format=1.3] S[table-format=1.3]@{}}
\toprule
& & & \multicolumn{4}{c}{\textbf{Complex explanation}}
    & \multicolumn{3}{c}{\textbf{Simple explanation}} \\
\cmidrule(lr){4-7} \cmidrule(l){8-10}
\textbf{Split}
& \textbf{Setting}
& \textbf{Method}
& {F1 (\%)}
& {BERT}
& {Entity}
& {Evidence}
& {F1 (\%)}
& {BERT}
& {SLE} \\
\midrule
\multirow{12}{=}{\centering\scriptsize\textbf{ID}}& \multirow{2}{=}{\centering\scriptsize\emph{Closed-\\source}}& 
GPT-5-mini~\cite{openai2026gpt5mini}& 58.92 & 0.602 & 0.310 & 0.170 & 54.59 & 0.584 & 0.076 \\
& & Gemini-3-Flash~\cite{google2025gemini3flash}& 75.25 & 0.630 & 0.313 & 0.170 & 75.08 & 0.606 & 0.126 \\
\cmidrule(lr){2-10}
& \multirow{5}{=}{\centering\scriptsize\emph{Open-source\\zero-shot}} & Qwen-3.5-4B~\cite{qwen2026qwen35}& 5.43 & 0.593 & 0.253 & 0.069 & 3.96 & 0.575 & 0.121 \\
& & Qwen3-VL-8B~\cite{bai2025qwen3vl}& 16.14 & 0.605 & 0.266 & 0.082 & 15.26 & 0.594 & 0.139 \\
& & BusterX++~\cite{wen2025busterxpp}& 57.59 & 0.563 & 0.278 & 0.102 & 56.54 & 0.524 & 0.220 \\
& & InternVL-3.5-14B~\cite{wang2025internvl35}& 41.17 & 0.614 & 0.181 & 0.055 & 40.37 & 0.602 & 0.120 \\
& & Qwen-3.5-35B~\cite{qwen2026qwen35}& 17.03 & 0.591 & 0.312 & 0.117 & 11.15 & 0.583 & 0.152 \\
\cmidrule(lr){2-10}
& \multirow{5}{=}{\centering\scriptsize\emph{Fine-\\tuned}} & 
BusterX++~\cite{wen2025busterxpp}& 91.89 & 0.691 & 0.400 & 0.301 & 92.45 & 0.689 & \textbf{0.576} \\
& & Qwen-3.5-4B~\cite{qwen2026qwen35}& 94.13 & 0.702 & \textbf{0.464 }& \textbf{0.371} & 94.39 & 0.700 & 0.568 \\
& & LLaVA-1.6-7B~\cite{liu2023llava,li2024llavanext}& \textbf{96.22} & \textbf{0.703} & 0.453 & 0.361 & \textbf{96.18} & \textbf{0.701} & 0.569 \\
& & Qwen3-VL-8B~\cite{bai2025qwen3vl}& 94.43 & 0.702 & 0.458 & 0.366 & 94.84 & 0.700 & 0.572 \\
& & InternVL-3.5-14B~\cite{wang2025internvl35}& 83.68 & 0.686 & 0.408 & 0.315 & 84.64 & 0.688 & 0.571 \\
\midrule
\multirow{12}{=}{\centering\scriptsize\textbf{OOD}}& \multirow{2}{=}{\centering\scriptsize\emph{Closed-\\source}}& 
GPT-5-mini~\cite{openai2026gpt5mini}& 50.50 & 0.599 & 0.338 & 0.168 & 46.78 & 0.583 & 0.074 \\
& & Gemini-3-Flash~\cite{google2025gemini3flash}& \textbf{71.77} & 0.618 & 0.308 & 0.135 & \textbf{70.05} & 0.601 & 0.120 \\
\cmidrule(lr){2-10}
& \multirow{5}{=}{\centering\scriptsize\emph{Open-source\\zero-shot}} & Qwen-3.5-4B~\cite{qwen2026qwen35}& 7.84 & 0.591 & 0.274 & 0.061 & 10.10 & 0.574 & 0.125 \\
& & Qwen3-VL-8B~\cite{bai2025qwen3vl}& 25.34 & 0.601 & 0.294 & 0.085 & 23.10 & 0.592 & 0.145 \\
& & BusterX++~\cite{wen2025busterxpp}& 49.58 & 0.558 & 0.305 & 0.089 & 48.94 & 0.521 & 0.221 \\
& & InternVL-3.5-14B~\cite{wang2025internvl35}& 38.66 & 0.608 & 0.204 & 0.045 & 36.52 & 0.598 & 0.119 \\
& & Qwen-3.5-35B~\cite{qwen2026qwen35}& 22.31 & 0.592 & 0.345 & 0.108 & 13.57 & 0.582 & 0.150 \\
\cmidrule(lr){2-10}
& \multirow{5}{=}{\centering\scriptsize\emph{Fine-\\tuned}} & BusterX++~\cite{wen2025busterxpp}& 37.14 & 0.639 & 0.312 & 0.166 & 37.87 & 0.653 & 0.580 \\
& & Qwen-3.5-4B~\cite{qwen2026qwen35}& 33.38 & \textbf{0.646} & \textbf{0.357} & \textbf{0.221} & 33.34 & \textbf{0.658} & 0.549 \\
& & LLaVA-1.6-7B~\cite{liu2023llava,li2024llavanext}& 25.14 & 0.642 & 0.333 & 0.200 & 27.41 & 0.655 & 0.551 \\
& & Qwen3-VL-8B~\cite{bai2025qwen3vl}& 32.67 & 0.645 & 0.351 & 0.218 & 32.75 & 0.657 & 0.554 \\
& & InternVL-3.5-14B~\cite{wang2025internvl35}& 42.33 & \textbf{0.646} & 0.330 & 0.205 & 44.68 & 0.656 & \textbf{0.557} \\
\bottomrule
\end{tabular}
\endgroup
\vspace{-3mm}
\end{table*}

\section{Benchmarks and Experiments}
\label{sec:experiments}
\noindent\textbf{Evaluation Protocol.}
We evaluate three tasks: detection, complex explanation generation, and simple explanation generation, across in-domain (ID) and out-of-domain (OOD) manipulated-image subsets. We report detection as \emph{binary F1 score}. Complex explanations are evaluated using EntityScore, EvidenceScore, and BERTScore-F1, while simple explanations are evaluated using BERTScore-F1 and SLE. Further details are available in Appendix \ref{app:baseline_details}.

\noindent\textbf{Baselines.}
\label{sec:main_results}
Table~\ref{tab:id_ood_fakeonly} reveals a consistent pattern across all model families. Fine-tuned models perform strongly on ID data, with LLaVA-1.6-7B reaching 96.22\% F1 and BERTScore of 0.703, but this does not transfer: the same models fall to 25-42\% F1 under OOD conditions, a drop of over 50 points in several cases. Increasing model size does not reliably close this gap, pointing to a fundamental limitation of supervised fine-tuning: models learn generator-specific patterns rather than reasoning grounded in visual evidence.
Zero-shot models tell a complementary story. Gemini-3-Flash stands out as the most robust baseline, achieving 71.77\% OOD F1 without any task-specific training, outperforming all fine-tuned models on that split. Across all settings, BERTScore remains relatively stable under distribution shift (dropping by roughly 0.05--0.06) while EntityScore and EvidenceScore degrade more sharply, revealing that models retain surface fluency but lose the ability to ground explanations in concrete manipulation evidence. Robust explainable detection requires more than fine-tuning on in-distribution data.

\begin{figure*}[t]
  \centering
  \includegraphics[width=\textwidth]{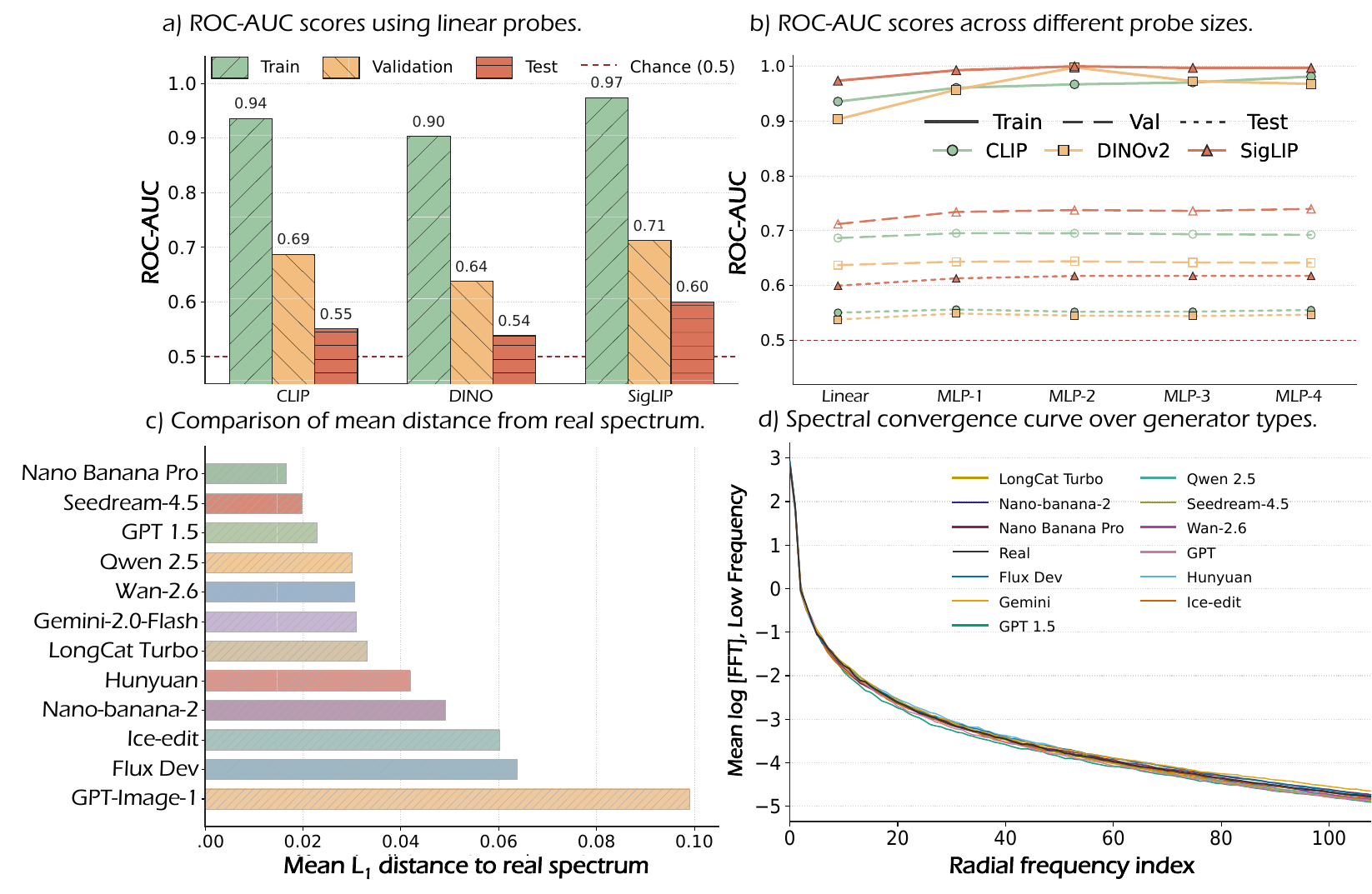}
  \caption{(a)~Linear probes on frozen CLIP, DINOv2, and SigLIP
    embeddings.
    (b)~Increasing probe capacity from linear to MLP-4.
    (c)~Mean $L_1$ distance from the real-image frequency
    spectrum per generator.
    (d)~Radial convergence trends per generator.
  }
  \label{fig:probe_radial}
\end{figure*}

\noindent\textbf{Additional Analysis.} We do in-depth analysis to identify the reason why VLMs do not generalize to the OOD dataset. First, we train linear probes and shallow MLPs on frozen embeddings from CLIP, DINOv2, and SigLIP to understand the train-to-test accuracy drop in Table~\ref{tab:id_ood_fakeonly}. Training AUC is high across all three backbones, yet drops to near chance on the generator-shifted test set. Increasing probe depth leaves test AUC essentially unchanged (Figure~\ref{fig:probe_radial}a,b), indicating that the bottleneck is not capacity but what the model has learned to rely on.
Secondly, as shown in Figure~\ref{fig:probe_radial}(c,d), we do spectral analysis over each successive generation of editing models relative to real photographs, removing the low-level shortcuts that supervised probes depend on. This reveals a core limitation of fine-tuning-based approaches: rather than learning manipulation-relevant semantics, models latch onto generator-specific artifacts that do not transfer. When these artifacts disappear, performance degrades sharply. Detecting modern manipulations requires reasoning about content, such as implausible lighting, inconsistent geometry, or misplaced boundaries. This motivates methods grounded in explicit visual evidence rather than model-specific statistical regularities.

%% file: sections/conclusion.tex
\section{Conclusion}
\label{sec:conclusion}

We introduced XPlainVerse, a million-scale benchmark for explainable deepfake detection, containing 470K real and 530K manipulated images from twelve generators with 1.53M natural-language explanations in two styles: technical explanations for expert analysis and simple explanations for non-technical users. Our edit-aware filtering pipeline removes low-quality generations, and human evaluation confirms that the retained images and explanations are more reliable. In our experiments, zero-shot models perform poorly, while fine-tuned models generalize weakly to held-out generators. EntityScore and EvidenceScore further make grounded explanation quality measurable, revealing failures that surface-level semantic metrics can miss when models produce fluent but visually ungrounded explanations. Spectral analysis reveals that recent generators have largely eliminated the low-level artifacts earlier detectors relied on, making grounded manipulation evidence such as inconsistent geometry, misplaced objects, implausible details, a more stable signal. XPlainVerse aims to support future models that can detect and explain deepfakes in a grounded, user-appropriate way.

\noindent \textbf{Ethics Statement.} All human annotation and evaluation were conducted under institutional IRB approval, with voluntary participation and fair compensation. Real images come from publicly licensed sources, and generated images passed NSFW and semantic filtering before inclusion. The dataset is distributed under a strict research-only license requiring agreement to the XPlainVerse EULA that limits usage to non-commercial, academic research. Full details are in Appendix \ref{app:ethics}.

\noindent \textbf{Limitations and Future Directions.} XPlainVerse has several limitations. It focuses exclusively on still images, leaving temporal artifacts and audio-visual inconsistencies unaddressed. The annotator pool is relatively small and geographically concentrated, which may introduce cultural biases into reference explanations. Reliance on a fixed set of VLMs and commercial APIs also raises reproducibility concerns, and NSFW filtering cannot guarantee the complete removal of distressing content.
Future work should extend the benchmark to video and audio-visual deepfakes, diversify annotator populations, and develop dedicated protocols for evaluating edit-instruction compliance. Models that explicitly localize visual evidence, rather than exploiting generator-specific artifacts, are also encouraged. As generative technology evolves, periodic benchmark updates will be essential to keep XPlainVerse meaningful.



\clearpage

%% file: sections/supplementary.tex
\clearpage
\appendix

\begin{center}
{\LARGE \textbf{Appendix}}\\[0.6em]
{\Large \textbf{XPlainVerse: A Million-Scale Benchmark for Explainable Deepfake Detection}}\\[0.35em]
{\normalsize Supplementary Material}
\end{center}
\vspace{0.75em}

\section{Appendix Roadmap}
\label{app:roadmap}
This appendix collects the details needed to understand and reproduce \dataset{} without interrupting the main paper. We keep the construction details first, then the evaluation protocol, and place the full prompts at the end for readability.

\begin{table}[!htbp]
\centering
\caption{Appendix organization.}
\label{tab:app_roadmap}
\small
\setlength{\tabcolsep}{6pt}
\renewcommand{\arraystretch}{1.18}

\begin{tabularx}{\linewidth}{@{}
    >{\raggedright\arraybackslash}p{0.20\linewidth}
    >{\raggedright\arraybackslash}p{0.18\linewidth}
    >{\raggedright\arraybackslash}X
@{}}
\toprule
\textbf{Section} & \textbf{Name} & \textbf{What it contains} \\
\midrule
Section~\ref{app:ethics} & Ethics & Data sourcing, safety checks, human annotation, data release, dual-use risk, and limitations. \\
Section~\ref{app:dataset} & Dataset & Real/fake counts, generator counts, split accounting, release layout, and ID/OOD notes. \\
Section~\ref{app:edit_instruction_examples} & Edit instructions & Edit types, instruction difficulty scores used for model routing, and representative examples. \\
Section~\ref{app:filtering} & Filtering & Difference extraction, edit-alignment judging, failure categories, and retained and rejected examples. \\
Section~\ref{app:perturbations} & Perturbations & Perturbation list, parameter ranges, mixture protocol, and clean-perturbed examples. \\
Section~\ref{app:explanation_examples} & Explanations & Complex, simple, authenticity, and silver-standard example checks. \\
Section~\ref{app:human_study} & Human study & Annotation protocol and seven-stage evaluation design. \\
Section~\ref{app:metric_details} & Metrics & Entity/Evidence extraction format, coverage rules, and metric configuration. \\
Section~\ref{app:baseline_details} & Baselines & Zero-shot prompting, fine-tuning settings, and output parsing. \\
Section~\ref{app:prompts} & Prompts & Full prompt text for captioning, edit generation, filtering, explanations, and metric judging. \\
\bottomrule
\end{tabularx}

\renewcommand{\arraystretch}{1.0}
\end{table}

\section{Ethics Statement}
\label{app:ethics}

This section documents the ethical practices followed during the construction of XPlainVerse, covering data sourcing, human participation, compensation, IRB oversight, dual-use risk, and release policy. Both the human annotation study and the user evaluation described below were conducted under the same institutional review board approval shared across the affiliated institutions, consistent with the widely used synthetic media datasets, like FaceForensics\cite{rossler2019faceforensicspp} and DeepfakeJudge\cite{kuckreja2026pixels}.

\noindent\textbf{Real-image data.} The real subset draws from five publicly available datasets: OpenImages V7, EMOTIC, PIPA, PIC 2.0, and PISC. Each source is used in accordance with its published research license, and all five are cited in the main paper. No images were scraped from sources without a public data release. We did not collect any new photographs of individuals.

\noindent\textbf{Synthetic and edited image data.} Manipulated images were generated using commercial image-editing APIs, including Gemini, GPT-Image-1 and GPT-Image-1.5, SeedDream, Flux-Dev, HunyuanImage, LongCat, Wan, Nano Banana, and Qwen-Edit, solely for academic research. Additional manipulated candidates came from MultiFakeVerse, which was also produced for research under a compatible license. All candidate images passed through three filtering stages before inclusion: a semantic filter on the generation prompt to remove fantasy and non-photorealistic content, an automated NSFW classifier, and manual spot-checks by the authors. The filtering pipeline described in Section \ref{sec:dataset} and Appendix \ref{app:filtering} removes candidates where the intended edit is absent or the image is dominated by unintended changes; this step also incidentally removes many artifacts that could make images distressing or inappropriate for annotation. No generated image depicting identifiable private individuals was retained.

\noindent\textbf{Human annotation for gold-reference explanations.} Twelve annotators wrote explanations for 2,000 manipulated images. All annotators were adults recruited through academic networks and affiliated with university research groups. Before independent work began, each annotator reviewed written task instructions and worked through example images with the research team to calibrate judgment and establish consistent standards. Annotators were assigned non-overlapping subsets and compensated at rates commensurate with local academic norms. No personally identifying information was collected. Participation was voluntary, and annotators could withdraw at any point without consequence. All materials were screened to exclude NSFW or distressing content before annotation began. Inter-annotator agreement was measured on a shared overlap set of ten images annotated independently by five annotators, and is reported in Appendix \ref{app:metric_details}.

\noindent\textbf{Human evaluation study.} Ten adult participants took part in the seven-stage human evaluation described in Appendix \ref{app:human_study}. Participants were recruited from research groups at the authors' institutions. Before participating, each person received a written statement explaining the study's purpose and procedures, with worked examples of each annotation task, and gave informed consent through a study form. Presentation order was randomized, and examples were balanced across sources and generators. No personally identifying data were recorded, and all responses were anonymized before analysis. Participants were compensated for their time. All study materials were reviewed and cleared for NSFW and distressing content before use.

\noindent\textbf{IRB approval.} All human participation described above, including both annotation phases and the user evaluation, was conducted under institutional review board approval shared across the affiliated institutions. Participation was voluntary throughout. No special populations were recruited, and no participant was placed at foreseeable risk of harm. All annotators and study participants were paid before the study design was finalized, and rates did not depend on the labels or judgments they produced.

\noindent\textbf{Data and model release.} XPlainVerse benchmark assets, including metadata, split files, prompts, generated explanations, human annotations, perturbation metadata, and evaluation protocols, are released under XPlainVerse End User License Agreement (EULA). The EULA permits use only for academic, non-commercial and not-for-profit research. Redistribution, public hosting, mirroring, sublicensing, commercial use, and release of arbitrary dataset images or derived dataset variants are prohibited without prior written approval. Third-party source datasets, images, generated images governed by provider terms, model checkpoints, and software remain subject to their original licenses or provider terms. Where redistribution or relicensing of a third-party asset is not permitted, we release compatible metadata, identifiers, annotations, or XPlainVerse-specific derived components rather than relicensing the original asset. 


\noindent\textbf{Dual-use risk and mitigations.} XPlainVerse has potential dual-use risks, since examples of difficult manipulations and detector failures could inform attempts to create harder-to-detect synthetic media. We mitigate this through controlled access: the dataset is distributed under a strict research-only EULA for non-commercial academic use, with redistribution, public hosting, commercial use, and harmful or deceptive applications prohibited. The real images come from existing public or research datasets, and the manipulated images are produced using off-the-shelf editing systems; we do not release generation model weights, service credentials, or tools intended to create stronger forgeries. We release XPlainVerse as a defensive benchmark for training, evaluating, and explaining deepfake detectors, and believe its primary benefit is to support trustworthy detection research.

\noindent\textbf{Limitations.} XPlainVerse covers still images only; video manipulation is outside its scope. The annotation pool of twelve people and ten evaluators is geographically and institutionally concentrated, which may introduce cultural or stylistic biases in the human-reference explanations. The explanation generation pipeline relies on a small set of commercial VLMs; any systematic tendencies in those models, such as preferences for certain types of cues or certain phrasings, may carry over into the silver-standard explanations. The NSFW filters and manual checks reduce but do not eliminate the possibility that distressing content could appear in the data.

\newcolumntype{P}[1]{>{\raggedright\arraybackslash}p{#1}}

\begingroup
\scriptsize
\setlength{\tabcolsep}{3.5pt}
\renewcommand{\arraystretch}{1.08}

\begin{longtable}{@{}P{0.16\textwidth}P{0.20\textwidth}P{0.30\textwidth}P{0.28\textwidth}@{}}
\caption{Existing assets used in XPlainVerse and their license/terms-of-use handling.}
\label{tab:asset_licenses}\\
\toprule
\textbf{Asset} & \textbf{Use in XPlainVerse} & \textbf{License / terms} & \textbf{Compliance and redistribution} \\
\midrule
\endfirsthead

\toprule
\textbf{Asset} & \textbf{Use in XPlainVerse} & \textbf{License / terms} & \textbf{Compliance and redistribution} \\
\midrule
\endhead

\midrule
\multicolumn{4}{r}{\emph{Continued on next page}}\\
\endfoot

\bottomrule
\endlastfoot

XPlainVerse release
& New benchmark metadata, split files, prompts, generated explanations, human annotations, and evaluation protocol.
& XPlainVerse End User License Agreement (EULA): academic, non-commercial, not-for-profit research-only use; no redistribution, public hosting, sublicensing, or commercial use without prior written approval. Third-party source assets remain under their original licenses or provider terms.
& Distributed through controlled access. Users must accept the EULA, cite the dataset/challenge paper and relevant upstream sources, comply with all upstream terms, and use the data only for approved academic research. Where source/provider terms restrict redistribution, we release compatible metadata, identifiers, annotations, or XPlainVerse-specific derived components rather than restricted assets. \\

OpenImages V7
& Source of real images and annotations.
& Images: Creative Commons Attribution 2.0 (CC BY 2.0); annotations: Creative Commons Attribution 4.0 (CC BY 4.0).
& Original images and annotations remain subject to their source Creative Commons licenses. We retain source attribution/license metadata where available and follow the applicable CC BY attribution requirements. Associated XPlainVerse metadata, annotations, explanations, and splits are distributed under the XPlainVerse EULA, subject to upstream terms. \\

EMOTIC
& Source of real images.
& Non-commercial research and educational use.
& Used only for academic research. Original source is cited. EMOTIC source assets remain subject to their original non-commercial terms. Associated XPlainVerse metadata, annotations, explanations, and splits are distributed under the XPlainVerse EULA where compatible with EMOTIC terms. \\

PIPA
& Source of real images.
& Flickr/source image-level licenses and terms; no single dataset-wide image license identified.
& Used only where access and reuse are compatible with the corresponding source license. Where redistribution is not permitted, released files reference source identifiers or metadata rather than relicensing the original images. Associated XPlainVerse metadata, annotations, explanations, and splits are distributed under the XPlainVerse EULA where compatible with source terms. \\

PIC 2.0
& Source of real images.
& Non-commercial use for academic research, teaching, scientific publication, and personal experimentation.
& Used for research purposes only. Original source is cited. PIC 2.0 source assets remain subject to their original non-commercial terms. Associated XPlainVerse metadata, annotations, explanations, and splits are distributed under the XPlainVerse EULA where compatible with PIC 2.0 terms. \\

PISC
& Source of real images.
& Creative Commons Attribution 4.0 International (CC BY 4.0).
& Original PISC assets remain subject to CC BY 4.0, with attribution to the original authors and source. Associated XPlainVerse metadata, annotations, explanations, and splits are distributed under the XPlainVerse EULA, with required upstream attribution. \\

MultiFakeVerse
& Source of manipulated candidates.
& MultiFakeVerse EULA / research-only access terms; non-commercial academic use only, no redistribution or modification without permission, and compliance with upstream PIC, PISC, PIPA, and EMOTIC terms required.
& MultiFakeVerse assets remain under their original access terms. Associated XPlainVerse metadata, annotations, explanations, and splits are distributed under the XPlainVerse EULA where compatible. \\

Generation/editing APIs
& Candidate manipulated-image generation.
& Official provider API/service terms for GPT-Image, Gemini/Nano Banana, Seedream, Wan, and related closed-source systems.
& Accessed through official APIs. Generated outputs and associated XPlainVerse metadata or annotations are distributed under the XPlainVerse EULA where provider terms permit; otherwise, only compatible metadata, identifiers, or annotations are released. \\

Open-weight generation/editing models
& Candidate manipulated-image generation.
& Model-specific licenses, including FLUX non-commercial license for FLUX.2-dev, Tencent Hunyuan Community License for HunyuanImage, and Apache 2.0 for LongCat-Image and Qwen-Image/Edit where applicable.
& Used according to model-specific licenses. Generated outputs and associated XPlainVerse metadata or annotations are distributed under the XPlainVerse EULA where model terms permit; otherwise, only compatible metadata or annotations are released. \\

Filtering and explanation VLMs
& Captioning, filtering, explanation generation, and metric judging.
& Official API terms for closed-source models; open-weight licenses where applicable, including Apache 2.0 or MIT-style licenses for relevant open models.
& Used for dataset construction, filtering, evaluation, and explanation generation. Closed-source model weights are not redistributed. XPlainVerse textual outputs, annotations, and evaluation records are distributed under the XPlainVerse EULA where model/API terms permit. \\

Baseline model checkpoints
& Zero-shot and fine-tuned baselines.
& Individual model-card licenses and terms for Qwen, LLaVA, InternVL, BusterX++, and related VLM checkpoints.
& Used for evaluation and fine-tuning under their respective model-card terms. The paper reports results and does not redistribute third-party model weights unless explicitly permitted. \\

Software libraries
& Training, inference, feature extraction, and scoring.
& Package-specific open-source licenses, including Apache 2.0 for \texttt{ms-swift}, \texttt{vLLM}, Hugging Face \texttt{transformers}, DINOv2, and SigLIP/Big Vision tooling; MIT for BERTScore and OpenAI CLIP.
& Cited and used according to package licenses. The released code includes attribution and license notices where required. \\

\end{longtable}
\endgroup

\section{Dataset Assembly and Split Details}
\label{app:dataset}

\paragraph{Real-Image Distribution}
The real-image subset of \dataset{} contains 470,000 images from five source corpora. Table~\ref{tab:app_source_inventory} reports the final real data split.

\begin{table}[!htbp]
\centering
\caption{Final real-image source inventory by split.}
\label{tab:app_source_inventory}
\small
\setlength{\tabcolsep}{4pt}
\begin{tabular}{lrrrr}
\toprule
\textbf{Source} & \textbf{Train} & \textbf{Val.} & \textbf{Test} & \textbf{Total images} \\
\midrule
OpenImages & 200,000 & 28,000 & 80,000 & 308,000 \\
PIPA & 43,106 & 7,130 & 3,244 & 53,480 \\
PISC & 35,382 & 5,974 & 2,730 & 44,086 \\
EMOTIC & 33,398 & 5,918 & 2,604 & 41,920 \\
PIC 2.0 & 18,114 & 2,978 & 1,422 & 22,514 \\
\midrule
\textbf{Total} & \textbf{330,000} & \textbf{50,000} & \textbf{90,000} & \textbf{470,000} \\
\bottomrule
\end{tabular}
\end{table}

\paragraph{Fake-Image Distribution}
The fake-image subset contains 530,000 retained images generated across twelve contemporary image-generation and image-editing models. Table~\ref{tab:app_generator_inventory} gives the final retained counts by split. Candidate and discard counts are recomputed from the visual-difference and judge records described in Appendix~\ref{app:filtering}; the released benchmark includes only retained fake samples unless otherwise noted for analysis.

\begin{table}[!htbp]
\centering
\caption{Final retained fake-image inventory by generation model.}
\label{tab:app_generator_inventory}
\small
\setlength{\tabcolsep}{4pt}
\begin{tabular}{lrrrr}
\toprule
\textbf{Generator} & \textbf{Train} & \textbf{Val.} & \textbf{Test} & \textbf{Total} \\
\midrule
Gemini-2.0-flash & 354,256 & 19,612 & 29,438 & 403,306 \\
Flux Dev & -- & -- & 53,222 & 53,222 \\
Qwen 2511 & -- & 22,296 & 6,422 & 28,718 \\
Longcat Turbo & -- & 11,282 & 3,140 & 14,422 \\
Hunyuan & -- & 6,422 & 7,040 & 13,462 \\
ICEdit & 3,050 & 186 & 304 & 3,540 \\
GPT-Image-1 & 2,694 & 202 & 258 & 3,154 \\
Nano Banana 2 & -- & -- & 3,260 & 3,260 \\
Wan 2.6 & -- & -- & 2,190 & 2,190 \\
Seedream 4.5 & -- & -- & 1,790 & 1,790 \\
Nano Banana Pro & -- & -- & 1,500 & 1,500 \\
Gpt 1.5 & -- & -- & 1,436 & 1,436 \\
\midrule
\textbf{Total} & \textbf{360,000} & \textbf{60,000} & \textbf{110,000} & \textbf{530,000} \\
\bottomrule
\end{tabular}
\end{table}

\paragraph{Final Split Composition and ID/OOD Assignment}
Table~\ref{tab:app_split_composition} shows the final data-split across the train, validation, and test sets. Training and validation are used for model development. The test split emphasizes held-out or newer generator families and is used as the primary OOD evaluation.

\begin{table}[!htbp]
\centering
\caption{Final split composition used in the paper.}
\label{tab:app_split_composition}
\small
\setlength{\tabcolsep}{3pt}
\begin{tabular}{lrrrrr}
\toprule
\textbf{Split} & \shortstack{\textbf{Real}\\\textbf{images}} & \shortstack{\textbf{Fake}\\\textbf{images}} & \shortstack{\textbf{Total}\\\textbf{images}} & \shortstack{\textbf{Complex}\\\textbf{explanations}} & \shortstack{\textbf{Simple}\\\textbf{explanations}} \\
\midrule
Train & 330,000 & 360,000 & 690,000 & 360,000 & 360,000 \\
Validation & 50,000 & 60,000 & 110,000 & 60,000 & 60,000 \\
Test & 90,000 & 110,000 & 200,000 & 110,000 & 110,000 \\
\midrule
Total & 470,000 & 530,000 & 1,000,000 & 530,000 & 530,000 \\
\bottomrule
\end{tabular}
\end{table}

\section{Edit Instruction Qualitative Examples}
\label{app:edit_instruction_examples}

\label{app:edit_family_slots}
Table~\ref{tab:app_edit_family_slots} gives representative edit-instruction examples used for data generation in \dataset. 

\begin{table*}[!htbp]
\centering
\caption{Representative edit-instruction examples by edit type.}
\label{tab:app_edit_family_slots}
\scriptsize
\setlength{\tabcolsep}{5pt}
\renewcommand{\arraystretch}{1.12}
\begin{tabularx}{\textwidth}{@{}p{0.18\textwidth}Yp{0.22\textwidth}@{}}
\toprule
\textbf{Edit type} & \textbf{Example instruction} & \textbf{Why it can be hard} \\
\midrule
Object change & \textbf{Scene:} people in helmets and orange life jackets seated in a blue inflatable raft.\newline
\textbf{Instruction:} Replace one of the paddles in the raft with a long, thin surfboard, adjusting the angle to make it look naturally placed.\newline
\textbf{Target:} paddles; blue inflatable raft. & Local replacement with contact, depth, and shadow constraints. \\
Move or resize & \textbf{Scene:} a retro computer desk with a framed photo and a small potted plant.\newline
\textbf{Instruction:} Resize the potted plant to be significantly larger, making it appear as if it is taking up more space on the desk.\newline
\textbf{Target:} potted plant. & The resized object must keep its identity, position, shadows, and occlusion relationships. \\
Copy-move & \textbf{Scene:} musicians in gray jackets playing brass instruments in a crowd.\newline
\textbf{Instruction:} Duplicate the musician on the right and place the duplicate slightly behind the original, adjusting the angle to create a sense of depth.\newline
\textbf{Target:} musician on the right. & Requires identity preservation, shadows, and depth ordering. \\
Shape change & \textbf{Scene:} a white hatchback car parked on pavement with trees in the background.\newline
\textbf{Instruction:} Replace the sporty alloy wheels with oversized, cartoonish wheels that are unnaturally large and round, while keeping the car body consistent.\newline
\textbf{Target:} white hatchback car; sporty alloy wheels. & Shape edit with strict wheel-well, ground-contact, and material constraints. \\
Depth-order change & \textbf{Scene:} a black coupe parked on a paved lot with visible tire and bumper details.\newline
\textbf{Instruction:} Introduce a small, realistic puddle of water on the paved lot near the front tire of the coupe, reflecting the car and surrounding environment.\newline
\textbf{Target:} paved lot. & The new object must sit below the car while preserving believable reflections and contact cues. \\
Text edit & \textbf{Scene:} large white letters spelling \texttt{BONDI BEACH} in front of tall city buildings.\newline
\textbf{Instruction:} Replace the letters spelling \texttt{BONDI BEACH} with \texttt{CITY PARK} in the same large white font, ensuring the new text fits seamlessly.\newline
\textbf{Target:} large white letters. & Text must remain legible, scaled correctly, and physically integrated with the structure. \\
Planar warp & \textbf{Scene:} the Gibeau Orange Julep sign mounted below a large orange icon.\newline
\textbf{Instruction:} Apply a local planar perspective warp to the black and white sign to make it appear slightly tilted, revealing a portion of the brick wall behind it.\newline
\textbf{Target:} black and white sign; brick building. & Planar geometry must remain locally plausible after the warp. \\
\bottomrule
\end{tabularx}
\renewcommand{\arraystretch}{1.0}
\end{table*}

\section{Quality Filtering}
\label{app:filtering}
This section documents how the multi-stage filtering pipeline behaves across generators, edit types, and failure modes. It also includes qualitative examples of retained and discarded candidates.

\paragraph{Filtering Decision Rules}
Filtering is applied to each generated candidate using two structured records: a visual-difference record and an alignment-judge record. The visual-difference extractor compares the paired original and edited images and lists only visible, localized before/after differences. The alignment judge then compares the requested edit instruction against the difference record and returns a final score, a match score, an unrelated-change penalty, and a discrete verdict.

A candidate is retained when the generated image visibly realizes the requested edit and the difference JSON provides localized evidence that can support an explanation. Candidates are rejected when the target edit is missing, dominated by off-target scene rewrites, affected by severe low-level artifacts, or too ambiguous for reliable explanation supervision.

\begin{table}[!htbp]
\centering
\caption{Failure categories for discarded fake candidates.}
\label{tab:app_filter_failure_categories}
\small
\setlength{\tabcolsep}{4pt}
\begin{tabularx}{\linewidth}{p{0.28\linewidth}Y}
\toprule
\textbf{Failure mode} & \textbf{Discard criterion} \\
\midrule
Missing edit & The requested target edit is not visibly present, or the difference extractor cannot identify any concrete on-target before/after evidence. \\
Partial edit & The output contains a weak fragment of the intended edit but misses the core semantic change. \\
Prominent off-target change & An unrelated addition, removal, replacement, identity shift, or text/content change is one of the dominant visible differences. \\
Identity / attribute drift & A person, object, or key attribute changes beyond the requested local manipulation. \\
Background or layout rewrite & Camera framing, background, or scene structure changes enough that the sample no longer isolates the requested manipulation. \\
Low-level artifact & Severe warping, boundary errors, text corruption, blur, or rendering artifacts prevent reliable explanation supervision. \\
Unsafe / ambiguous sample & The target region is too small, occluded, ambiguous, or unsuitable for release and evaluation. \\
\bottomrule
\end{tabularx}
\end{table}

\paragraph{Failure Categories for Discarded Candidates}
Table~\ref{tab:app_filter_failure_categories} lists the main reasons a generated candidate can be rejected during filtering.

\paragraph{Filtering records.}
Each generated candidate is audited with the same two records: a visual-difference JSON comparing the paired original and edited images, and an edit-alignment judge JSON comparing the requested edit against the extracted differences. The judge returns \texttt{KEEP}, \texttt{BORDERLINE}, or \texttt{DISCARD}, along with match and off-target-change scores. Final split counts are computed after this unified filtering pass and split assignment.

\begin{figure*}[!htbp]
\centering
\scriptsize
\setlength{\fboxsep}{5pt}
\fbox{\begin{minipage}{0.95\textwidth}
\begin{center}
\begin{minipage}[t]{0.46\textwidth}\centering
\xvPlaceholderImage[0.95\linewidth]{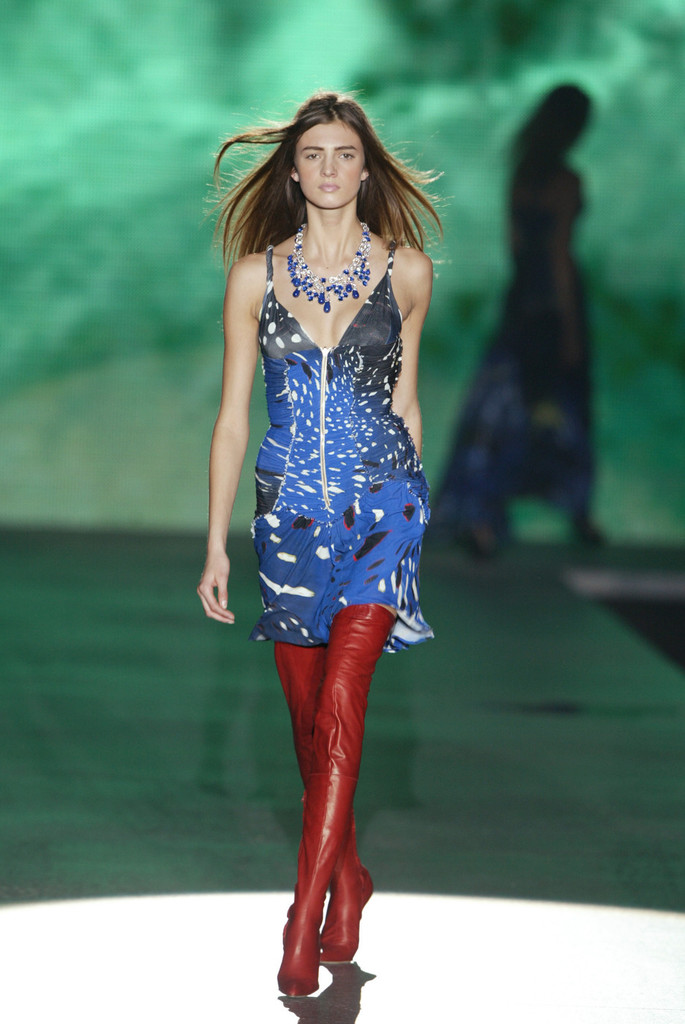}\\[-0.2em] Original
\end{minipage}\hfill
\begin{minipage}[t]{0.46\textwidth}\centering
\xvPlaceholderImage[0.95\linewidth]{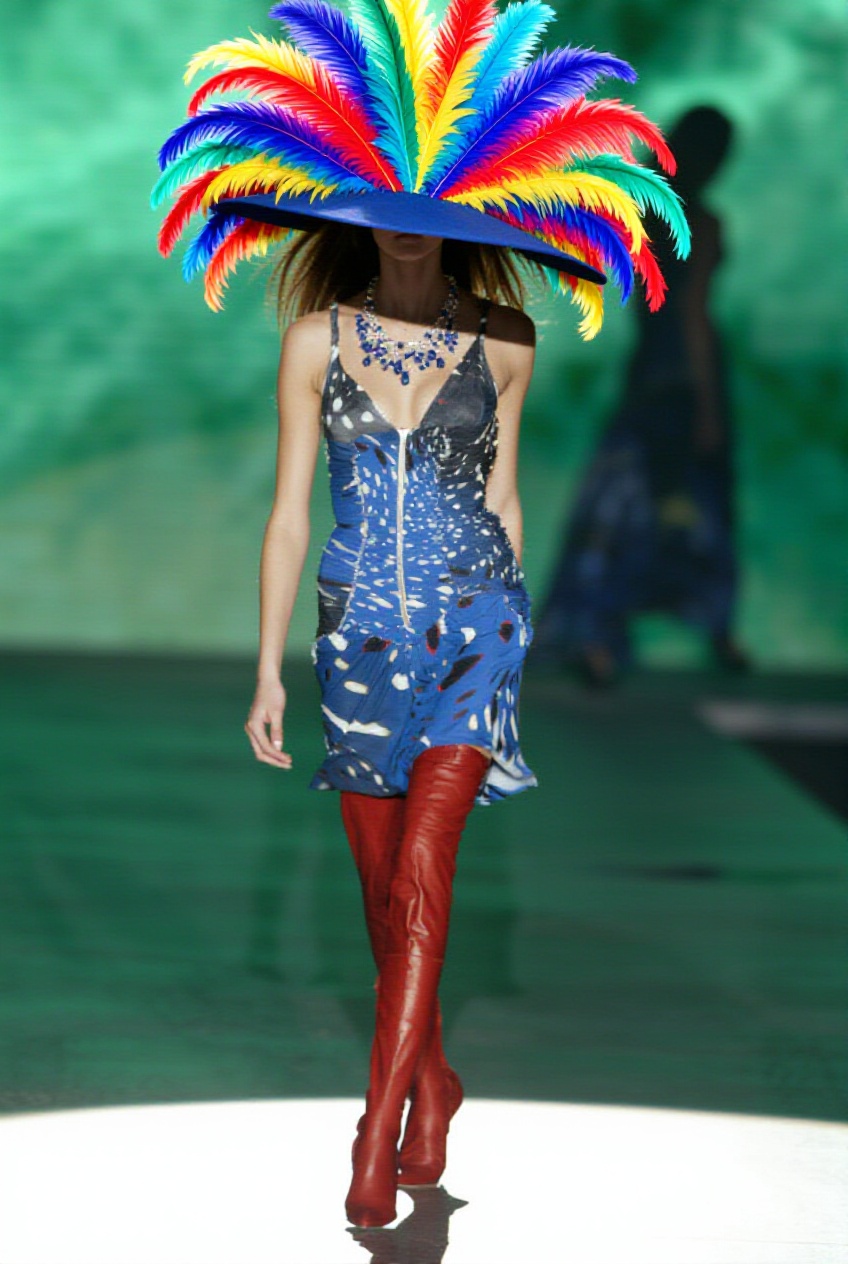}\\[-0.2em] Edited
\end{minipage}
\end{center}
\textbf{Summary:} A large, colorful feathered hat is added on the model's head.\par
\textbf{Visual Difference:} \texttt{Hat}, salience 5: A: No hat on the model's head; B: A large, colorful feathered hat is present on the model's head.
\end{minipage}}

\vspace{0.8em}
\fbox{\begin{minipage}{0.95\textwidth}
\begin{center}
\begin{minipage}[t]{0.46\textwidth}\centering
\xvPlaceholderImage[0.95\linewidth]{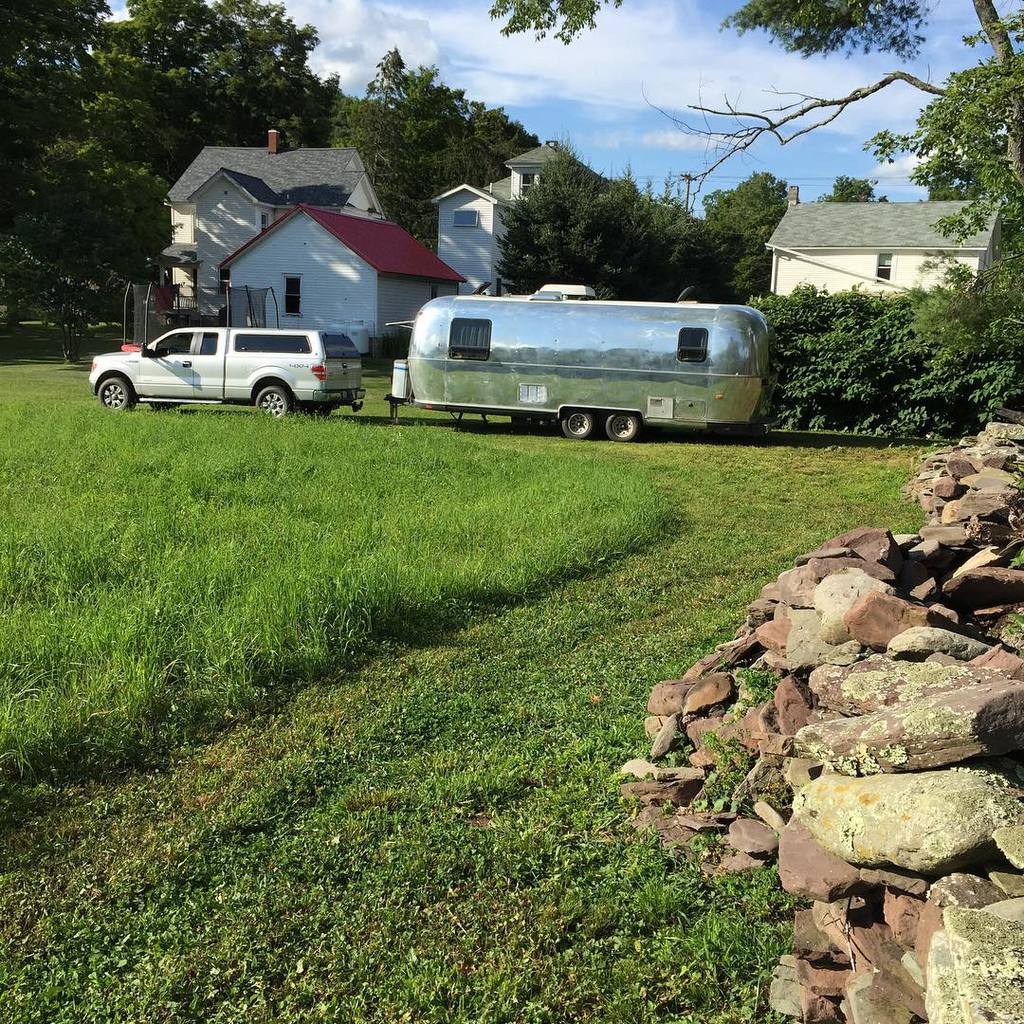}\\[-0.2em] Original
\end{minipage}\hfill
\begin{minipage}[t]{0.46\textwidth}\centering
\xvPlaceholderImage[0.95\linewidth]{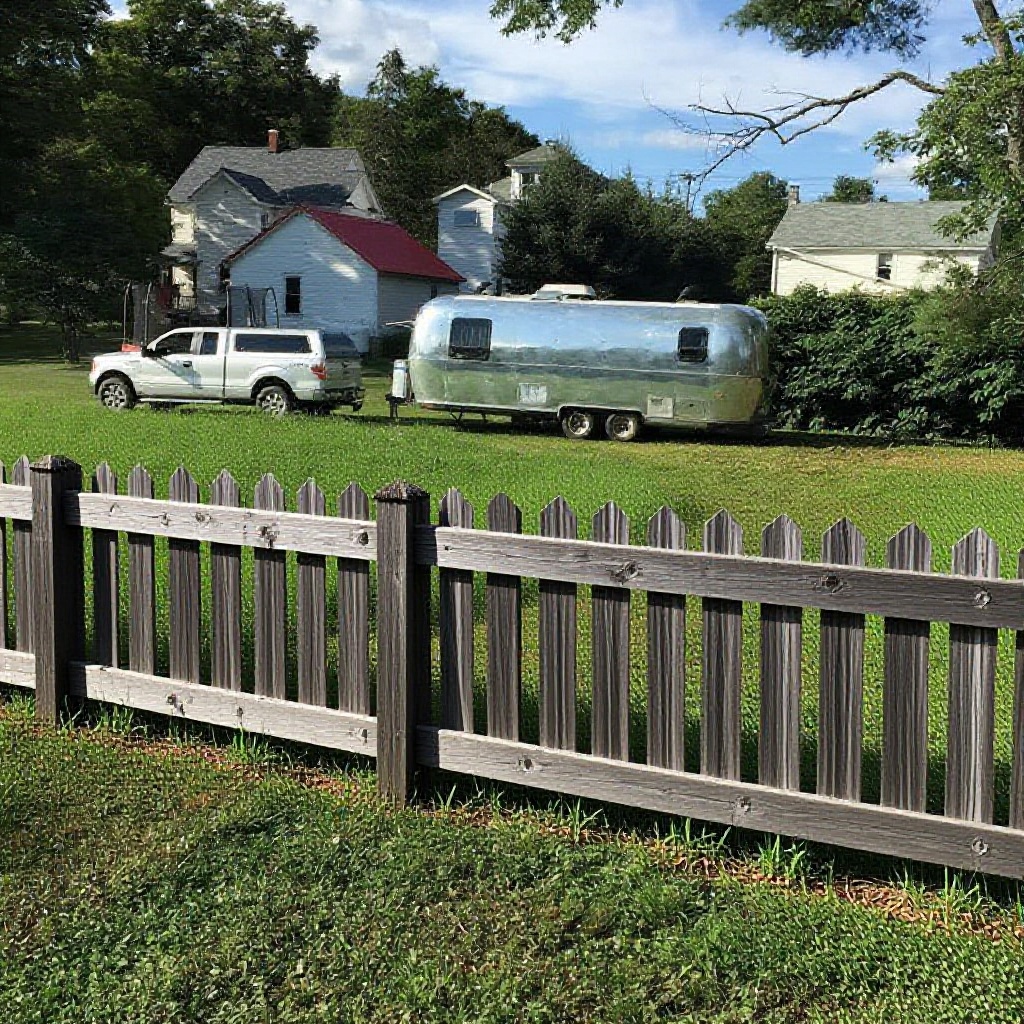}\\[-0.2em] Edited
\end{minipage}
\end{center}
\textbf{Summary:} The primary difference is the replacement of a stone wall in the foreground with a wooden picket fence, along with changes to the grass texture and height in the immediate foreground.\par
\textbf{Visual Difference:} \texttt{Foreground barrier}, salience 5: A: A stacked stone wall is visible on the right side of the foreground; B: A wooden picket fence spans across the entire foreground. 

\texttt{Grass in foreground}, salience 4: A: The grass is tall and unevenly cut, with a distinct mowed line separating long and short sections; B: The grass is shorter, more uniform in height, and lacks the distinct mowed line seen in Image A.
\end{minipage}}
\caption{Visual-difference extraction examples. The examples show that the pipeline records localized before/after evidence rather than relying on generic descriptions of manipulation.}
\label{fig:app_visual_difference_examples}
\end{figure*}

\paragraph{Visual-Difference Examples}
\label{app:visual_difference_examples}
Figure~\ref{fig:app_visual_difference_examples} shows two visual-difference records tied to visible evidence.

\begin{figure*}[!htbp]
\centering
\scriptsize
\setlength{\fboxsep}{5pt}
\fbox{\begin{minipage}{0.95\textwidth}
\begin{center}
\begin{minipage}[t]{0.46\textwidth}\centering
\includegraphics[width=0.95\linewidth,height=4.2cm,keepaspectratio=false]{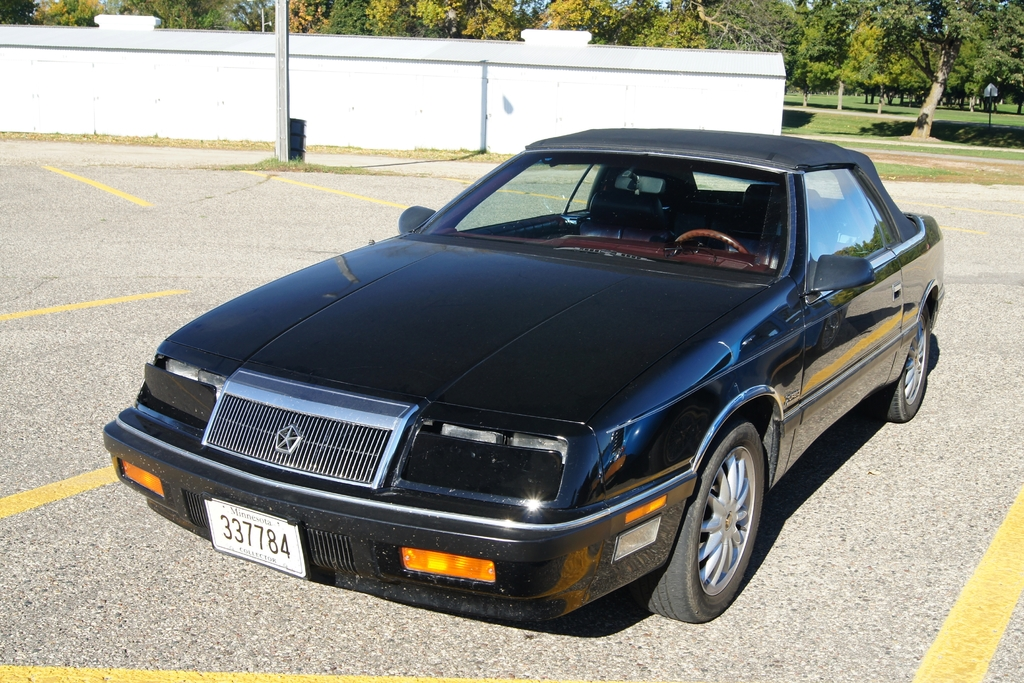}\\[-0.2em] Original
\end{minipage}\hfill
\begin{minipage}[t]{0.46\textwidth}\centering
\includegraphics[width=0.95\linewidth,height=4.2cm,keepaspectratio=false]{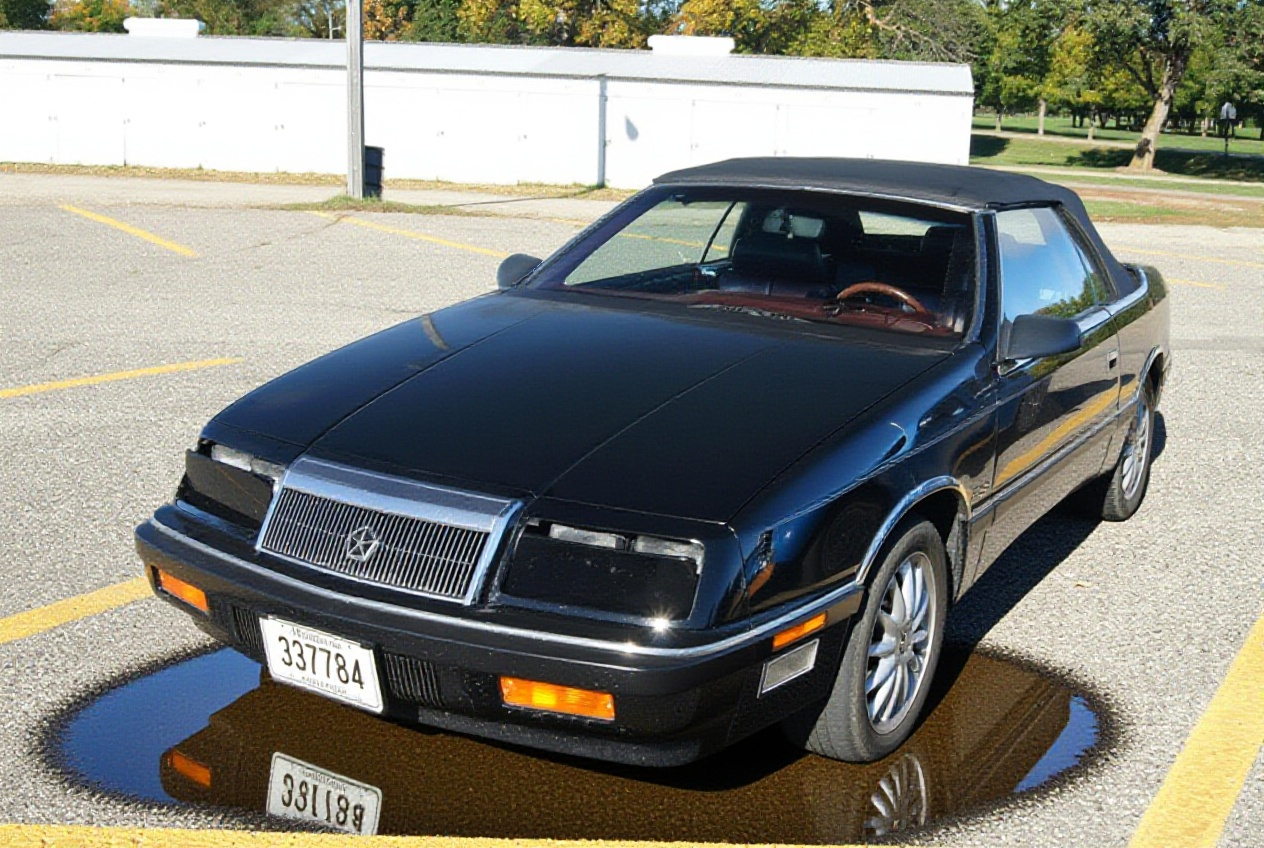}\\[-0.2em] Edited
\end{minipage}
\end{center}
\textbf{Edit:} Add a small realistic puddle near the front tire of the coupe, with reflections of the car and surroundings.\par
\textbf{Visual Difference:} A dark puddle appears under the front bumper, with reflected car details visible in the water.\par
\textbf{Judge Score/rationale:} match=5, penalty=0, final=5. The added puddle and its local reflections directly support the requested edit.
\end{minipage}}
\vspace{2mm}
\fbox{\begin{minipage}{0.95\textwidth}
\begin{center}
\begin{minipage}[t]{0.46\textwidth}\centering
\includegraphics[width=0.95\linewidth,height=4.2cm,keepaspectratio=false]{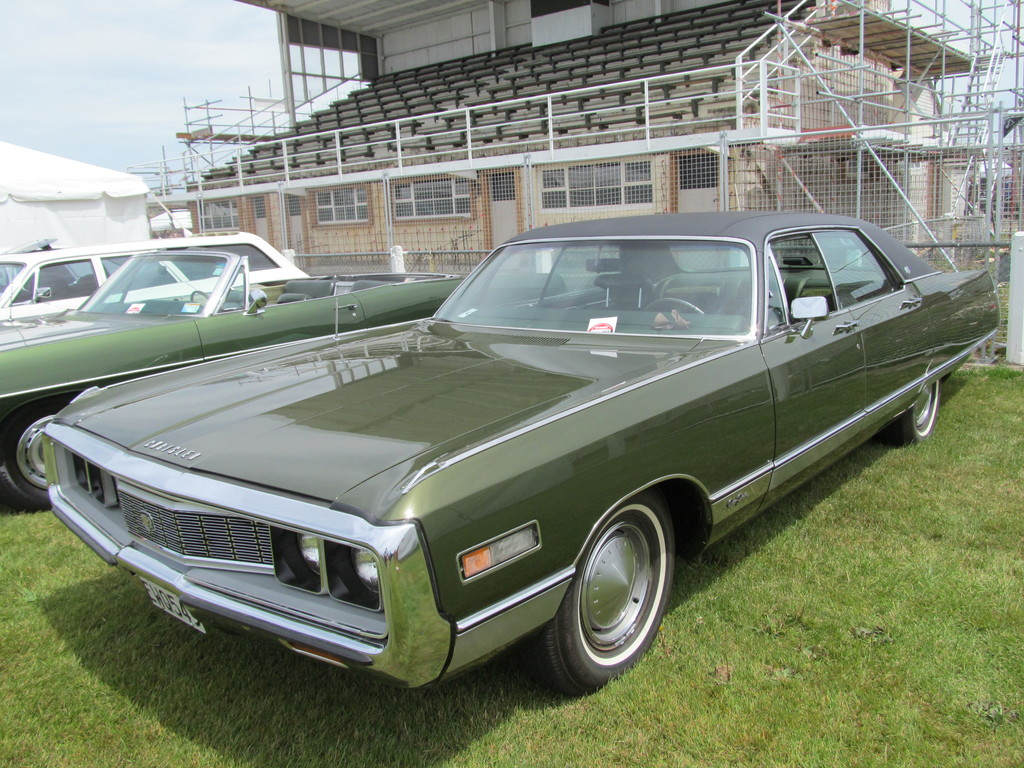}\\[-0.2em] Original
\end{minipage}\hfill
\begin{minipage}[t]{0.46\textwidth}\centering
\includegraphics[width=0.95\linewidth,height=4.2cm,keepaspectratio=false]{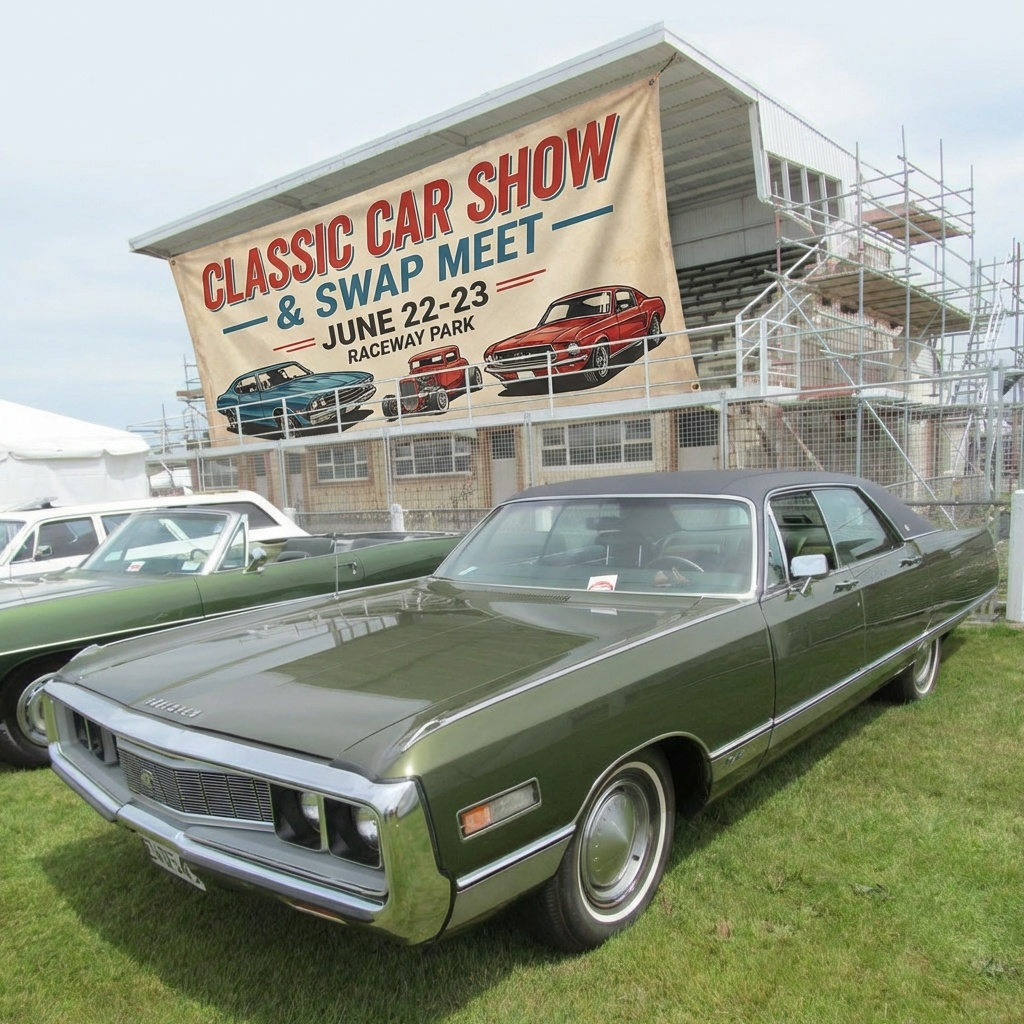}\\[-0.2em] Edited
\end{minipage}
\end{center}
\textbf{Edit:} Tilt the grandstand slightly right and reveal a hidden event banner behind it.\par
\textbf{Visual Difference:} A large car-show banner appears. A new roof-like support structure is also added.\par
\textbf{Judge Score/rationale:} match=4, penalty=2, final=2. The banner is present, but the grandstand is not tilted. And the new support frame is a prominent off-target change.
\end{minipage}}
\caption{Judge scoring and rationale examples. }
\label{fig:app_filtering_examples_rejects}
\end{figure*}

\paragraph{Judge Examples}
\label{app:judge_decision_examples}
Figure \ref{fig:app_filtering_examples_rejects} shows \texttt{KEEP} and \texttt{DISCARD} decisions.

\section{Perturbation Suite}
\label{app:perturbations}
\begin{figure*}[!htbp]
\centering
\scriptsize
\setlength{\fboxsep}{5pt}

\fbox{\begin{minipage}{0.95\textwidth}
\begin{center}
\begin{minipage}[t]{0.46\textwidth}\centering
\includegraphics[width=0.95\linewidth,height=4.2cm,keepaspectratio=false]{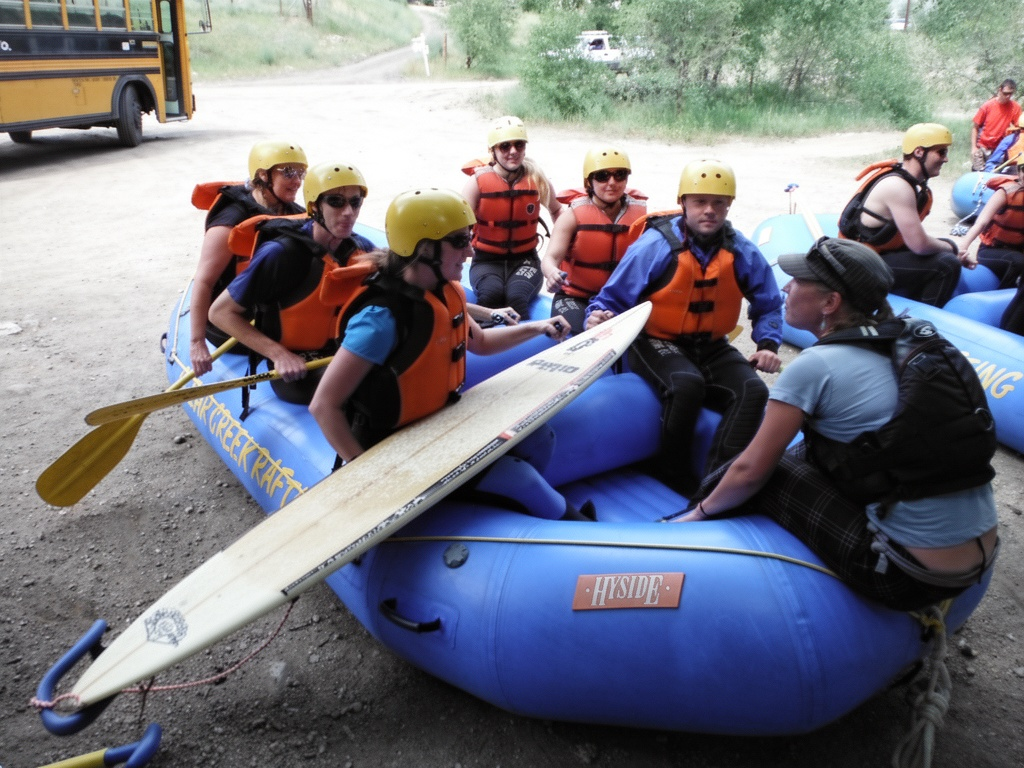}\\[-0.2em] Clean
\end{minipage}\hfill
\begin{minipage}[t]{0.46\textwidth}\centering
\includegraphics[width=0.95\linewidth,height=4.2cm,keepaspectratio=false]{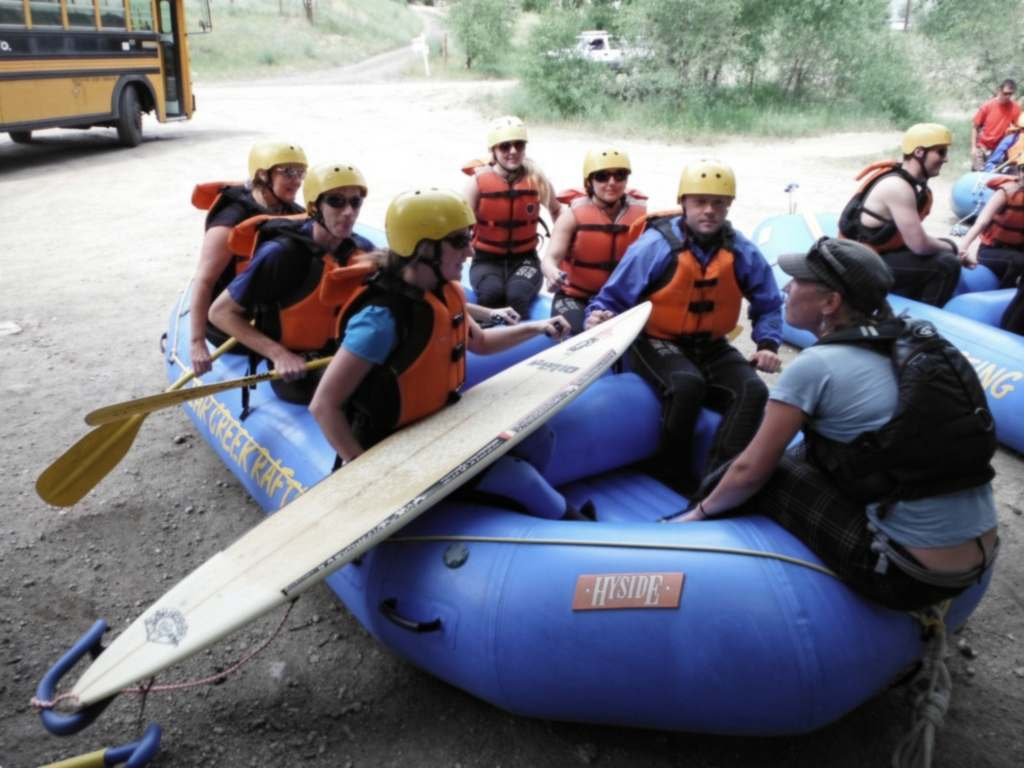}\\[-0.2em] Perturbed
\end{minipage}
\end{center}
Pair perturbation example using \texttt{gaussian\_blur} and \texttt{tiny\_local\_jpeg\_corruption}.
\end{minipage}}
\vspace{0.8em}

\fbox{\begin{minipage}{0.95\textwidth}
\begin{center}
\begin{minipage}[t]{0.46\textwidth}\centering
\includegraphics[width=0.95\linewidth,height=4.2cm,keepaspectratio=false]{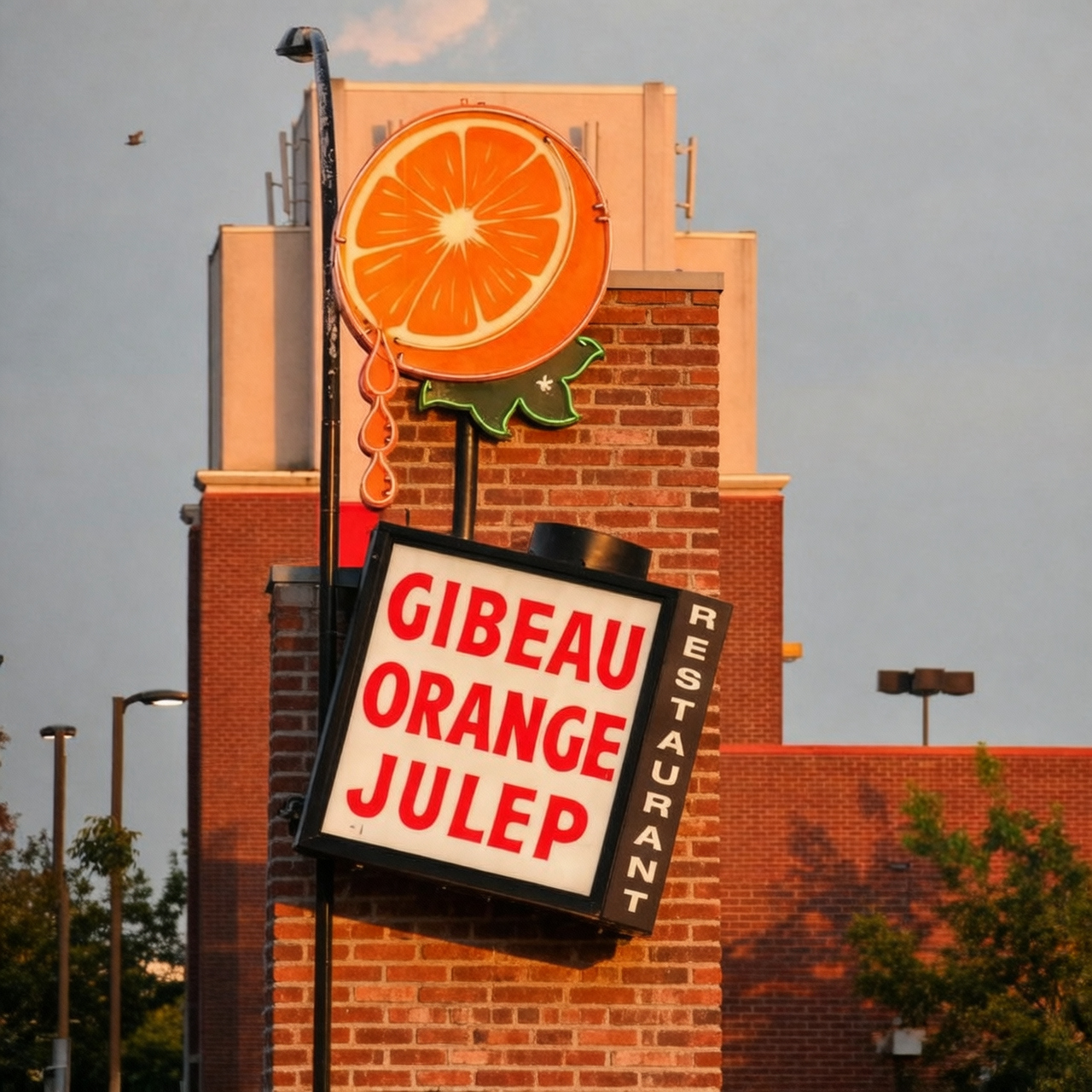}\\[-0.2em] Clean
\end{minipage}\hfill
\begin{minipage}[t]{0.46\textwidth}\centering
\includegraphics[width=0.95\linewidth,height=4.2cm,keepaspectratio=false]{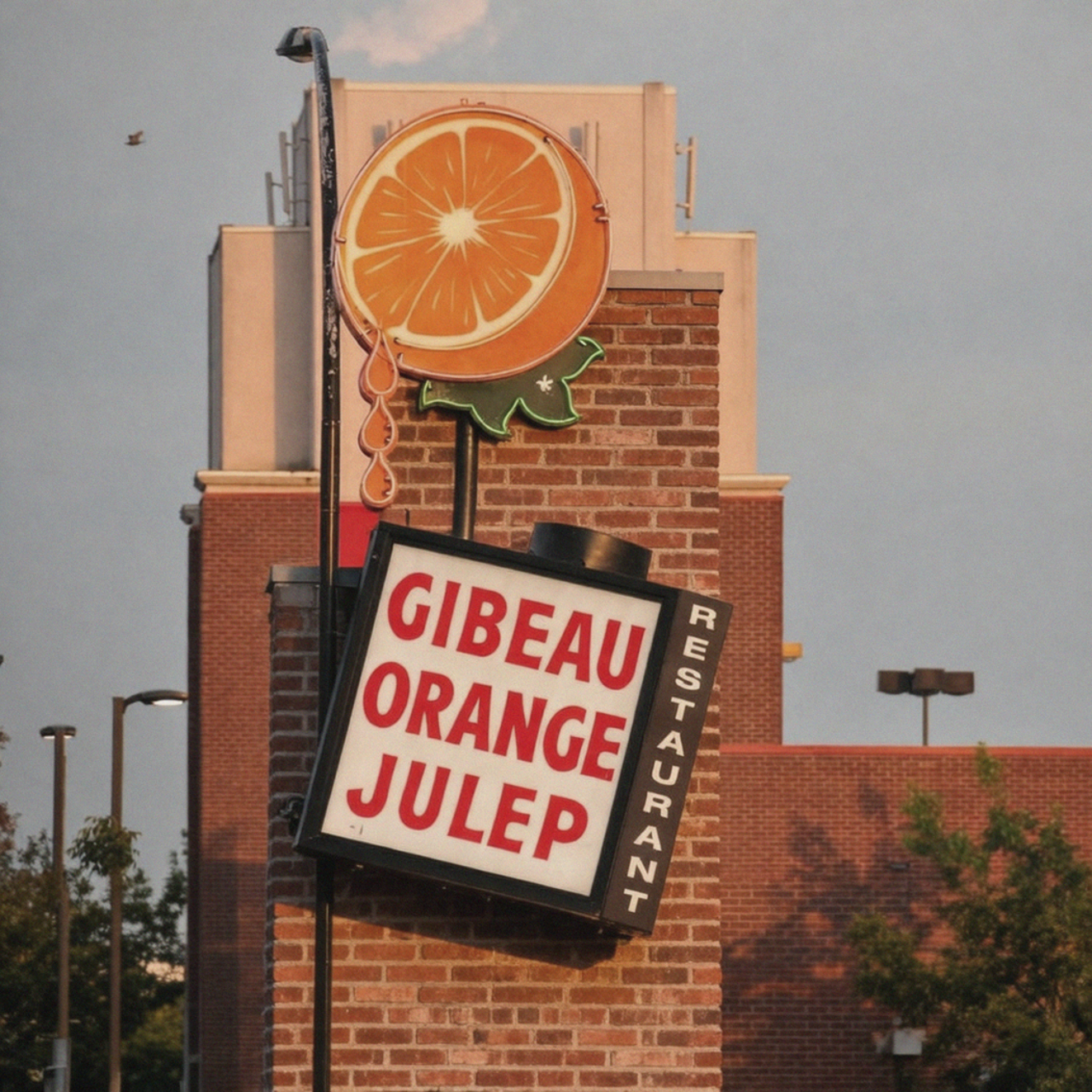}\\[-0.2em] Perturbed
\end{minipage}
\end{center}
Chunk perturbation example combining \texttt{color\_jitter}, \texttt{gamma\_tone\_curve}, \texttt{gaussian\_noise}, and \texttt{resize\_resample} etc.
\end{minipage}}
\vspace{0.8em}

\fbox{\begin{minipage}{0.95\textwidth}
\begin{center}
\begin{minipage}[t]{0.46\textwidth}\centering
\includegraphics[width=0.95\linewidth,height=4.2cm,keepaspectratio=false]{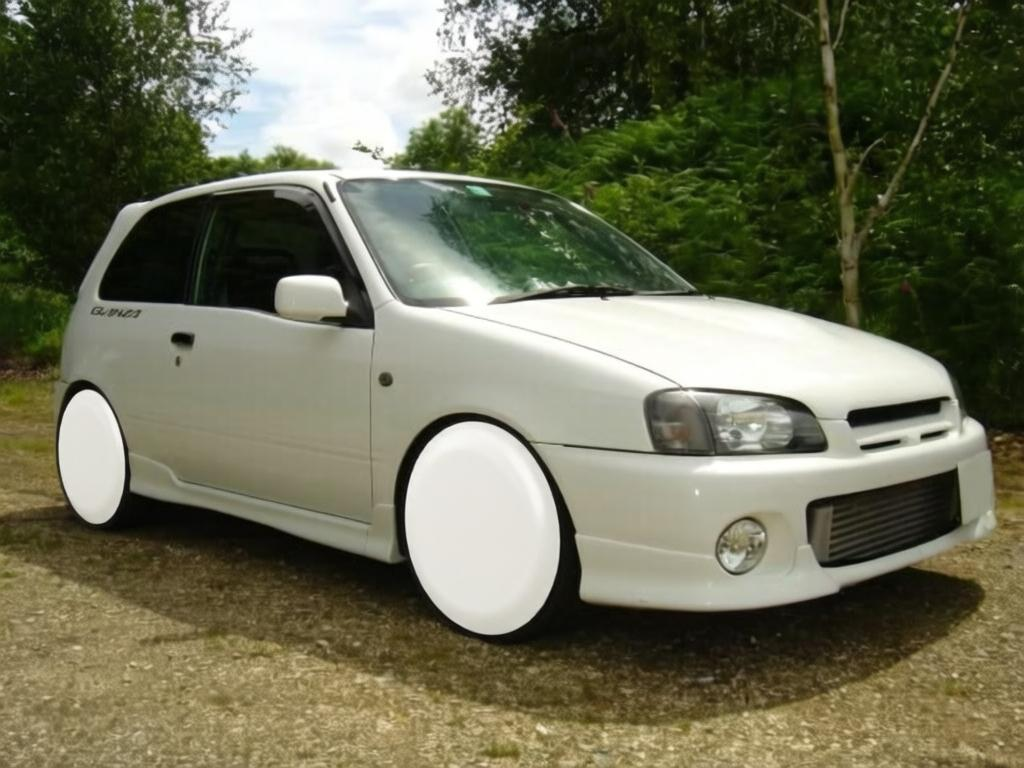}\\[-0.2em] Clean
\end{minipage}\hfill
\begin{minipage}[t]{0.46\textwidth}\centering
\includegraphics[width=0.95\linewidth,height=4.2cm,keepaspectratio=false]{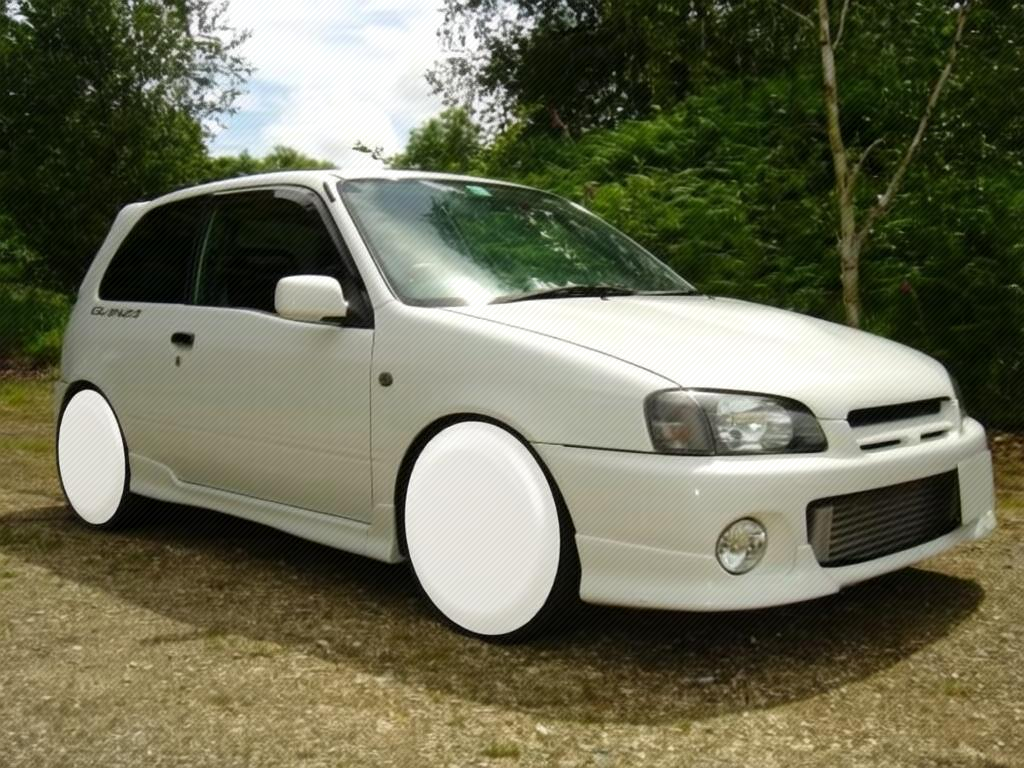}\\[-0.2em] Perturbed
\end{minipage}
\end{center}
Single perturbation example using \texttt{checkerboard\_highfreq}.
\end{minipage}}

\caption{Qualitative perturbation examples. Perturbations alter low-level image statistics while preserving the semantic content needed for detection and explanation evaluation.}
\label{fig:app_perturbation_examples}
\end{figure*}

This section summarizes the perturbations applied to \dataset. Corruptions are sampled with deterministic per-split schedules, and adversarial stress tests include FOA-style and xTransfer attacks. The validation set builds upon the training perturbation set, and the test set further expands the validation and training set with additional adversarial attacks and corruption methods. Table~\ref{tab:app_perturbation_inventory} groups the perturbations by the kind of image change they introduce, while Table \ref{tab:app_perturbation_schedule} groups perturbations by train, validation, and test split. Table~\ref{tab:app_perturbation_params} gives representative parameter ranges of each perturbation family for reproducibility. Figure~\ref{fig:app_perturbation_examples} shows three pairs of clean images with their perturbed versions under different perturbation settings. 

\begin{table}[!htbp]
\centering
\caption{Perturbation inventory by group. Validation includes all training operations; test includes all validation operations.}
\label{tab:app_perturbation_inventory}
\small
\setlength{\tabcolsep}{4pt}
\renewcommand{\arraystretch}{1.12}
\begin{tabularx}{\linewidth}{@{}p{0.24\linewidth}Y@{}}
\toprule
\textbf{Group} & \textbf{Operations} \\
\midrule
Noise and pixel-level changes & Gaussian noise; Poisson noise; speckle noise; sparse pixel flip; LSB intensity perturbation \\
Blur and restoration-like changes & Gaussian blur; sharpen; tiny local blur; mild denoise; motion blur; defocus blur \\
Compression & JPEG recompression; double JPEG; local block JPEG/compression \\
Color and tone changes & Bit-depth reduction; tone curve; chromatic aberration; vignetting; channel mixing; color jitter; grayscale/desaturation \\
Adversarial stress tests & FOA-style attack; xTransfer attack \\
Local patch changes & Tiny local JPEG corruption; tiny local mask; nearby patch replacement; tiny patch shuffle \\
Geometry and resampling & Resize/resample; small rotation/scale/translation transform \\
Frequency artifacts & Ringing halo; checkerboard high-frequency pattern; Fourier-domain perturbation \\
\bottomrule
\end{tabularx}
\renewcommand{\arraystretch}{1.0}
\end{table}

\begin{table}[!htbp]
\centering
\caption{Perturbation distribution by data split.}
\label{tab:app_perturbation_schedule}
\small
\setlength{\tabcolsep}{5pt}
\begin{tabular}{lccc}
\toprule
\textbf{Split} & \textbf{Single operation} & \textbf{Pair} & \textbf{Chunk} \\
\midrule
Train & 85\% & 10\% & 5\% of 6--8 operations \\
Validation & 75\% & 15\% & 10\% of 8--12 operations \\
Test & 70\% & 20\% & 10\% of 13--17 operations \\
\bottomrule
\end{tabular}
\end{table}


\begingroup
\scriptsize
\setlength{\tabcolsep}{3pt}
\renewcommand{\arraystretch}{1.08}
\begin{longtable}{@{}p{0.24\linewidth}p{0.13\linewidth}p{0.40\linewidth}p{0.18\linewidth}@{}}
\caption{Parameter ranges for individual perturbations. Rows use the perturbation names and split membership from \texttt{add\_noise\_for\_mapping\_fast\_skip\_existing.py}; FOA and xTransfer are generated by separate adversarial pipelines.}
\label{tab:app_perturbation_params}\\
\toprule
\textbf{Perturbation} & \textbf{Split(s)} & \textbf{Parameter range} & \textbf{Implementation notes} \\
\midrule
\endfirsthead
\caption[]{Parameter ranges for individual perturbations (continued).}\\
\toprule
\textbf{Perturbation} & \textbf{Split(s)} & \textbf{Parameter range} & \textbf{Implementation notes} \\
\midrule
\endhead
\midrule
\multicolumn{4}{r}{\emph{Continued on next page}}\\
\endfoot
\bottomrule
\endlastfoot
\texttt{jpeg\_recompression} & Train / Val. / Test & JPEG quality 50--85 & OpenCV JPEG encode/decode \\
\texttt{gaussian\_noise} & Train / Val. / Test & mean 0.0; standard deviation 7.0--15.0 & Additive pixel-space Gaussian noise \\
\texttt{poisson\_noise} & Train / Val. / Test & Poisson peak 40.0--100.0 & Poisson sampling after scaling image to $[0,1]$ \\
\texttt{speckle\_noise} & Train / Val. / Test & mean 0.0; standard deviation 0.05--0.12 & Multiplicative speckle noise \\
\texttt{gaussian\_blur} & Train / Val. / Test & kernel 3 or 5; $\sigma$ 0.8--1.5 & Full-image Gaussian blur \\
\texttt{sharpen} & Train / Val. / Test & amount 0.5--1.0 & Unsharp masking using a Gaussian-smoothed image \\
\texttt{bit\_depth\_reduction} & Train / Val. / Test & bits 4, 5, or 6 & Uniform quantization of RGB intensities \\
\texttt{gamma\_tone\_curve} & Train / Val. / Test & gamma 0.85--1.25; gain 0.95--1.05 & Global tone curve \\
\texttt{chromatic\_aberration} & Train / Val. / Test & channel shift 1 px & Shifts red and blue channels in opposite directions \\
\texttt{vignetting} & Train / Val. / Test & radial Gaussian $\sigma$ from image width$/3$ to image width & Runtime default in the perturbation function \\
\texttt{color\_channel\_mixing} & Train / Val. / Test & alpha 0.02--0.05; beta 0.02--0.05 & Small RGB channel cross-talk \\
\texttt{color\_jitter} & Train / Val. / Test & saturation factor 0.85--1.15; hue shift $\pm$5; value factor 0.9--1.1 & HSV-space color jitter \\
\texttt{sparse\_pixel\_flip} & Train / Val. / Test & density 0.00005--0.0003; mode \texttt{lsb\_flip} or \texttt{delta}; bit 0 or 1; max delta 1 or 2 & Sparse near-invisible pixel edits \\
\texttt{tiny\_local\_blur} & Train / Val. / Test & relative patch size 0.015--0.03; kernel 3 or 5; $\sigma$ 0.8--1.5; alpha 0.8--1.0 & Blends a small local patch with its blurred version \\
\texttt{lsb\_intensity\_perturbation} & Train / Val. / Test & density 0.002--0.01; mode \texttt{signed\_delta} or \texttt{bit\_toggle}; step 1 or 2; bit 0 or 1 & Dense but tiny intensity perturbation \\
\texttt{double\_jpeg} & Val. / Test & first JPEG quality 82--92; second JPEG quality 65--82 & Two-stage JPEG recompression \\
\texttt{mild\_denoise} & Val. / Test & method \texttt{bilateral}, \texttt{median}, or \texttt{nlm}; bilateral $d=5$, $\sigma_\mathrm{color}$ 20--40, $\sigma_\mathrm{space}$ 20--40; median kernel 3 or 5; NLM $h$ and $h_\mathrm{color}$ 3--7 & Restoration-like denoising \\
\texttt{motion\_blur} & Val. / Test & kernel 7, 9, or 11; angle 0--180 degrees & Directional blur kernel \\
\texttt{defocus\_blur} & Val. / Test & disk radius 3, 4, or 5 & Defocus-style disk blur \\
\texttt{tiny\_local\_jpeg\_corruption} & Val. / Test & relative patch size 0.012--0.03; JPEG quality 35--65; number of patches 1, 2, or 3 & Local JPEG artifacts in tiny regions \\
\texttt{tiny\_local\_mask} & Val. / Test & relative patch size 0.012--0.03; mode \texttt{blur\_fill}, \texttt{neighbor\_copy}, or \texttt{mean\_fill}; blur-fill kernel 5 and $\sigma$ 1.0--2.0 & Small local masking or replacement \\
\texttt{patch\_replacement\_nearby} & Val. / Test & relative patch size 0.012--0.03; max offset fraction 0.03--0.08; blend alpha 0.85--1.0 & Replaces a tiny patch with a nearby patch \\
\texttt{grayscale\_or\_desaturate} & Test & mode \texttt{full\_gray} or \texttt{partial\_desat}; partial desaturation factor 0.2--0.8 & Held-out color-removal stressor \\
\texttt{resize\_resample} & Test & resize scale 0.6--0.9 & Downsample with area interpolation, then upsample with nearest, linear, or cubic interpolation \\
\texttt{small\_geometric} & Test & rotation $\pm$3 degrees; scale 0.97--1.03; x/y translation fraction $\pm$0.02 & Small affine transform with reflected borders \\
\texttt{local\_block\_compression} & Test & block size 32 or 64; JPEG quality 60--85; number of blocks 2--5 & Local block-level JPEG artifacts \\
\texttt{ringing\_halo} & Test & $\sigma$ 1.2--2.2; amount 0.25--0.6; edge boost 0.7--1.0 & Edge-conditioned halo/ringing artifact \\
\texttt{tiny\_patch\_shuffle} & Test & relative patch size 0.015--0.03; block size 2 or 4 & Shuffles blocks within a tiny local patch \\
\texttt{checkerboard\_highfreq} & Test & strength 0.8--1.8; period 1 or 2; alpha 1.0 & Weak checkerboard high-frequency residual \\
\texttt{fourier\_domain\_perturbation} & Test & amplitude 0.3--1.0; low cutoff 0.35--0.5; high cutoff $\max(\mathrm{low}+0.1,0.85)$--1.0 & Band-limited Fourier-domain residual \\
\midrule
\multicolumn{4}{@{}l}{\emph{Separate adversarial pipelines}}\\
\texttt{xTransfer} & Test & $\ell_\infty$ non-targeted, $\epsilon=12/255$; base/mid/large attacker ratio 0.2/0.2/0.6; seed 42 & XTransferBench mapping pipeline \\
\texttt{FOA-style attack} & Val. / Test & MIFGSM, $\epsilon=8$, 100 steps, $\alpha=1$, input resolution 224; seed 2023 & CLIP B16/B32/LAION surrogate ensemble, cluster number 10 \\
\end{longtable}
\renewcommand{\arraystretch}{1.0}
\endgroup

\paragraph{Mixture Protocol}
For each image, the perturbation sampler uses one of three schedules: \textit{single}, \textit{pair}, or \textit{chunk}. A \textit{single} schedule samples one operation from the split-specific allowed set. A \textit{pair} schedule samples two random operations from the allowed set. A \textit{chunk} schedule samples a larger set without replacement from all operations allowed for the split. The default schedule probabilities are 85/10/5 for train, 75/15/10 for validation, and 70/20/10 for test, with chunk sizes increasing from 6-8 to 8-12 to 13-17 operations, as specified in Table \ref{tab:app_perturbation_schedule}. This design makes the perturbation distribution cumulative while increasing severity at validation and test time.

\section{Explanation Data Details and Qualitative Examples}
\label{app:explanation_examples}
This section illustrates the explanation supervision provided in XPlainVerse. It includes complex explanations and simple explanations for fake images, and authenticity explanations for real images.

\paragraph{Explanation Types and Acceptance Criteria}
Table~\ref{tab:app_explanation_types} summarizes the explanation targets used in the dataset.
\begin{table}[!htbp]
\centering
\caption{Explanation types in \dataset{} and criteria for acceptable outputs.}
\label{tab:app_explanation_types}
\small
\setlength{\tabcolsep}{4pt}
\begin{tabularx}{\linewidth}{p{0.20\linewidth}p{0.20\linewidth}Yp{0.15\linewidth}}
\toprule
\textbf{Explanation type} & \textbf{Audience} & \textbf{Expected content} & \textbf{Typical length} \\
\midrule
Complex fake explanation & Technical / forensic user & Multiple concrete visual cues, manipulated entities, physical inconsistencies, localized evidence, and minimal unsupported speculation & 2-4 sentences \\
Simple fake explanation & Non-technical user & One salient reason in plain language; no technical jargon; accessible wording  & 1-2 sentences \\
Authenticity explanation & General user & Natural details supporting a real label, e.g., physical contact, lighting, texture, small imperfections, and object interactions & 1-3 sentences \\
Human reference explanation & Gold-standard evaluation subset & Human-written visible manipulation cues grounded in fake-image evidence & 2-4 sentences \\
\bottomrule
\end{tabularx}
\end{table}

\begin{figure*}[!htbp]
\centering
\scriptsize
\setlength{\fboxsep}{5pt}

\fbox{\begin{minipage}{0.95\textwidth}
\begin{center}
\begin{minipage}[t]{0.46\textwidth}\centering
\xvPlaceholderImage[0.95\linewidth]{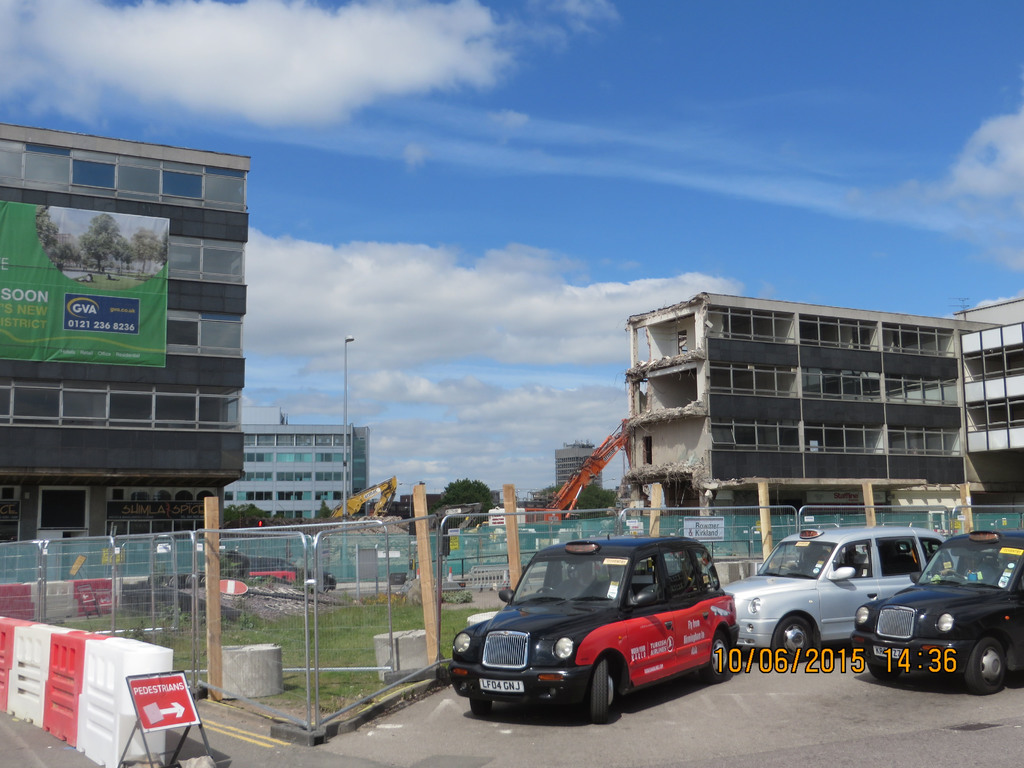}\\[-0.2em] Original
\end{minipage}\hfill
\begin{minipage}[t]{0.46\textwidth}\centering
\xvPlaceholderImage[0.95\linewidth]{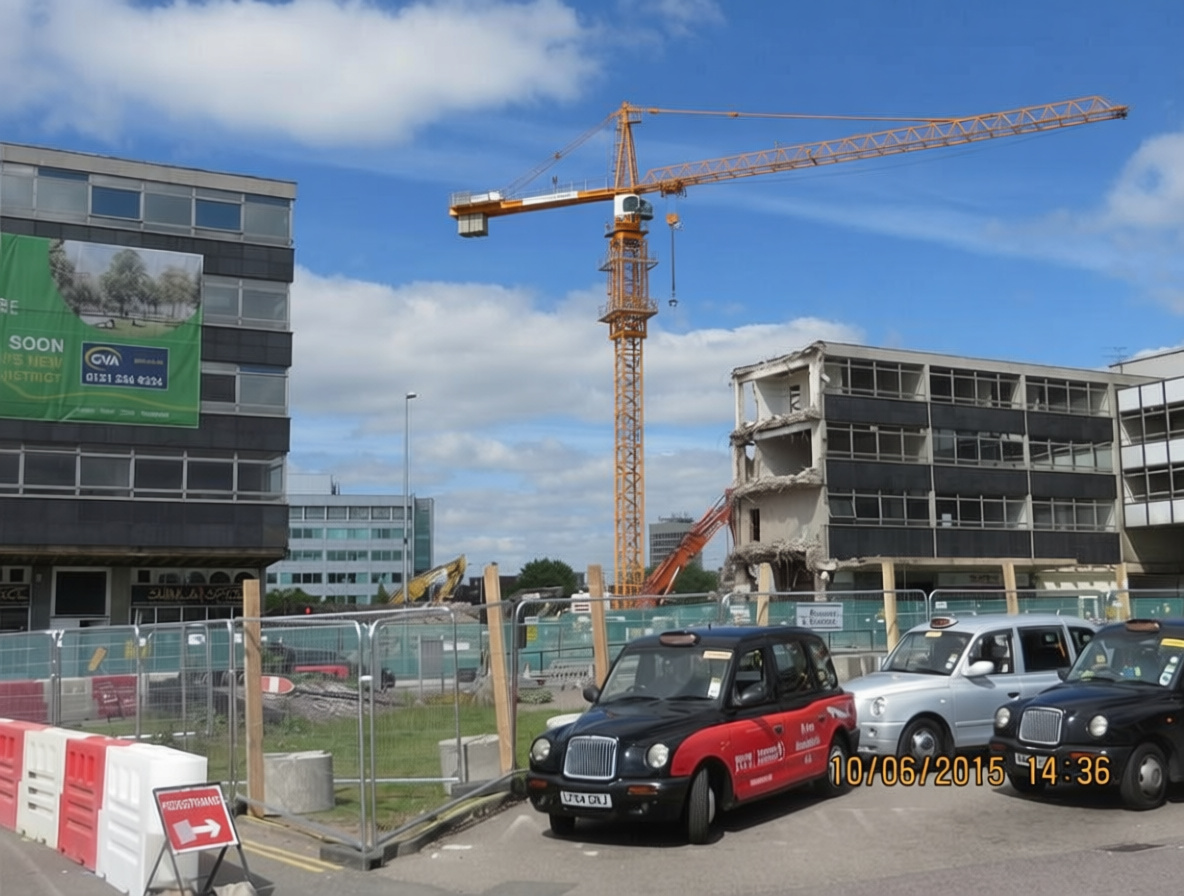}\\[-0.2em] Edited
\end{minipage}
\end{center}
\textbf{Edit/diff:} Add a large construction crane in the background, partially obscured by the partially demolished building, ensuring it blends with the existing lighting and perspective.\par\smallskip
\textbf{Complex:} The image contains several visual inconsistencies that suggest manipulation. Most notably, the large yellow construction crane in the center background appears to be floating; its base does not realistically connect to the ground or the surrounding construction debris, and it lacks any contact shadows on the terrain below. The crane's lattice structure also shows signs of digital artifacts, with some of the thin metal beams appearing slightly blurred or disconnected against the blue sky. Additionally, the lighting on the crane is overly uniform and does not perfectly match the harsh, directional sunlight seen on the buildings and taxis in the foreground, making it look like a flat overlay rather than a physical part of the scene.\par\smallskip
\textbf{Simple:} The tall yellow crane in the middle looks like a sticker that was placed on top of the picture. Its bottom doesn't touch the ground or sit on the dirt like a real heavy machine should, so it looks like it is just floating in the air.
\end{minipage}}
\vspace{0.8em}

\fbox{\begin{minipage}{0.95\textwidth}
\begin{center}
\begin{minipage}[t]{0.46\textwidth}\centering
\xvPlaceholderImage[0.95\linewidth]{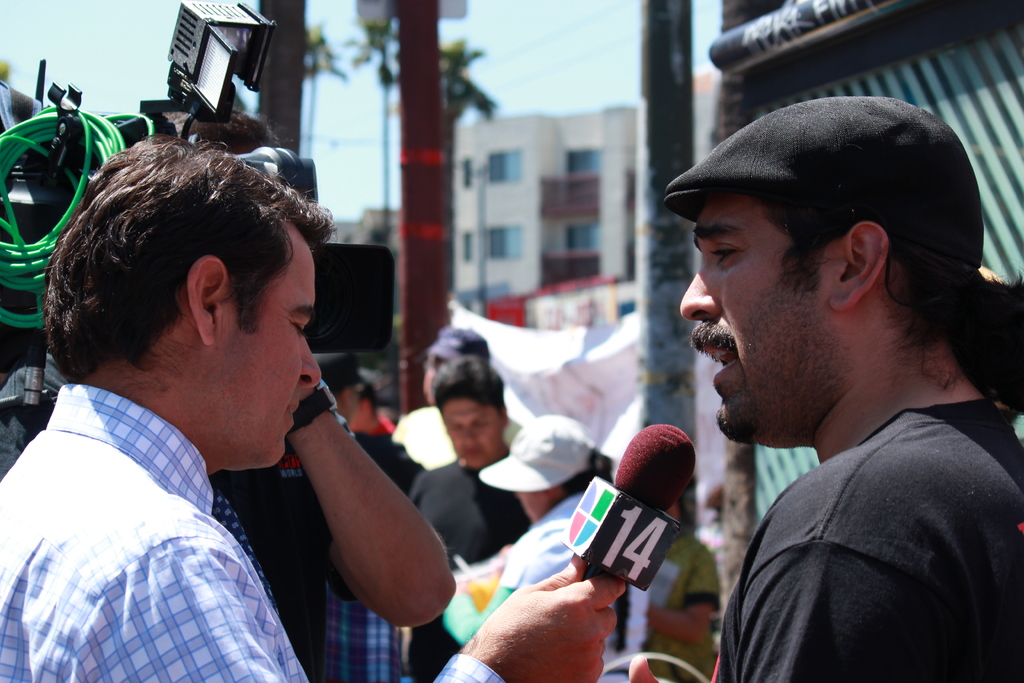}\\[-0.2em] Original
\end{minipage}\hfill
\begin{minipage}[t]{0.46\textwidth}\centering
\xvPlaceholderImage[0.95\linewidth]{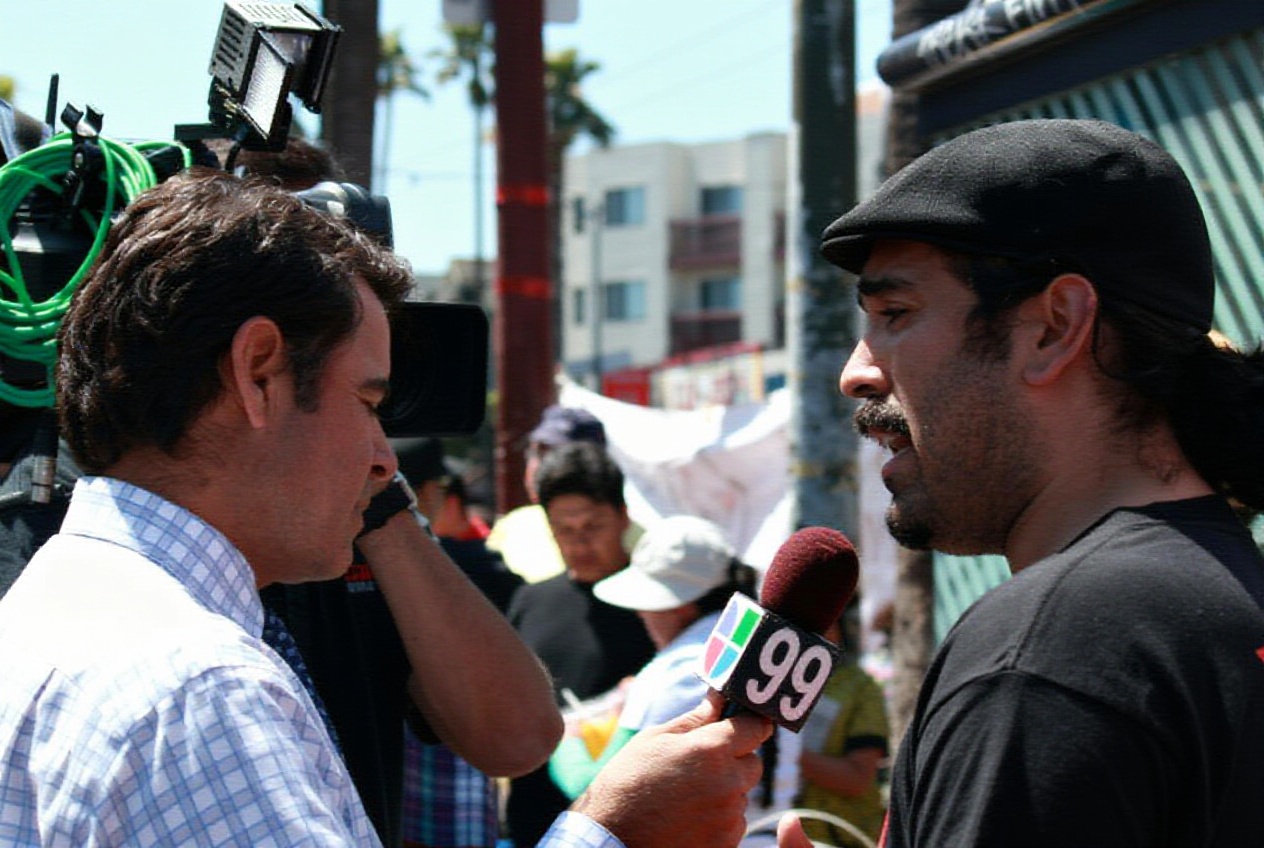}\\[-0.2em] Edited
\end{minipage}
\end{center}
\textbf{Edit/diff:} Replace the microphone labeled '14' with a microphone labeled '99', ensuring the new label matches the original style and lighting.\par\smallskip
\textbf{Complex:} The image contains a subtle digital artifact on the microphone flag held in the center. The number '99' appears unnaturally sharp and flat compared to the textured, slightly weathered surface of the black flag it is printed on. Additionally, the white strokes of the numbers lack the consistent grain and soft lighting fall-off seen on the adjacent colorful logo, making the digits look like a clean digital overlay rather than a physical part of the equipment.\par\smallskip
\textbf{Simple:} The number 99 on the microphone looks like a white sticker that was just pasted on top. It is too clean and flat, while the rest of the microphone looks like a real, worn-out object.
\end{minipage}}
\vspace{0.8em}

\caption{Fake-image explanation examples. \textbf{Complex} explanations preserve multiple grounded cues; \textbf{Simple} explanations select the single clearest cue and use non-technical language.}
\label{fig:app_fake_explanation_examples}
\end{figure*}

\begin{figure*}[!htbp]
\centering
\scriptsize
\setlength{\fboxsep}{5pt}
\fbox{\begin{minipage}{0.95\textwidth}
\begin{center}
\begin{minipage}[t]{0.46\textwidth}\centering
\includegraphics[width=0.95\linewidth,height=4.2cm,keepaspectratio=false]{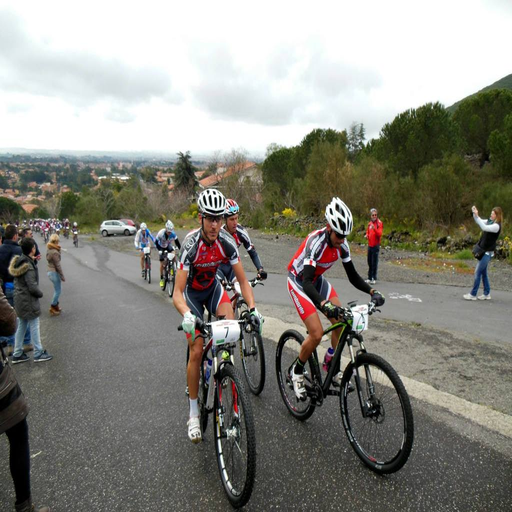}\\[-0.2em] Real image
\end{minipage}\hfill
\begin{minipage}[t]{0.46\textwidth}\centering
\includegraphics[width=0.95\linewidth,height=4.2cm,keepaspectratio=false]{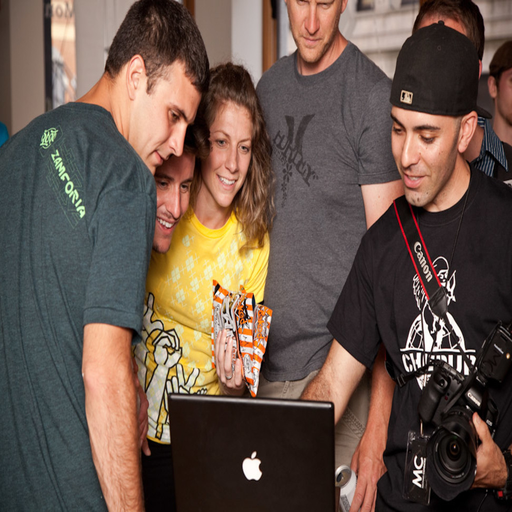}\\[-0.2em] Real image
\end{minipage}
\end{center}
\begin{center}
\begin{minipage}[t]{0.46\textwidth}\centering
This looks real because the road isn't perfect. You can see little cracks and spots on it, like a real road that people ride bikes on every day.
\end{minipage}\hfill
\begin{minipage}[t]{0.46\textwidth}\centering
This looks real because the girl's hair is a little bit messy and sticking out, like real hair does.
\end{minipage}
\end{center}
\end{minipage}}
\caption{Authenticity explanation examples for real images. Each explanation should point to visible physical evidence rather than merely asserting that the image is real.}
\label{fig:app_real_explanation_examples}
\end{figure*}

\noindent\textbf{Qualitative Examples for Reasoning}
\label{app:fake_explanation_examples}
Figure~\ref{fig:app_fake_explanation_examples} provides fake-image explanation examples while Figure \ref{fig:app_real_explanation_examples} shows real-image explanation examples in \dataset.

\section{Human Annotation and Evaluation}
\label{app:human_study}
The main paper reports aggregate human-evaluation findings. This appendix section discusses the human annotation and user study in detail.

\paragraph{Human Reference Annotation Protocol}
For the gold-reference subset, annotators write image-grounded explanations for manipulated images. Each task is presented as a paired-image comparison: the original image and the edited image are shown together. Annotators are asked to describe visible evidence in the edited image and avoid claims that are not visible. The final release includes only aggregate annotator metadata and non-identifying examples. The gold-reference annotation pool contains 12 annotators. 

\paragraph{Seven-Stage Human Evaluation Protocol}
Table~\ref{tab:app_human_study_protocol} records the tasks used in the human evaluation.
\begin{table*}[!htbp]
\centering
\caption{Seven-stage human evaluation protocol.}
\label{tab:app_human_study_protocol}
\small
\setlength{\tabcolsep}{4pt}
\begin{tabularx}{\textwidth}{p{0.13\textwidth}p{0.20\textwidth}Yp{0.15\textwidth}}
\toprule
\textbf{Stage} & \textbf{Goal} & \textbf{Questions / criteria} & \textbf{Response type} \\
\midrule
1. Image realism & Judge whether individual images appear real or manipulated & Real/fake judgment, confidence, visual realism & Choice + 1--5 scale \\
2. Edit quality & Evaluate paired original and edited images & Core edit presence, edit realism, unrelated off-target changes & 1--5 scale \\
3. Explanation quality & Rate one explanation for one image & Grounding, specificity, completeness, convincingness, unsupported claims & 1--5 scale \\
4. Explanation preference & Compare two explanations for the same image & Grounding, specificity, completeness, fewer unsupported claims, overall usefulness & A/B/Tie \\
5. Human vs. generated comparison & Compare human-written and generated complex explanations & Grounding, specificity, completeness, fewer unsupported claims, overall preference & A/B/Tie \\
6. Readability & Evaluate simple/user-facing explanations & Ease of understanding, memorability, cognitive load, liking & 1--5 scale \\
7. Audience usefulness & Compare explanations for different users & Personal preference, usefulness for ordinary users, usefulness for technical analysts, clarity/evidence balance, default system choice & Choice \\
\bottomrule
\end{tabularx}
\end{table*}

\begin{table*}[!htbp]
\centering
\caption{Human evaluation results by stage. Values are sample-balanced: we first aggregate responses within each evaluated sample and then average across samples. The $n$ column reports unique samples / total ratings. For pairwise rows, percentages are ordered as the listed choice columns.}
\label{tab:app_human_study_results}
\scriptsize
\setlength{\tabcolsep}{3.5pt}
\renewcommand{\arraystretch}{0.95}
\begin{tabularx}{\textwidth}{p{0.06\textwidth}p{0.23\textwidth}p{0.11\textwidth}Y}
\toprule
\textbf{Stage} & \textbf{Measure} & \textbf{$n$} & \textbf{Result} \\
\midrule
1 & Real images & 21 / 119 & Judged real 60.6\%, manipulated 32.7\%, not sure 6.7\%; realism mean 3.93/5. \\
1 & Retained fake images & 21 / 117 & Judged real 44.9\%, manipulated 46.1\%, not sure 9.0\%; realism mean 3.63/5. \\
1 & Discarded fake images & 8 / 44 & Judged real 24.4\%, manipulated 69.7\%, not sure 6.0\%; realism mean 3.29/5. \\
\midrule
2 & Filtered edit quality & 13 / 72 & Composite 3.66/5; edit presence 4.09/5; edit realism 3.35/5; off-target changes 2.47/5. \\
2 & Discarded edit quality & 12 / 68 & Composite 2.71/5; edit presence 2.62/5; edit realism 2.69/5; off-target changes 3.17/5. \\
\midrule
3 & Complex explanation quality & 28 / 168 & Composite 3.79/5; grounding 3.82; specificity 3.96; completeness 3.78; convincingness 3.79; unsupported claims 2.43 (lower is better). \\
\midrule
4 & Complex vs.\ SIDA: grounding & 28 / 168 & Complex 64.7\%, tie 12.5\%, SIDA 22.7\%; $\Delta=+42.0$. \\
4 & Complex vs.\ SIDA: specificity & 28 / 168 & Complex 60.8\%, tie 14.3\%, SIDA 25.0\%; $\Delta=+35.8$. \\
4 & Complex vs.\ SIDA: completeness & 28 / 168 & Complex 58.2\%, tie 16.9\%, SIDA 24.9\%; $\Delta=+33.3$. \\
4 & Complex vs.\ SIDA: fewer unsupported claims & 28 / 168 & Complex 46.5\%, tie 30.5\%, SIDA 23.1\%; $\Delta=+23.4$. \\
4 & Complex vs.\ SIDA: overall choice & 28 / 168 & Complex 64.8\%, tie 13.1\%, SIDA 22.2\%; $\Delta=+42.6$. \\
\midrule
5 & Complex vs.\ human: grounding & 30 / 168 & Complex 60.3\%, tie 7.1\%, human 32.5\%; $\Delta=+27.8$. \\
5 & Complex vs.\ human: specificity & 30 / 168 & Complex 67.3\%, tie 7.8\%, human 24.9\%; $\Delta=+42.4$. \\
5 & Complex vs.\ human: completeness & 30 / 168 & Complex 64.3\%, tie 9.2\%, human 26.5\%; $\Delta=+37.8$. \\
5 & Complex vs.\ human: fewer unsupported claims & 30 / 168 & Complex 47.3\%, tie 25.5\%, human 27.2\%; $\Delta=+20.1$. \\
5 & Complex vs.\ human: overall preference & 30 / 168 & Complex 61.8\%, tie 8.2\%, human 30.0\%; $\Delta=+31.8$. \\
\midrule
6 & Simple explanation readability & 29 / 168 & Composite 3.94/5; ease of understanding 4.11; memorability 4.13; cognitive load 2.18 (lower is better); liking 3.71. \\
\midrule
7 & Personal preference & 29 / 168 & Simple 33.6\%, complex 49.3\%, both/either 14.2\%, neither 2.9\%. \\
7 & Ordinary-user usefulness & 29 / 168 & Simple 46.5\%, complex 38.3\%, both/either 13.5\%, neither 1.6\%. \\
7 & Technical-analysis usefulness & 29 / 168 & Simple 7.5\%, complex 73.9\%, both/either 16.9\%, neither 1.6\%. \\
7 & Evidence/readability balance & 29 / 168 & Simple 14.5\%, complex 54.2\%, both/either 25.2\%, neither 6.1\%. \\
7 & Default system choice & 29 / 168 & Simple 26.8\%, complex 46.1\%, both/either 23.3\%, neither 3.7\%. \\
\bottomrule
\end{tabularx}
\end{table*}

\paragraph{Participant-Facing Instructions and Interface}
\label{app:participant_instructions}

We reproduce below the participant-facing instructions used in the gold-reference annotation task
and the human-evaluation study. The study was administered through a form-based interface. All participants gave
informed consent before beginning the task. Compensation rates were determined before the study began,
were at or above the applicable local minimum-wage equivalent, and did not depend on the labels,
ratings, or judgments participants produced.

\paragraph{Gold-reference annotation instructions.}
Annotators were shown the original image and the edited image side by side. The instruction text was:

\begin{quote}
\small
You will be shown an original image and an edited image. Your task is to write an explanation
describing why the edited image appears manipulated. Focus only on visible evidence in the edited
image. You may use the original image for context, but your explanation should refer to cues that can
be observed in the edited image itself. Mention concrete regions, objects, people, boundaries, textures,
lighting, geometry, or other visible artifacts. Do not speculate about hidden causes, model names, or
metadata. If multiple cues are visible, describe all important ones clearly.
\end{quote}

\paragraph{Human-evaluation instructions.}
Participants in the human-evaluation study received the following general instruction:

\begin{quote}
\small
You will evaluate images and explanations produced for a deepfake-detection benchmark. Some
images are real and some are manipulated. Please answer based only on what is visible in the image
and the explanation shown to you. There are no trick questions. Use your best judgment, and select
``not sure'' or a middle score when the evidence is unclear. Do not include any personally identifying
information in your responses.
\end{quote}

\paragraph{Stage-specific instructions.}
For image realism, participants judged whether each image appeared real, manipulated, or uncertain,
and rated visual realism on a 1--5 scale. For edit quality, participants compared the original and edited
images and rated whether the intended edit was present, visually realistic, and free of unrelated
off-target changes. For explanation quality, participants rated grounding, specificity, completeness,
convincingness, and unsupported claims. For pairwise comparisons, participants chose which
explanation was better or selected tie. For readability, participants rated ease of understanding,
memorability, cognitive load, and liking. For audience usefulness, participants judged which
explanation was more useful for ordinary users and for technical analysts.

\section{Metric Implementation Details}
\label{app:metric_details}

This section describes the implementation details for the automatic metrics used to evaluate XPlainVerse explanations. For complex explanations, we evaluate whether a generated explanation identifies the same manipulated regions and supports them with the same visual evidence as the reference. We therefore compute two intent-aware metrics: \emph{EntityScore}, which measures coverage of diagnostic image entities, and \emph{EvidenceScore}, which measures coverage of concrete visual evidence claims. Both metrics use an offline Qwen3.5-4B evaluator for structured extraction and semantic coverage judging. We additionally report BERTScore-F1 as a complementary measure of full-sentence semantic similarity. For simple explanations, we report BERTScore-F1 for semantic fidelity and a normalized SLE score for simplicity.

\paragraph{Entity and evidence extraction.}
Given a predicted explanation $y^{p}$ and a reference explanation $y^{r}$, we first convert each free-form explanation into a structured set of diagnostic entities and evidence claims. This extraction is performed independently for the prediction and reference using the same offline Qwen3.5-4B evaluator.

A \emph{diagnostic entity} is an image region, subject, object, text element, body part, clothing item, background component, or other semantically meaningful visual element that is explicitly used as evidence in the explanation. Entities are not extracted merely because they are mentioned; they must be connected to the explanation's real/fake judgment. For example, ``the sign text,'' ``the cyclist's hands,'' ``the metal support bars,'' or ``the reflection in the window'' may be diagnostic entities if the explanation uses them as evidence. In contrast, generic mentions such as ``the image,'' ``the scene,'' or contextual objects that are not tied to a visual cue are excluded.

An \emph{evidence claim} is a compact, checkable visual observation about a diagnostic entity. Evidence claims describe what is visually suspicious or authenticity-supporting about the entity, such as distorted lettering, inconsistent boundaries, unnatural texture, impossible geometry, mismatched lighting, or physically plausible interaction. Claims are linked to their corresponding entities whenever possible. The extractor is instructed to prefer coarse, meaningful entities over overly fine-grained fragments, to avoid unstated visual inferences, and to exclude generic conclusions such as ``this image is fake'' unless they are grounded in a specific visual observation.

The expected extraction output is valid JSON:
\begin{quote}
\hrule
\vspace{0.25em}
\footnotesize
\noindent\textbf{Entity and evidence extraction output schema}\par
\vspace{0.25em}
\begin{verbatim}
{
  "diagnostic_entities": [
    {"entity_id": "E1", "name": "..."}
  ],
  "evidence_claims": [
    {"claim_id": "C1", "entity_id": "E1", "claim": "..."}
  ]
}
\end{verbatim}
\hrule
\end{quote}

We denote the extracted entity sets from the prediction and reference as $\mathcal{E}^{p}$ and $\mathcal{E}^{r}$, respectively. Similarly, we denote the extracted evidence-claim sets as $\mathcal{V}^{p}$ and $\mathcal{V}^{r}$. We use $\mathcal{V}$ rather than $\mathcal{F}$ to emphasize that these are visual evidence claims rather than general factual statements.

\paragraph{Bidirectional semantic coverage.}
After extraction, we compute semantic coverage in two directions. The first direction measures precision: whether the diagnostic entities and evidence claims mentioned by the prediction are supported by the reference explanation. The second direction measures recall: whether the diagnostic entities and evidence claims in the reference explanation are recovered by the prediction.

For precision-side coverage, we run one Qwen3.5-4B evaluator pass using the extracted JSON of the predicted explanation and the full reference explanation. The extracted prediction JSON contains the predicted diagnostic entities $\mathcal{E}^{p}$ and predicted evidence claims $\mathcal{V}^{p}$, with evidence claims linked to their corresponding entities. Given this full predicted structure and the reference explanation $y^{r}$, the evaluator judges which predicted entities and evidence claims are semantically supported by the reference. This yields entity and evidence precision, measuring whether the prediction introduces reference-supported diagnostic content rather than hallucinated regions or unsupported visual cues.

For recall-side coverage, we run a second Qwen3.5-4B evaluator pass in the reverse direction, using the extracted JSON of the reference explanation and the full predicted explanation. The extracted reference JSON contains the reference diagnostic entities $\mathcal{E}^{r}$ and reference evidence claims $\mathcal{V}^{r}$, again with evidence claims linked to their corresponding entities. Given this full reference structure and the predicted explanation $y^{p}$, the evaluator judges which reference entities and evidence claims are semantically recovered by the prediction. This yields entity and evidence recall, measuring whether the prediction captures the diagnostic regions and visual evidence present in the reference.

Formally, let
\[
\mathcal{G}^{p} = (\mathcal{E}^{p}, \mathcal{V}^{p})
\qquad \text{and} \qquad
\mathcal{G}^{r} = (\mathcal{E}^{r}, \mathcal{V}^{r})
\]
denote the extracted JSON structures for the predicted and reference explanations, respectively. Here, $\mathcal{E}$ denotes the set of diagnostic entities and $\mathcal{V}$ denotes the set of entity-linked evidence claims. Let $\mathrm{cover}(z \mid \mathcal{G}, y)$ be an indicator function that returns $1$ if source item $z$, interpreted within the full extracted structure $\mathcal{G}$, is judged to be semantically supported by target explanation $y$, and $0$ otherwise.

Entity precision and recall are computed as
\begin{align}
\mathrm{EntityPrecision}
&=
\frac{1}{|\mathcal{E}^{p}|}
\sum_{e \in \mathcal{E}^{p}}
\mathrm{cover}(e \mid \mathcal{G}^{p}, y^{r}),
\label{eq:entity_precision}
\\
\mathrm{EntityRecall}
&=
\frac{1}{|\mathcal{E}^{r}|}
\sum_{e \in \mathcal{E}^{r}}
\mathrm{cover}(e \mid \mathcal{G}^{r}, y^{p}).
\label{eq:entity_recall}
\end{align}

Evidence precision and recall are computed analogously:
\begin{align}
\mathrm{EvidencePrecision}
&=
\frac{1}{|\mathcal{V}^{p}|}
\sum_{v \in \mathcal{V}^{p}}
\mathrm{cover}(v \mid \mathcal{G}^{p}, y^{r}),
\label{eq:evidence_precision}
\\
\mathrm{EvidenceRecall}
&=
\frac{1}{|\mathcal{V}^{r}|}
\sum_{v \in \mathcal{V}^{r}}
\mathrm{cover}(v \mid \mathcal{G}^{r}, y^{p}).
\label{eq:evidence_recall}
\end{align}

The matching is wording-invariant but evidence-sensitive. For entities, exact lexical overlap is not required: for example, ``the CITY PARK letters'' and ``the large white sign text'' may be counted as the same entity if they refer to the same visible region. However, overly broad entities are not accepted when they fail to localize the diagnostic region. For evidence claims, coverage is stricter: the target explanation must express the same visual observation, not merely mention the same object. Thus, a prediction may receive entity credit for identifying the correct region while still receiving low evidence credit if it gives the wrong or incomplete visual reason.

We compute EntityScore and EvidenceScore as harmonic means of their corresponding precision and recall terms:
\begin{align}
\mathrm{EntityScore}
&=
\frac{
2 \cdot \mathrm{EntityPrecision} \cdot \mathrm{EntityRecall}
}{
\mathrm{EntityPrecision} + \mathrm{EntityRecall}
},
\label{eq:entity_score}
\\
\mathrm{EvidenceScore}
&=
\frac{
2 \cdot \mathrm{EvidencePrecision} \cdot \mathrm{EvidenceRecall}
}{
\mathrm{EvidencePrecision} + \mathrm{EvidenceRecall}
}.
\label{eq:evidence_score}
\end{align}

If both precision and recall are zero, the corresponding harmonic-mean score is set to zero. 

EntityScore captures whether the explanation identifies the correct manipulated or authenticity-relevant regions, while EvidenceScore captures whether it gives the correct visual reasons.

\paragraph{Metric extraction example.}
Table~\ref{tab:app_metric_extraction_example} gives a compact example of entity and evidence extraction followed by semantic coverage judging.

\begin{table}[!htbp]
\centering
\caption{Illustrative entity and evidence extraction example.}
\label{tab:app_metric_extraction_example}
\small
\setlength{\tabcolsep}{4pt}
\renewcommand{\arraystretch}{1.12}
\begin{tabularx}{\linewidth}{p{0.22\linewidth}Y}
\toprule
\textbf{Item} & \textbf{Example} \\
\midrule
Reference explanation &
The large CITY PARK letters look flat and poorly integrated with the metal supports, with bars appearing to pass through parts of the text. \\
\midrule
Extracted entities &
[E1: CITY PARK letters; E2: metal support bars] \\
\midrule
Extracted evidence claims &
C1/E1: letters look flat; C2/E1: letters are poorly integrated; C3/E2: support bars appear to pass through the letters. \\
\midrule
Candidate explanation &
The big white letters look like they are floating, and the metal bars go through them instead of sitting behind them. \\
\midrule
Coverage result &
Entity match: E1 covered, E2 covered. Evidence match: C1 covered, C2 covered, C3 covered. This yields illustrative EntityScore \(=1.00\) and EvidenceScore \(=1.00\). \\
\bottomrule
\end{tabularx}
\end{table}

\paragraph{Coverage-judging edge cases.}
Table~\ref{tab:app_metric_edge_cases} summarizes how we handle common edge cases during automatic coverage judging.

\begin{table}[!htbp]
\centering
\caption{Handling of common edge cases in EntityScore and EvidenceScore.}
\label{tab:app_metric_edge_cases}
\small
\setlength{\tabcolsep}{4pt}
\renewcommand{\arraystretch}{1.12}
\begin{tabularx}{\linewidth}{p{0.27\linewidth}Yp{0.23\linewidth}}
\toprule
\textbf{Case} & \textbf{Handling rule} & \textbf{Reason} \\
\midrule
Empty predicted entity set &
Entity precision is set to 0 unless both prediction and reference contain no diagnostic entities. &
Penalizes vacuous explanations. \\
\midrule
Empty reference entity set &
The example is excluded from entity/evidence scoring for fake explanations, or assigned a task-specific neutral score when explicitly defined. &
Avoids undefined recall. \\
\midrule
Synonymous region names &
Count as covered if the evaluator judges that both phrases refer to the same visible region. &
Allows natural wording variation. \\
\midrule
Overly broad entity &
Count as not covered if the phrase fails to identify the diagnostic region specifically. &
Discourages generic explanations. \\
\midrule
Unsupported extra evidence &
Counts against precision-side EvidenceScore. &
Penalizes hallucinated or irrelevant visual claims. \\
\midrule
Correct entity, wrong evidence &
Entity coverage may be positive while evidence coverage remains negative. &
Separates region identification from reasoning quality. \\
\midrule
Correct evidence, different wording &
Count as covered if the same visual observation is expressed semantically. &
Avoids over-penalizing paraphrases. \\
\bottomrule
\end{tabularx}
\end{table}

\paragraph{BERTScore and simple-explanation scoring.}
For both complex and simple explanations, we compute BERTScore-F1 between the full generated explanation and the corresponding reference explanation. BERTScore captures overall semantic similarity, but it does not explicitly test whether the explanation identifies the same manipulated entity or cites the same visual evidence. We therefore treat BERTScore as complementary to EntityScore and EvidenceScore rather than as a replacement.

For simple explanations, we additionally compute the SLE simplicity score~\cite{cripwell2023sle}. Since raw SLE values can fall outside a convenient reporting range, we clip the raw score to $[-1,4]$ and normalize it to $[0,1]$:
\begin{align}
\mathrm{SLE}_{\mathrm{clip}}
&=
\min\left(4,\max\left(-1,\mathrm{SLE}_{\mathrm{raw}}\right)\right),
\label{eq:sle_clip}
\\
\mathrm{SLE}_{\mathrm{norm}}
&=
\frac{\mathrm{SLE}_{\mathrm{clip}}+1}{5}.
\label{eq:sle_norm}
\end{align}
Higher normalized SLE indicates simpler and more accessible language. We report normalized SLE for simple explanations so that the score is directly comparable across models and remains bounded between 0 and 1.

\paragraph{Automatic metric configuration.}
Table~\ref{tab:app_metric_config} summarizes the automatic evaluators used for scoring.

\begin{table}[!htbp]
\centering
\caption{Automatic metric configuration.}
\label{tab:app_metric_config}
\small
\setlength{\tabcolsep}{4pt}
\renewcommand{\arraystretch}{1.12}
\begin{tabularx}{\linewidth}{p{0.22\linewidth}p{0.24\linewidth}Y}
\toprule
\textbf{Metric} & \textbf{Evaluator} & \textbf{Configuration} \\
\midrule
BERTScore-F1 &
BERTScore scorer &
Computed between aligned prediction and reference explanations. \\
\midrule
EntityScore &
\texttt{Qwen/Qwen3.5-4B} evaluator &
Extract diagnostic entities from prediction and reference explanations, judge bidirectional semantic coverage, and compute the harmonic mean of entity precision and recall. \\
\midrule
EvidenceScore &
\texttt{Qwen/Qwen3.5-4B} evaluator &
Extract entity-linked visual evidence claims from prediction and reference explanations, judge bidirectional semantic coverage, and compute the harmonic mean of evidence precision and recall. \\
\midrule
Normalized SLE &
SLE scorer &
Clip raw SLE to $[-1,4]$ and normalize using $(\mathrm{SLE}_{\mathrm{clip}}+1)/5$. \\
\midrule
Detection Accuracy / F1 &
Label parser and scorer &
Parse real/fake labels from model outputs and compute standard classification metrics. \\
\bottomrule
\end{tabularx}
\end{table}

\section{Baseline and Training Details}
\label{app:baseline_details}

\paragraph{Evaluation protocol.}
All baselines follow the same detection-and-explanation protocol. For each input image, the model receives a single image and is prompted to output a binary prediction, either \emph{real} or \emph{fake}, together with a natural-language explanation. We evaluate two explanation tracks separately. The \emph{complex} track asks for detailed forensic reasoning, while the \emph{simple} track asks for short, jargon-free explanations. Both tracks are evaluated on the same image lists, but use separate prompts tailored to the requested explanation style.

We evaluate on both in-distribution (ID) and out-of-distribution (OOD) test sets. ID fake images are sampled from generator families represented during training, while OOD fake images are sampled from held-out generators introduced only at evaluation time. Each evaluation set contains 18,500 images, consisting of 10,000 fake images and 8,500 real images. The same ID and OOD splits are used for both explanation tracks, ensuring that detection and explanation metrics are computed on aligned examples.

\paragraph{Prompt templates.}
Both zero-shot and fine-tuned baselines use the same task instruction and output schema, with separate prompts for the complex and simple explanation tracks. For the complex track, we use:
\begin{quote}
\small\ttfamily
Detect whether the image is real or fake and provide reasoning for it.\\[0.4em]
Respond in the following format:\\
\textless reasoning\textgreater your reasoning here\textless/reasoning\textgreater\\
\textless answer\textgreater real or fake\textless/answer\textgreater
\end{quote}

For the simple track, we use:
\begin{quote}
\small\ttfamily
Detect whether the image is real or fake and provide reasoning for it in simple, easy-to-understand, jargon-free language.\\[0.4em]
Respond in the following format:\\
\textless reasoning\textgreater your reasoning here\textless/reasoning\textgreater\\
\textless answer\textgreater real or fake\textless/answer\textgreater
\end{quote}

\paragraph{Zero-shot protocol.}
Zero-shot baselines are evaluated directly using the prompt templates above. The text inside \texttt{\textless reasoning\textgreater} is used for explanation evaluation, while the label inside \texttt{\textless answer\textgreater} is used for detection evaluation.

\paragraph{Fine-tuning setup.}
For fine-tuned baselines, we use a balanced 50\% subset of the training split, sampled evenly across real and fake images. We construct separate JSONL training files for the complex and simple explanation tracks using the same prompt templates and output schema as in zero-shot evaluation. The two files use the same sample identifiers and image paths, but differ in the target explanation style. Each track is fine-tuned independently so that the model is optimized for the corresponding explanation format.

\begin{table}[!htbp]
\centering
\caption{Key training, inference, and compute settings for fine-tuned VLM baselines.}
\label{tab:app_baseline_settings}
\footnotesize
\setlength{\tabcolsep}{4pt}
\renewcommand{\arraystretch}{1.06}
\begin{tabularx}{\linewidth}{p{0.38\linewidth}Y}
\toprule
\textbf{Setting} & \textbf{Value} \\
\midrule
\multicolumn{2}{l}{\textbf{Fine-tuning}} \\
\midrule
Framework & \texttt{ms-swift} supervised fine-tuning \\
Hardware & 4 AMD MI210 GPUs, each with 64 GB VRAM \\
Fine-tuning method & LoRA \\
Fine-tuning tracks & Complex and simple tracks trained separately \\
Training subset & Balanced 50\% subset of the training split \\
LoRA rank / alpha & 16 / 32 \\
Target modules & \texttt{all-linear} \\
Vision encoder & Frozen \\
Epochs & 1 \\
Per-device batch size & 8 \\
Gradient accumulation & 2 steps \\
Effective batch size & 64 \\
Learning rate & \(2\times10^{-4}\) \\
Learning-rate schedule & Cosine decay with warmup ratio 0.05 \\
Maximum sequence length & 2048 tokens \\
Precision & bfloat16 \\
Optimizer & AdamW, with \(\beta_1=0.9\), \(\beta_2=0.95\), and weight decay 0.1 \\
Maximum gradient norm & 1.0 \\
Runtime / compute per fine-tuning run
& 4B model: 6--7 h, 24--28 GPU-h; 7B/8B-class models: 9--10 h, 36--40 GPU-h; 14B model: \(\approx\)13 h, \(\approx\)52 GPU-h. Each run uses 4 GPUs. \\
Total fine-tuning compute
& 10 runs: 5 baselines \(\times\) 2 explanation tracks; approximately 368--400 GPU-h, assuming BusterX++ follows the 7B/8B-class runtime. \\
\midrule
\multicolumn{2}{l}{\textbf{Inference}} \\
\midrule
Inference backend & vLLM through \texttt{ms-swift} \\
Hardware & 1 AMD MI210 GPU \\
Maximum new tokens & 2048 \\
Decoding & Greedy decoding with temperature 0.0 \\
Precision & bfloat16 \\
Maximum model length & 4096 tokens \\
Maximum batch size & 16 \\
Runtime / compute per inference run
& 3--4 h per model-track evaluation on 1 GPU, corresponding to 3--4 GPU-h. \\
Total inference compute
& 20 open-source inference runs: 5 zero-shot baselines and 5 fine-tuned baselines, each evaluated on 2 explanation tracks; approximately 60--80 GPU-h. \\
\midrule
\multicolumn{2}{l}{\textbf{Automatic scoring}} \\
\midrule
Metric computation & EntityScore, EvidenceScore, BERTScore-F1, and SLE \\
Scoring backend & Offline Qwen3.5-4B evaluator for EntityScore/EvidenceScore; local metric computation for BERTScore-F1 and SLE \\
Runtime / compute per scoring run
& 3--4 h per scoring run on 1 GPU, corresponding to 3--4 GPU-h. \\
Total scoring compute
& Approximately 60--80 GPU-h across the corresponding open-source evaluation outputs. \\
\bottomrule
\end{tabularx}
\end{table}


\paragraph{Scoring.}
For each run, we parse the predicted label from the \texttt{<answer>} field and the explanation from the \texttt{<reasoning>} field. Detection is evaluated using the predicted real/fake label, while explanation metrics are computed separately for the complex and simple tracks using the corresponding generated explanation.

\paragraph{Compute accounting.}
GPU-hours are computed as the number of GPUs multiplied by the wall-clock runtime. The values in Table~\ref{tab:app_baseline_settings} report approximate compute for the final runs used in the paper and exclude small debugging runs. Dataset construction also required caption generation, edit-instruction generation, candidate image generation, filtering, explanation generation, and perturbation generation. For closed-source API models used during dataset construction and evaluation, provider-side compute is not observable; we therefore report processed-example counts/API requests in the project code and dataset documentation rather than provider-side GPU-hours.

\FloatBarrier
\section{Prompt Templates and Model Inputs}
\label{app:prompts}
This final section records the prompts used for captioning, edit-instruction generation, instruction-difficulty rating, visual-difference extraction, alignment judging, explanation generation, and metric judging. We place the prompt text at the end so that the main appendix reads continuously.

\begin{table}[!htbp]
\centering
\caption{Prompt and model-version index.}
\label{tab:app_prompt_versions}
\small
\setlength{\tabcolsep}{5pt}
\renewcommand{\arraystretch}{1.12}
\begin{tabularx}{\linewidth}{@{}p{0.25\linewidth}p{0.27\linewidth}Y@{}}
\toprule
\textbf{Stage} & \textbf{Model(s)} & \textbf{Output} \\
\midrule
Caption generation & GPT-4o-mini family & Dense image caption \\
Edit-instruction generation & GPT-4o-mini family & Structured edit-instruction JSON \\
Instruction difficulty audit & GPT-4o-mini family & Difficulty rating and rationale \\
Visual-difference extraction & Gemini-3-Flash, GPT-5-mini, GLM-4.6V, Qwen3.5-27B & Difference JSON with changed entities, locations, and evidence \\
Alignment judging & DeepSeek-V3.2, GPT-OSS-120B & \texttt{KEEP}, \texttt{BORDERLINE}, or \texttt{DISCARD} plus rationale \\
Complex explanation & Gemini-3-Flash, GPT-5-mini, GLM-4.6V, Qwen3.5-27B & Expert-facing explanation \\
Simple explanation & Gemini-3-Flash, GPT-5-mini, GLM-4.6V, Qwen3.5-27B & User-facing explanation \\
Authenticity explanation & Gemini-3-Flash, GPT-5-mini, GLM-4.6V, Qwen3.5-27B, Gemini-2.5-Pro & Real-image authenticity explanation \\
Metric extraction / judging & Qwen/Qwen3.5-4B & Entity/Evidence extraction and semantic coverage decisions \\
\bottomrule
\end{tabularx}
\renewcommand{\arraystretch}{1.0}
\end{table}

\FloatBarrier
\input{appendix_prompts}

%% file: appendix_prompts.tex

\subsection{Caption Generation Prompt}

\begin{xvpromptbox}{User prompt}
You are a professional vision-language annotator. Review the photo and produce a concise, information-dense caption (2–3 sentences) that:
- Clearly names the primary subject(s) as well as important secondary objects that are visible.
- Describes appearance details such as colors, materials, and notable textures, along with approximate positions (foreground/background/left/right).
- Mentions the environment or setting, lighting conditions, and any visible interactions or relationships between objects or people.
- Avoids speculation about unseen details, emotions, or origins, and never introduces objects that do not appear in the image.
- Uses straightforward, photographic language suitable for a professional photo-editing workflow.
Output only the caption text.
\end{xvpromptbox}

\subsection{Edit-Instruction Generation Prompt}

\begin{xvpromptbox}{System prompt}
You are an expert photo-editing director. Given a caption of a real photograph, you must propose concrete edit instructions that change visible content in the image and would be considered a manipulated/deepfake-like edit.

Key constraints:
- Avoid edits that are ONLY global/camera transforms (e.g., rotating/resizing/warping the entire image). Edits must modify scene content or localized geometry/spatial relationships.
- You MAY include people in the scene; however, at least ONE edit must target a NON-person object/region (environment, objects, signage, screens, etc.).
- Keep the result photorealistic (consistent lighting/materials/shadows). You MAY introduce “geometrically impossible” object shapes (e.g., square wheel, unnaturally thin object) as long as it still looks like a realistically manipulated photo (not fantasy/magic).
- Prefer 2-4 edits total, each unambiguous and easy to execute.
\end{xvpromptbox}

\begin{xvpromptbox}{Few-shot messages}
Example 1 user message:
Caption: A red vintage bicycle leans against a brick wall beside a green wooden door, while a potted fern rests on the stone steps.

Example 1 assistant message:
{
  "edits": [
    {
      "instruction": "Add a small wicker basket to the front handlebars of the red bicycle.",
      "target_objects": [
        "red bicycle"
      ]
    },
    {
      "instruction": "Repaint the green wooden door to a deep navy blue color.",
      "target_objects": [
        "green wooden door"
      ]
    },
    {
      "instruction": "Remove the potted fern from the stone steps and realistically fill in the exposed stone step surface.",
      "target_objects": [
        "potted fern",
        "stone steps"
      ]
    }
  ]
}

Example 2 user message:
Caption: Three coworkers gather around a glass conference table reviewing a printed sales chart, while a whiteboard behind them displays quarterly figures in blue marker.

Example 2 assistant message:
{
  "edits": [
    {
      "instruction": "Add a fourth team member standing on the left side of the glass conference table holding a tablet.",
      "target_objects": [
        "glass conference table"
      ]
    },
    {
      "instruction": "Change the sales chart on the table to display the title 'Q4 Forecast' instead of the current heading.",
      "target_objects": [
        "printed sales chart"
      ]
    },
    {
      "instruction": "Erase the blue marker text on the whiteboard and redraw it in green marker showing a 15
      "target_objects": [
        "whiteboard"
      ]
    }
  ]
}

Example 3 user message:
Caption: A skateboard rests on a concrete sidewalk next to a metal drain grate, with a paper coffee cup nearby.

Example 3 assistant message:
{
  "edits": [
    {
      "instruction": "Change the skateboard’s front wheels so they appear square-shaped instead of round, keeping the wheel material, dirt, and lighting consistent.",
      "target_objects": [
        "skateboard",
        "front wheels"
      ]
    },
    {
      "instruction": "Duplicate the paper coffee cup using a copy-move style edit, place the duplicate closer to the drain grate, and slightly rotate it so it looks naturally dropped; inpaint the original source area realistically.",
      "target_objects": [
        "paper coffee cup",
        "concrete sidewalk",
        "drain grate"
      ]
    }
  ]
}

Example 4 user message:
Caption: A storefront window has a large poster taped inside the glass, with a small sticker on the lower corner and brick wall around the frame.

Example 4 assistant message:
{
  "edits": [
    {
      "instruction": "Apply a local planar perspective warp to the large poster so it appears slightly tilted as if viewed more from the right side, without warping the entire window or bricks.",
      "target_objects": [
        "large poster"
      ]
    },
    {
      "instruction": "Use the newly revealed strip created by the tilt to reveal a faded graffiti on the glass, and blend it realistically.",
      "target_objects": [
        "newly revealed area behind the poster",
        "storefront window glass"
      ]
    },
    {
      "instruction": "Adjust tape edges/shadows so the poster still looks taped to the glass after tilting, and keep reflections/materials consistent.",
      "target_objects": [
        "tape",
        "large poster",
        "storefront window glass"
      ]
    }
  ]
}
\end{xvpromptbox}

\begin{xvpromptbox}{Final user prompt template}
Using the caption below, propose 2-4 concrete editing operations that intentionally change visible content in a deepfake-like way. Follow the rules:

- Reference the same object names or regions mentioned in the caption so the editor knows exactly what to modify.
- Avoid edits that are ONLY global/camera transforms (no whole-image rotate/resize/crop/perspective warp).
- At least ONE edit must target a NON-person object/region (objects, environment, signage, screens, printed text, etc.).

Allowed edits include:
  * add/remove/replace an object (and realistically inpaint the exposed background)
  * reposition/resize a specific object (localized, not whole-image)
  * copy-move duplication of an object/region, optionally with small rotate/scale/shear to blend
  * object shape deformation / “impossible geometry” (square wheel, unnaturally thin object, bent rigid object) while staying photorealistic
  * occlusion/depth-order change: make object A appear in front of/behind object B, generating newly revealed (disoccluded) areas realistically
  * local planar perspective warp on a planar object (poster/sign/screen/paper).
    - If the warp creates newly revealed (disoccluded) areas, DO NOT just “invent random stuff”.
      Instead, specify ONE of:
        (a) extend the same surface’s design/texture into the newly visible region (e.g., continue poster artwork), OR
        (b) reveal a physically plausible previously-hidden detail behind/under it (e.g., older poster layer, graffiti, tape residue, hidden label, crack), OR
        (c) reveal a object that could realistically have been occluded (e.g., a sticker under the poster edge), keeping it consistent with lighting and contact/occlusion.

- Keep edits photorealistic (materials/lighting/shadows consistent).
- Avoid fantasy/supernatural elements and generic enhancements (like “improve lighting”).
- Return only 2-4 edits.

Format the answer as JSON with this structure (no markdown, no explanation):
{{"edits": [
  {{"instruction": "<clear edit action>", "target_objects": ["<object name>", ...]}},
  ... up to four entries total ...
]}}

Caption:
{caption}
\end{xvpromptbox}

\subsection{Instruction Difficulty Rating Prompt}

\begin{xvpromptbox}{System prompt}
You are an edit-instruction difficulty rater for routing edits to image-editing models.

You will receive:
- A caption describing a real image (you do NOT see the image)
- A list of edit instructions

Your job:
For each instruction, estimate the probability that a typical image-editing model (given the real image + the instruction) will successfully achieve the intended change WITHOUT breaking other parts of the image that should remain unchanged.
Convert that probability into a difficulty score from 1 to 5.

CRITICAL PRINCIPLE:
Difficulty is driven primarily by how *constraint-tight and brittle* the instruction is, especially for *subtle/surgical* changes.
Large “generative” changes are often easier when they allow freedom and do not require matching the exact original instance details.

BASELINE ASSUMPTION:
Assume all edits are intended to remain photorealistic unless explicitly stated otherwise.
Do NOT increase difficulty just because the instruction says “realistic”, “match lighting”, or “preserve texture”.
Only increase difficulty for constraints beyond baseline photorealism.

Score = success probability band:
1 = Very easy  (P(success) > 0.85)
2 = Easy       (0.70–0.85)
3 = Moderate   (0.50–0.70)
4 = Hard       (0.30–0.50)
5 = Very hard  (P(success) < 0.30)

How to estimate P(success): weigh factors in this order (most important first)

1) Subtlety / under-edit risk (MOST IMPORTANT)
- Is the intended change small, fine-grained, or easy for models to ignore/half-apply?
- If yes, reduce P(success) aggressively, even if the edit sounds “small”.

2) Instance/identity lock (replacement not allowed)
- Must remain the same subject/object instance (not swapped for a different-looking one)?
- Strongly reduce P(success) if identity must remain stable while only one attribute changes.

3) Locality + collateral-damage intolerance
- Must be confined to a specific region while everything else stays identical?
- Higher intolerance to spillover => lower P(success).

4) Structure continuity / reveal-cover requirements
- Does the edit require plausible reconstruction of previously occluded areas or seamless continuation of edges/textures/geometry?
- If yes, lower P(success), especially when the reconstructed region must align with existing structures.

5) Caption underspecification
- If the caption doesn’t provide enough detail to judge what “correct” means (placement/appearance), treat the instruction as harder (lower P(success)).

Magnitude rule (secondary):
- Do NOT automatically rate large changes as hard. If the instruction permits freedom (any plausible result acceptable), it can have higher P(success) than a subtle, brittle edit.

Bundling rule:
- If an instruction bundles multiple required outcomes, rate it by the hardest required outcome.
- If missing any sub-part fails the instruction, prefer the higher difficulty.

Output MUST be valid JSON only (no markdown, no extra text), exactly:
{
  "difficulty_scores": [
    { "instruction": "<exactly copy instruction text>", "difficulty_1_to_5": 1-5, "reason": "<one short sentence>" }
  ]
}
\end{xvpromptbox}

\begin{xvpromptbox}{User prompt template}
Rate each edit instruction independently.

CAPTION:
{caption}

EDIT INSTRUCTIONS:
1) {instruction_1}
2) {instruction_2}
...
\end{xvpromptbox}

\subsection{Visual-Difference Extraction Prompt}

\begin{xvpromptbox}{System prompt}
You are a Visual Difference Extractor.

INPUTS:
- Image A: original/real image
- Image B: generated/edited image

GOAL:
List only the differences that are VISUALLY VERIFIABLE between Image A and Image B.
Differences may involve people, objects, text, surfaces, local geometry, spatial relationships, or scene structure.

OUTPUT (STRICT):
Return ONLY a valid JSON object (no markdown, no code fences, no extra text).
The JSON MUST have exactly these two keys:
1) "overall_summary": string
2) "changes": array of objects

Each object in "changes" MUST have exactly:
- "entity": string
- "salience": integer from 1 to 5
- "differences": array of strings

FORMAT for each string in "differences":
"A: <what is visible in Image A>; B: <what is visible in Image B>."

HOW TO COMPARE (be thorough):
- Compare Image A and Image B systematically by region: foreground, midground, background, and text-bearing regions.
- For each candidate changed region, compare not only object identity, but also:
  • presence/absence
  • count
  • position
  • orientation
  • size
  • outline / contour / silhouette
  • local shape
  • surface continuity
  • boundary alignment
  • overlap with nearby objects
  • support/contact with nearby surfaces
  • depth-order relationship
  • local perspective on flat or structured surfaces
  • text/content if legible

PAY SPECIAL ATTENTION TO SUBTLE BUT REAL DIFFERENCES SUCH AS:
- localized shape deformation rather than full object replacement
- bends, bulges, compressions, stretching, warping, or unnatural straightening/curving
- changes in edge flow, contour consistency, or region symmetry
- duplicated, missing, or newly introduced localized elements
- local repositioning of one object/part while the rest of the scene remains the same
- changes in which object overlaps, blocks, or reveals another
- changes in whether an object appears attached to, resting on, behind, or in front of another object
- planar or perspective inconsistencies on signs, screens, papers, posters, labels, packaging, walls, floors, tables, or other structured surfaces
- changed text, numbers, symbols, or logos when legible
- changes to pose, hands, face, gaze, clothing, or accessories only if clearly visible

IMPORTANT COMPARISON BEHAVIOR:
- When a region looks like the same object in both images, first ask whether its SHAPE, POSITION, RELATION, or STRUCTURE changed before describing it as a different object or a different appearance.
- Prefer describing a localized structural or spatial difference over inventing new semantic detail.
- If the main change is geometric, relational, or positional, report that directly.
- Distinguish between:
  • object identity change
  • object attribute change
  • local geometry change
  • spatial relationship change
  • text/content change
- Treat texture, pattern, material, and fine detail conservatively: only mention them if they are clearly visible in both images and are central to the difference.

STRICT VERIFIABILITY RULES:
- Do NOT include any statement unless you are confident it is directly supported by both images.
- Do NOT guess. Do NOT infer intent, cause, or editing process.
- Do NOT replace a subtle visible change with a generic object description.
- Do NOT invent fine-grained semantic details that are not clearly visible.
- Do NOT use vague filler such as:
  "modified", "repositioned", "different shapes", "changed in arrangement", "appears altered"
  unless you name the specific object/region and describe the concrete before/after difference.
- Do NOT invent changes for clutter or ambiguous background details.
- Only report background changes if they are specific and clearly visible.
- For subtle pose/expression changes: include only if the relevant body parts are clearly visible.
- For text: quote it only if legible.
- If a detail is small, blurry, occluded, or ambiguous, leave it out.

GROUPING RULES:
- Group differences under one entity only if they belong to the same localized object or region.
- Separate unrelated changes into separate entities.
- Prefer concrete entity names over vague labels.

SALience GUIDELINES:
- 5 = dominant change that strongly affects the image
- 4 = large obvious local change
- 3 = clearly visible secondary change
- 2 = subtle but reliable change
- 1 = minor but still verifiable detail

WRITING STYLE:
- Be concrete, visual, and concise.
- Describe the exact visible before/after difference.
- Name the object or region explicitly.
- Prefer structural, positional, and relational descriptions when those are the true source of the difference.
- Avoid generic recognition-style descriptions that do not identify the actual change.

- If you find no clear differences, output "changes": [] and say so in "overall_summary".
\end{xvpromptbox}

\begin{xvpromptbox}{User prompt}
Compare Image A (original) and Image B (generated). Produce the JSON response exactly as specified.
\end{xvpromptbox}

\subsection{Edit-Instruction Alignment Judge Prompt}

\begin{xvpromptbox}{System prompt}
ROLE: TEXT-ONLY JUDGE (no images)

SCENARIO:
Two versions of the same scene exist:
- Image A = before/original
- Image B = after/edited

You do NOT see images. You only see:
(1) EDIT_INSTRUCTION_JSON: the requested edit that Image B is supposed to contain relative to Image A.
(2) DIFFERENCE_JSON: another model's claimed A→B differences (may be lossy, incomplete, noisy, over-split, or somewhat imprecise in how it describes a local change).

IMPORTANT:
- Treat "A:" and "B:" only as labels inside the provided evidence lines.
- Do not assume anything not written.
- DIFFERENCE_JSON may contain false positives, missed details, partial descriptions, or slightly imperfect wording for the same underlying local edit.
- Judge semantic consistency carefully, but do not demand exact wording overlap.
- Prefer localized, concrete evidence over generic statements.

SEMANTIC COMPARISON RULE:
- Do NOT compare text literally or word-for-word.
- Judge meaning at the level of visual gist, localized scene content, and coarse attribute family.
- Treat near-synonyms, paraphrases, level-of-detail differences, and approximate descriptor wording as potentially referring to the same visual change.
- Small wording drift in color, material, texture, shape, position, or relation should NOT by itself count as a mismatch, contradiction, or unrelated change if the overall localized meaning is still consistent.
- Prefer semantic equivalence over exact lexical overlap.
- Only treat two descriptions as truly different if they imply a materially different visible change in the scene.

GOAL:
Decide whether the claimed A→B differences are consistent with the requested edit.

-------------------------
STEP 0: Determine target scope
From EDIT_INSTRUCTION_JSON:
- Identify the requested edit action.
- Identify the listed target_objects.
- Determine the LOCAL EDIT SCOPE.

IN-SCOPE includes:
- direct changes to the named target object(s)/region(s)
- direct local consequences of the edit on the same localized area, such as:
  - changed overlap/contact between the target and nearby objects
  - newly visible or newly occluded local regions caused by the target edit
  - local geometry/perspective changes on the same surface
  - local text/content changes on the edited object
  - local color/material/pattern/appearance changes if they plausibly belong to the requested edit
  - nearby local relational changes that naturally follow from editing the target

OUT-OF-SCOPE includes:
- unrelated changes to other objects, people, text, or background regions not plausibly tied to the requested edit
- broad scene rewrites beyond the local target area
- unrelated semantic additions/removals elsewhere in the scene

Important:
- Scope should be based on the requested edit as a localized manipulation, not the whole image.
- Do NOT treat every nearby difference as off-target if it is a natural local consequence of the intended edit.

-------------------------
STEP 1: Build a single CORE EDIT INTENT (do NOT over-split)
Create ONE sentence capturing the intended visual change from EDIT_INSTRUCTION_JSON.

Then extract 2–6 SUPPORTING CUES from the instruction, but treat them as:
- illustrative hints for what evidence might look like,
- NOT a checklist,
- NOT individually required,
- and not to be scored by counting matches.

Supporting cues may include:
- edited object identity
- add/remove/replace action
- before/after content
- relative location or relation
- text content change
- color/material/pattern/appearance change
- shape/deformation/structure change
- local perspective/warp change
- depth-order/occlusion change
- duplication/count change
- localized pose/part change
- local interaction/contact change

IMPORTANT MATCHING PRINCIPLE:
- The cues describe ONE intended edit; cues often overlap.
- If the core intent is supported, missing fine-grained cues are allowed.
- Merge/normalize cues that express the same underlying change.
- Judge semantic consistency, not wording overlap.
- A partially correct or somewhat imprecise description may still support the core intent if it is clearly about the same localized edit.
- Normalize wording into coarse visual meaning before judging.
- Collapse superficially different phrases that likely describe the same localized appearance or change.
- Differences in specificity are allowed: a broader or slightly narrower description may still match if it points to the same local edit.
- Do not require exact agreement on low-level descriptive words when the underlying visual change is the same.

-------------------------
STEP 2: Evidence matching (semantic, not word-for-word)
Search DIFFERENCE_JSON for evidence lines.

Classify evidence strength as:

- STRONG:
  concrete before/after about a named object/region and its visible change;
  explicit add/remove/replace;
  explicit text/content change;
  explicit position/overlap/depth change;
  explicit color/material/shape/structure change;
  or a clearly localized statement that directly supports the core intent.

- MODERATE:
  evidence is localized to the correct object/region and plausibly supports the intended edit,
  but is incomplete, somewhat imprecise, less specific than the instruction, or does not fully spell out the before/after.
  This includes semantically adjacent descriptions that likely refer to the same local change.

- WEAK:
  vague claims without a clear object/region anchor, such as "looks different", "appears altered", "changed layout",
  "more detailed", or similarly generic statements that do not clearly identify the edited content.

Decide CORE_MATCH as one of:
- MATCHED_STRONG: strong evidence directly supports the core intent.
- MATCHED_IMPLIED: moderate-to-strong evidence supports the same localized intended edit, even if the wording is incomplete or somewhat imprecise.
- PARTIAL_MATCH: some on-target support exists, but it is limited, underspecified, or ambiguous.
- NOT_MATCHED: insufficient evidence for the core intent.

Rules:
- STRONG evidence can support MATCHED_STRONG.
- STRONG or MODERATE evidence can support MATCHED_IMPLIED.
- MODERATE evidence, or a combination of weakly aligned but correctly localized evidence, can support PARTIAL_MATCH.
- WEAK evidence alone should not support MATCHED_STRONG or MATCHED_IMPLIED.
- Do not fail the sample just because one fine detail is absent, if the core edit is reasonably supported.
- If evidence supports the same localized edit but phrases it differently, count it as support.
- If evidence is somewhat imprecise but clearly refers to the same target area and same underlying edit family, allow partial or implied support.
- Compare the intended edit and the evidence after mentally normalizing paraphrases and approximate descriptors into the same coarse visual meaning.
- A wording difference is not a mismatch if both descriptions plausibly refer to the same localized visible change.
- Prefer agreement on target region + edit gist over exact agreement on attribute words.
- Do NOT treat small descriptor mismatches as contradictions if they are plausibly alternate descriptions of the same local appearance.
- If two descriptions differ mainly in wording granularity, specificity, or approximate attribute naming, prefer a partial/implied match rather than a failure.
- Contradiction requires a genuinely incompatible visual claim, not merely different wording.
- If the evidence points to the opposite of the requested change, treat that as contradiction.

-------------------------
STEP 3: Unrelated changes and hard-discard rule
Consider only OUT-OF-SCOPE changes with STRONG or clearly MODERATE evidence.
Ignore purely WEAK-only claims.

Important caution:
- DIFFERENCE_JSON may over-report tiny side details.
- Do NOT penalize local secondary effects that are natural consequences of the requested edit.
- Do NOT penalize slightly imperfect descriptions of the same local edit as separate off-target edits.
- Do NOT penalize wording drift that likely refers to the same underlying local edit.
- Do NOT create an off-target penalty from a merely imprecise attribute description.
- Before penalizing, ask whether the claimed difference is truly a separate semantic edit, or just a noisy textual rendering of the same local change.
- Only penalize changes that are clearly separate semantic edits outside the local target scope.

Classify OUT-OF-SCOPE changes as:

A) COSMETIC/NO-OP (penalty 0):
   - low-salience distractors
   - minor style/layout differences with the same semantic content
   - small text imperfections or character-level noise plausibly due to imperfect rendering
   - tiny local inconsistencies that do not amount to a separate semantic edit
   - nearby secondary details plausibly caused by the intended local edit
   - approximate wording differences that preserve the same coarse visual meaning
   - minor descriptor drift in color/material/pattern/shape/position wording that does not imply a separate semantic edit
   - alternate phrasings of the same localized change

B) MODERATE OFF-TARGET (penalty 1):
   - clear semantic change outside scope that is noticeable but not dominant
   - a separate added/removed/replaced object outside the target area
   - a noticeable unrelated text/content change when text is not the intended target
   - one clear off-target change elsewhere in the scene

C) PROMINENT OFF-TARGET / REWRITE (penalty 2 → hard discard):
   - major unrelated object/person addition, removal, or replacement
   - dominant unrelated scene rewrite
   - multiple strong off-target semantic changes
   - a clear separate edit affecting one of the most noticeable regions of the image
   - a dominant unrelated text/content change central to the scene

Compute unrelated_change_penalty:
- 2 if any C exists
- else 1 if any B exists (or multiple moderate off-target changes exist)
- else 0

HARD DISCARD:
- If unrelated_change_penalty == 2, verdict MUST be DISCARD.

SEMANTIC PENALTY GATE:
Before adding any penalized_unrelated_change, verify that it describes a genuinely separate visible edit outside the intended local scope.
If it could reasonably be an imprecise, approximate, or differently-worded description of the intended local change, do NOT penalize it.

-------------------------
STEP 4: Scoring and verdict
If unrelated_change_penalty == 2:
  final_score = 0..2 (choose based on how well core intent matches), verdict = DISCARD.
Else:
  Assign match_score 0..5 based on CORE_MATCH (do NOT score by counting matched cues):

  5 = MATCHED_STRONG
      core intent directly supported by strong, localized evidence; no contradiction

  4 = MATCHED_IMPLIED
      core intent supported by strong/moderate evidence that is clearly on-target and semantically consistent,
      even if not perfectly specific

  3 = PARTIAL_MATCH
      core intent has meaningful on-target support, but evidence is incomplete, underspecified, or somewhat ambiguous

  2 = weak/minimal overlap with intent
      some limited alignment exists, but support is too vague, too partial, or too uncertain

  1 = almost no support

  0 = contradiction
      evidence suggests the opposite of the requested edit, or clearly mismatches the target/edit family

  final_score = clamp(match_score - unrelated_change_penalty, 0..5)

Verdict:
- KEEP if final_score >= 4
- BORDERLINE if final_score == 3
- DISCARD otherwise

-------------------------
GUIDANCE FOR matched_requirements / missing_requirements
- matched_requirements:
  list the main parts of the core intent that are supported, at the level of meaning, not exact wording.
- missing_requirements:
  list only important missing parts of the core intent.
  Do NOT list every tiny omitted detail.
  Do NOT mark something missing only because the wording differs if the semantic gist is already covered.
  If the core intent is broadly supported, this list may be short or empty.

-------------------------
OUTPUT (STRICT JSON ONLY)
Return ONLY a JSON object with exactly these keys:
- "final_score": integer 0..5
- "match_score": integer 0..5
- "unrelated_change_penalty": integer 0..2
- "verdict": "KEEP" | "BORDERLINE" | "DISCARD"
- "matched_requirements": array of strings
- "missing_requirements": array of strings
- "penalized_unrelated_changes": array of strings
- "evidence_used": array of strings
\end{xvpromptbox}

\subsection{Complex Fake-Explanation Prompt}

\begin{xvpromptbox}{System prompt}
ROLE: Authenticity Critic: Generated-Image-Only Explanation

You will be given:
- GENERATED_IMAGE (this is the ONLY thing you may describe in the final explanation)
- AUXILIARY REFERENCES (for private guidance only): REAL_IMAGE, EDIT_INSTRUCTION_JSON, DIFFERENCE_JSON

Purpose of auxiliary references:
- REAL_IMAGE: a reliable reference of the original scene.
- EDIT_INSTRUCTION_JSON: the intended semantic change (the generated image may follow it partially).
- DIFFERENCE_JSON: a VLM-produced candidate list of changes (helpful as a checklist, but not authoritative).

TASK:
Write a natural, human-style explanation of why the GENERATED_IMAGE looks fake/synthetic/manipulated,
as if you ONLY saw the generated image.

NON-NEGOTIABLE RULES (no leakage):
1) Do NOT mention or imply: any other image, comparisons, “before/after”, “edit instruction”, “difference JSON”,
   “Image A/B”, “original/real/reference”, or “compared to”.
2) Every claim must be directly and clearly supported by visible evidence in the GENERATED_IMAGE alone.
   - If you only know it because of the auxiliary references, DO NOT include it.
3) You may use the auxiliary references ONLY to decide where to inspect more carefully.
   - Do NOT copy wording from DIFFERENCE_JSON.
   - Do NOT introduce claims that you cannot see in the GENERATED_IMAGE.

WHAT TO LOOK FOR (examples):
- warped geometry (hands/face asymmetry, bent straight lines)
- texture issues (plastic skin, smeared detail, patchy/over-smoothed areas)
- edge/blending artifacts (halos, cutout edges, seams)
- lighting/shadow inconsistencies (wrong direction, missing contact shadow)
- repeated/tiling patterns or unnatural background
- text artifacts (garbled letters, inconsistent strokes)
- inconsistent occlusion/reflections or depth cues
- noise/sharpness mismatch (one region too sharp/too blurry)

INTERNAL PROCESS:
- First inspect the GENERATED_IMAGE and form your own observations.
- Then consult the auxiliary references only as a checklist for regions to re-check.
- Keep only strong, visible cues.

OUTPUT FORMAT (STRICT):
Return ONLY valid JSON with exactly one key:
{
  "explanation": string
}

STYLE CONSTRAINTS:
- Mention concrete cues with lightweight location hints in-text (e.g., “around the face…”, “near the text…”, “in the background…”).
- No speculation about editing steps; stay visual.

FINAL SELF-CHECK:
- If any forbidden terms appear (real/original/reference/instruction/difference/compared/before/after/Image A/Image B),
  rewrite before answering.
\end{xvpromptbox}

\begin{xvpromptbox}{User prompt}
Generate the explanation JSON exactly as specified.
Remember: ONLY describe the GENERATED_IMAGE in the final explanation.
\end{xvpromptbox}

\subsection{Simple Fake-Explanation Prompt}

\begin{xvpromptbox}{System prompt}
You are a Digital Detective. Your specialty is spotting image manipulations and explaining them in a way a five-year-old can understand, without any technical jargon

Your job is to look at an edited image and explain, in a very simple way, why one part of it feels fake, strange, or out of place.

You will be given:
- GENERATED_IMAGE
- COMPLEX_EXPLANATION
- DIFFERENCE_JSON
- EDIT_INSTRUCTION_JSON

Your goal is to write one short explanation that even a five-year-old could understand.

IMPORTANT IDEA:
This simple explanation should NOT just be a shorter version of the complex explanation.
It can say something new.
But it should stay focused on the same main changed thing, person, object, or part of the scene suggested by the COMPLEX_EXPLANATION and/or DIFFERENCE_JSON.

You do NOT need to explain every changed thing. You can pick only one thing.

MASTER RULE:
Trust what you can see in the GENERATED_IMAGE most.

The other inputs are only clues:
- COMPLEX_EXPLANATION helps you know the main changed thing
- DIFFERENCE_JSON helps you know what likely changed most
- EDIT_INSTRUCTION_JSON helps you know what kind of change may have been wanted

But the final explanation must sound like it came from looking at the GENERATED_IMAGE only.

Use:
- very easy words
- short sentences
- everyday ideas
- one clear reason
- an analogy if it helps

Do NOT use hard or technical words.

DO NOT use words like:
- lighting
- reflection
- shadow
- smooth
- sharp
- texture
- blend
- artifact
- proportion
- occlusion
- illumination
- forensic
- manipulated region
- inconsistent
etc.

Instead, say things in simple ways, like:
- too shiny
- too blurry
- too stiff
- does not match
- looks pasted on
- looks like a toy
- looks like a sticker
- looks mixed up
- does not fit the moment
- does not belong there

Look for simple reasons like:
- a face that does not match the moment
- a body pose that looks odd
- hands or fingers that look wrong
- words that look mixed up
- clothes that do not fit the place
- something that looks pasted on
- something that looks like a cartoon in a real picture
- something that feels out of place in an easy, common-sense way

GOOD EXAMPLES:
- "A person is standing on a sandy beach, but their feet aren't sinking into the sand at all and they don't have any footprints behind them. It looks like they are floating on top of the sand, which doesn't happen in real life."
- "A man is wearing a big, thick winter coat and a wool hat, but he is standing on a sunny beach with palm trees. This doesn't make sense because people wear swimsuits at the beach, not winter clothes."
- "There is a fluffy, cartoon-style cat sitting in a field of real grass. The cat looks like it came from a TV show, while the grass looks like a real photograph. The two styles don't match."

BAD EXAMPLES:
- "The image contains lighting inconsistencies."
- "The face has blending artifacts."
- "The manipulated region is unrealistic."
- "This looks fake."
- "The subject is unnatural."

VERY IMPORTANT RULES:
1) Write the final explanation as if you only saw the GENERATED_IMAGE.
2) Do NOT mention or hint at:
   - the original image
   - a reference image
   - before/after
   - edit instruction
   - difference JSON
   - intended change
   - “A” or “B”
3) Do NOT use comparison phrases like:
   - changed from
   - used to
   - originally
   - compared to
   - should be
   - restored
4) Do NOT introduce a totally new main person, object, or story that is not connected to the COMPLEX_EXPLANATION and/or DIFFERENCE_JSON.
5) You MAY say new simple things about the same changed thing, as long as they are easy to see in the GENERATED_IMAGE.
6) If you are unsure, choose the clearest and most obvious reason only.
7) Do not try to sound smart. Sound clear.

OUTPUT FORMAT:
Return ONLY valid JSON with exactly one key:
{
  "explanation": string
}

FINAL CHECK BEFORE ANSWERING:
- Is this about just one main changed thing?
- Is this easy for a five-year-old to understand?
- Did I avoid hard words?
- Did I explain WHY it feels off, instead of just saying it is fake?
- Does it sound like I only looked at the GENERATED_IMAGE?

If any answer is no, rewrite it before returning.
\end{xvpromptbox}

\begin{xvpromptbox}{User prompt}
Generate the explanation JSON exactly as specified.

Remember:
- Focus on only one main changed thing.
- Pick the easiest and most obvious one to explain.
- Keep the explanation very simple.
- It should sound like something a normal person, or even a child, could understand.
- It may say something new, but only about the same main changed thing suggested by the COMPLEX_EXPLANATION and/or DIFFERENCE_JSON.
- Use the GENERATED_IMAGE as the main evidence.
- Do not mention hidden inputs or compare to another image.
\end{xvpromptbox}

\subsection{Real-Image Authenticity Prompt}

\begin{xvpromptbox}{System prompt}
You are an expert "Digital Detective." Your special skill is explaining why a picture is real (that is, not AI-manipulated), using super simple language that even a five-year-old can understand.
You will be given one input:
[Image]: A single, real-world photograph.
Your Mission:
Write a short, simple, and convincing explanation for why the image appears authentic. Your entire output must be a single JSON object.
Guiding Principles & Rules:
Your goal is to write a simple one or two-sentence paragraph that explains why the image feels like a real snapshot of a moment. You must point out a compelling, physical detail that a child could see and understand.
To build your case, look for real-world details that are hard to fake. Focus only on simple, physical things:
Little Messes and Imperfections: Is there a wrinkle in a shirt, a scuff on a shoe, a crumb on a table, or a little bit of dirt on someone's face? Real life isn't perfectly clean.
How Things Touch and Interact: Does a person's hair look messy because of the wind? Do their feet sink into the sand? Does a spoon make a dent in ice cream?
Real Textures and Patterns: Does the cat's fur look fuzzy and soft? Does the bark on the tree look rough and bumpy? Does a sweater look like it's really made of yarn?
The most important rule is to use zero grown-up words. Explain what you see in the simplest way possible.
Required Output Format:
Your entire output must be a single, valid JSON object with one key
{
  "authenticity_explanation": "Your simple and convincing explanation of why the image appears authentic."
}
Example of a High-Quality Output:
(Input: Image of a person writing in a notebook at a wooden desk)
{
  "authenticity_explanation": "This looks real because the wooden desk isn't perfect. You can see the little lines that all wood has, and there are even some tiny scratches on it, like a real desk that someone uses every day."
}
(Input: Image of a dog playing with a ball in a park)
{
  "authenticity_explanation": "You can tell this is a real dog because he's a little bit messy. There's some mud on his nose and his paws are dirty, which shows he was really outside playing in the park."
}
\end{xvpromptbox}

\begin{xvpromptbox}{User prompt}
Analyze the REAL_IMAGE and return the JSON exactly in the required format.
Only use visible evidence from the image.
\end{xvpromptbox}

\subsection{Metric Extraction and Coverage Prompts}

\begin{xvpromptbox}{Entity and fact extraction prompt}
You are an information extraction system.

Task:
Given an explanation of why an image looks fake or real, extract only:

1. diagnostic_entities:
   the entities that are actually used as evidence for the authenticity judgment

2. evidence_claims:
   the specific claims that explain why those entities make the image look fake or real

Important:
Do NOT extract all entities mentioned in the explanation.
Do NOT extract all factual statements.
Extract only the entities and claims that function as evidence for the fake/real judgment.

Definitions:

Diagnostic entity:
A person, object, text element, clothing element, body region, or scene component that is explicitly used as evidence for why the image looks fake or real.

Evidence claim:
A compact, checkable claim about a diagnostic entity that directly supports the authenticity judgment.

What to exclude:
- entities mentioned only for context
- descriptive facts that do not support the fake/real reasoning
- generic conclusions such as "the image looks fake" or "the image looks real"
- identity, location, color, or presence details unless they are themselves part of the evidence

Entity selection rule:
Include an entity only if the explanation uses it as evidence for authenticity or manipulation.

Claim selection rule:
Include a claim only if removing it would weaken the explanation for why the image looks fake or real.

Granularity policy:
- Prefer coarse, meaningful entities
- Prefer compact, meaningful claims
- Do not over-split one natural observation into many tiny claims
- But do split clearly distinct pieces of evidence into separate claims

Examples of the intended behavior:
- If a person is mentioned only to identify another evidence-bearing item, do not include the person as an entity
- If a phrase is only contextual and not part of the authenticity reasoning, do not extract it as a claim
- If an entity has multiple distinct evidence-based issues, extract multiple claims for that entity
- If multiple descriptors belong to one natural evidence unit, keep them together

Rules:
1. Use only information explicitly stated in the explanation
2. Do not infer unstated causes, intentions, or visual details
3. Extract only evidence-bearing entities
4. Extract only evidence-bearing claims
5. Each claim must be specific and checkable
6. Keep wording normalized and concise
7. Avoid redundancy
8. Do not create entities for minor sub-parts unless the explanation is clearly centered on them
9. Do not include non-evidential context
10. Return valid JSON only

Output schema:
{
  "diagnostic_entities": [
    {
      "entity_id": "E1",
      "name": "..."
    }
  ],
  "evidence_claims": [
    {
      "claim_id": "C1",
      "entity_id": "E1",
      "claim": "..."
    }
  ]
}

Additional guidance:
- Prefer the entity that actually carries the evidence
- Do not include broader context entities when the evidence is really about a more specific visual element
- Exclude claims whose role is only to identify or locate an entity
- Include only claims that explain why the image appears fake or real
- If a claim can be removed without weakening the fake/real reasoning, exclude it

Now process the following explanation:

{{EXPLANATION_TEXT}}
\end{xvpromptbox}

\begin{xvpromptbox}{Semantic coverage prompt}
You are an evaluation system.

Task:
You will be given:
1. a reference JSON containing:
   - diagnostic_entities
   - evidence_claims
2. a candidate explanation

Your job is to determine, for each diagnostic entity and each evidence claim in the reference JSON, whether it is present in the candidate explanation.

This is a LENIENT SEMANTIC MATCHING task.
Do NOT require exact wording.
Match based on core meaning.

What counts as a match:

Entity present:
A diagnostic entity is present if the candidate explanation clearly refers to the same evidence-bearing thing, either directly or indirectly, even if the wording is different.

Claim present:
An evidence claim is present if the candidate explanation conveys the same core evidence or substantially the same observation, even if:
- the wording is different
- the phrasing is less specific
- the claim is merged with another claim
- the claim is split across multiple sentences
- the candidate uses synonyms or paraphrases
- the candidate expresses the same idea more simply

Be lenient:
- Accept paraphrases
- Accept indirect references if clear enough
- Accept slightly broader or slightly less specific wording if the same core idea is preserved
- Accept one candidate sentence covering multiple reference claims
- Accept one reference claim being covered across multiple candidate sentences

But do NOT mark present if:
- the candidate only mentions the entity without the relevant evidence
- the candidate gives only generic criticism or praise without the actual reference idea
- the candidate discusses a related but meaningfully different issue
- the match is too vague or speculative

Decision rule:
Ask:
"Would a reasonable human judge say that the candidate explanation covers this same evidence, even if phrased differently?"
- If yes, mark present
- If only weakly related or too vague, mark false

Important:
Focus on evidence, not wording.
Do not invent support that is not actually present in the candidate explanation.

Output format:
Return valid JSON only.

Use this schema:
{
  "entity_matches": [
    {
      "entity_id": "E1",
      "entity_name": "...",
      "present": true,
      "matched_text": "...",
      "reason": "brief explanation"
    }
  ],
  "claim_matches": [
    {
      "claim_id": "C1",
      "entity_id": "E1",
      "reference_claim": "...",
      "present": true,
      "matched_text": "...",
      "reason": "brief explanation"
    }
  ]
}

Output requirements:
- `present` must be either true or false
- `matched_text` should be the shortest relevant span from the candidate explanation if present, otherwise ""
- `reason` should be short and based on semantic matching
- Evaluate each entity and claim independently
- The same candidate text may support multiple items
- If the candidate covers only a very generic version of the claim without the core evidence, mark false

REFERENCE_JSON:
{{REFERENCE_JSON}}

CANDIDATE_EXPLANATION:
{{CANDIDATE_EXPLANATION}}
\end{xvpromptbox}

%% file: xplainverse_references.bib
@article{kuznetsova2020openimages,
   title={The Open Images Dataset V4: Unified Image Classification, Object Detection, and Visual Relationship Detection at Scale},
   volume={128},
   ISSN={1573-1405},
   url={http://dx.doi.org/10.1007/s11263-020-01316-z},
   DOI={10.1007/s11263-020-01316-z},
   number={7},
   journal={International Journal of Computer Vision},
   publisher={Springer Science and Business Media LLC},
   author={Kuznetsova, Alina and Rom, Hassan and Alldrin, Neil and Uijlings, Jasper and Krasin, Ivan and Pont-Tuset, Jordi and Kamali, Shahab and Popov, Stefan and Malloci, Matteo and Kolesnikov, Alexander and Duerig, Tom and Ferrari, Vittorio},
   year={2020},
   month=Mar, pages={1956–1981} }

@misc{kuckreja2026pixels,
      title={Pixels Don't Lie (But Your Detector Might): Bootstrapping MLLM-as-a-Judge for Trustworthy Deepfake Detection and Reasoning Supervision}, 
      author={Kartik Kuckreja and Parul Gupta and Muhammad Haris Khan and Abhinav Dhall},
      year={2026},
      eprint={2602.19715},
      archivePrefix={arXiv},
      primaryClass={cs.CV},
      url={https://arxiv.org/abs/2602.19715}, 
}

@article{kosti2020emotic,
   title={Context Based Emotion Recognition using EMOTIC Dataset},
   ISSN={1939-3539},
   url={http://dx.doi.org/10.1109/TPAMI.2019.2916866},
   DOI={10.1109/tpami.2019.2916866},
   journal={IEEE Transactions on Pattern Analysis and Machine Intelligence},
   publisher={Institute of Electrical and Electronics Engineers (IEEE)},
   author={Kosti, Ronak and Alvarez, Jose and Recasens, Adria and Lapedriza, Agata},
   year={2019},
   pages={1–1} }

@misc{oh2015person,
      title={Person Recognition in Personal Photo Collections}, 
      author={Seong Joon Oh and Rodrigo Benenson and Mario Fritz and Bernt Schiele},
      year={2015},
      eprint={1509.03502},
      archivePrefix={arXiv},
      primaryClass={cs.CV},
      url={https://arxiv.org/abs/1509.03502}, 
}

@misc{li2020visualsocial,
      title={Visual Social Relationship Recognition}, 
      author={Junnan Li and Yongkang Wong and Qi Zhao and Mohan S. Kankanhalli},
      year={2018},
      eprint={1812.05917},
      archivePrefix={arXiv},
      primaryClass={cs.CV},
      url={https://arxiv.org/abs/1812.05917}, 
}

@misc{liu2022humancentric,
      title={Human-centric Relation Segmentation: Dataset and Solution}, 
      author={Si Liu and Zitian Wang and Yulu Gao and Lejian Ren and Yue Liao and Guanghui Ren and Bo Li and Shuicheng Yan},
      year={2021},
      eprint={2105.11168},
      archivePrefix={arXiv},
      primaryClass={cs.CV},
      url={https://arxiv.org/abs/2105.11168}, 
}

@misc{rossler2019faceforensicspp,
      title={FaceForensics++: Learning to Detect Manipulated Facial Images}, 
      author={Andreas Rössler and Davide Cozzolino and Luisa Verdoliva and Christian Riess and Justus Thies and Matthias Nießner},
      year={2019},
      eprint={1901.08971},
      archivePrefix={arXiv},
      primaryClass={cs.CV},
      url={https://arxiv.org/abs/1901.08971}, 
}

@misc{dolhansky2020dfdc,
      title={The DeepFake Detection Challenge (DFDC) Dataset}, 
      author={Brian Dolhansky and Joanna Bitton and Ben Pflaum and Jikuo Lu and Russ Howes and Menglin Wang and Cristian Canton Ferrer},
      year={2020},
      eprint={2006.07397},
      archivePrefix={arXiv},
      primaryClass={cs.CV},
      url={https://arxiv.org/abs/2006.07397}, 
}

@misc{dang2020dffd,
      title={On the Detection of Digital Face Manipulation}, 
      author={Hao Dang and Feng Liu and Joel Stehouwer and Xiaoming Liu and Anil Jain},
      year={2020},
      eprint={1910.01717},
      archivePrefix={arXiv},
      primaryClass={cs.CV},
      url={https://arxiv.org/abs/1910.01717}, 
}

@misc{zhu2023genimage,
      title={GenImage: A Million-Scale Benchmark for Detecting AI-Generated Image}, 
      author={Mingjian Zhu and Hanting Chen and Qiangyu Yan and Xudong Huang and Guanyu Lin and Wei Li and Zhijun Tu and Hailin Hu and Jie Hu and Yunhe Wang},
      year={2023},
      eprint={2306.08571},
      archivePrefix={arXiv},
      primaryClass={cs.CV},
      url={https://arxiv.org/abs/2306.08571}, 
}

@misc{wang2023dire,
      title={DIRE for Diffusion-Generated Image Detection}, 
      author={Zhendong Wang and Jianmin Bao and Wengang Zhou and Weilun Wang and Hezhen Hu and Hong Chen and Houqiang Li},
      year={2023},
      eprint={2303.09295},
      archivePrefix={arXiv},
      primaryClass={cs.CV},
      url={https://arxiv.org/abs/2303.09295}, 
}

@misc{pal2024semitruths,
      title={Semi-Truths: A Large-Scale Dataset of AI-Augmented Images for Evaluating Robustness of AI-Generated Image detectors}, 
      author={Anisha Pal and Julia Kruk and Mansi Phute and Manognya Bhattaram and Diyi Yang and Duen Horng Chau and Judy Hoffman},
      year={2024},
      eprint={2411.07472},
      archivePrefix={arXiv},
      primaryClass={cs.CV},
      url={https://arxiv.org/abs/2411.07472}, 
}

@misc{liu2021spsl,
      title={Spatial-Phase Shallow Learning: Rethinking Face Forgery Detection in Frequency Domain}, 
      author={Honggu Liu and Xiaodan Li and Wenbo Zhou and Yuefeng Chen and Yuan He and Hui Xue and Weiming Zhang and Nenghai Yu},
      year={2021},
      eprint={2103.01856},
      archivePrefix={arXiv},
      primaryClass={cs.CV},
      url={https://arxiv.org/abs/2103.01856}, 
}

@misc{li2020facexray,
      title={Face X-ray for More General Face Forgery Detection}, 
      author={Lingzhi Li and Jianmin Bao and Ting Zhang and Hao Yang and Dong Chen and Fang Wen and Baining Guo},
      year={2020},
      eprint={1912.13458},
      archivePrefix={arXiv},
      primaryClass={cs.CV},
      url={https://arxiv.org/abs/1912.13458}, 
}

@misc{li2018eye,
      title={In Ictu Oculi: Exposing AI Generated Fake Face Videos by Detecting Eye Blinking}, 
      author={Yuezun Li and Ming-Ching Chang and Siwei Lyu},
      year={2018},
      eprint={1806.02877},
      archivePrefix={arXiv},
      primaryClass={cs.CV},
      url={https://arxiv.org/abs/1806.02877}, 
}

@inproceedings{afchar2018mesonet,
   title={MesoNet: a Compact Facial Video Forgery Detection Network},
   url={http://dx.doi.org/10.1109/WIFS.2018.8630761},
   DOI={10.1109/wifs.2018.8630761},
   booktitle={2018 IEEE International Workshop on Information Forensics and Security (WIFS)},
   publisher={IEEE},
   author={Afchar, Darius and Nozick, Vincent and Yamagishi, Junichi and Echizen, Isao},
   year={2018},
   month=Dec, pages={1–7} 
}

@INPROCEEDINGS{guera2018deepfake,
  author={Güera, David and Delp, Edward J.},
  booktitle={2018 15th IEEE International Conference on Advanced Video and Signal Based Surveillance (AVSS)}, 
  title={Deepfake Video Detection Using Recurrent Neural Networks}, 
  year={2018},
  volume={},
  number={},
  pages={1-6},
  doi={10.1109/AVSS.2018.8639163}}

@misc{zhao2021multiattentional,
      title={Multi-attentional Deepfake Detection}, 
      author={Hanqing Zhao and Wenbo Zhou and Dongdong Chen and Tianyi Wei and Weiming Zhang and Nenghai Yu},
      year={2021},
      eprint={2103.02406},
      archivePrefix={arXiv},
      primaryClass={cs.CV},
      url={https://arxiv.org/abs/2103.02406}, 
}

@inproceedings{zhang2022patchdiffusion,
  title={Patch diffusion: a general module for face manipulation detection},
  author={Zhang, Baogen and Li, Sheng and Feng, Guorui and Qian, Zhenxing and Zhang, Xinpeng},
  booktitle={Proceedings of the AAAI conference on artificial intelligence},
  volume={36},
  number={3},
  pages={3243--3251},
  year={2022}
}

@misc{zhang2024commonsense,
      title={Common Sense Reasoning for Deepfake Detection}, 
      author={Yue Zhang and Ben Colman and Xiao Guo and Ali Shahriyari and Gaurav Bharaj},
      year={2024},
      eprint={2402.00126},
      archivePrefix={arXiv},
      primaryClass={cs.CV},
      url={https://arxiv.org/abs/2402.00126}, 
}

@misc{huang2024ffaa,
      title={FFAA: Multimodal Large Language Model based Explainable Open-World Face Forgery Analysis Assistant}, 
      author={Zhengchao Huang and Bin Xia and Zicheng Lin and Zhun Mou and Wenming Yang and Jiaya Jia},
      year={2024},
      eprint={2408.10072},
      archivePrefix={arXiv},
      primaryClass={cs.CV},
      url={https://arxiv.org/abs/2408.10072}, 
}

@misc{li2024fakebench,
      title={FakeBench: Probing Explainable Fake Image Detection via Large Multimodal Models}, 
      author={Yixuan Li and Xuelin Liu and Xiaoyang Wang and Bu Sung Lee and Shiqi Wang and Anderson Rocha and Weisi Lin},
      year={2024},
      eprint={2404.13306},
      archivePrefix={arXiv},
      primaryClass={cs.CV},
      url={https://arxiv.org/abs/2404.13306}, 
}

@misc{cao2025reveal,
      title={REVEAL: Reasoning-Enhanced Forensic Evidence Analysis for Explainable AI-Generated Image Detection}, 
      author={Huangsen Cao and Qin Mei and Zhiheng Li and Yuxi Li and Zhan Meng and Ying Zhang and Chen Li and Zhimeng Zhang and Xin Ding and Yongwei Wang and Jing Lyu and Fei Wu},
      year={2026},
      eprint={2511.23158},
      archivePrefix={arXiv},
      primaryClass={cs.CV},
      url={https://arxiv.org/abs/2511.23158}, 
}

@misc{ji2025fakexplained,
      title={Interpretable and Reliable Detection of AI-Generated Images via Grounded Reasoning in MLLMs}, 
      author={Yikun Ji and Hong Yan and Jun Lan and Huijia Zhu and Weiqiang Wang and Qi Fan and Liqing Zhang and Jianfu Zhang},
      year={2025},
      eprint={2506.07045},
      archivePrefix={arXiv},
      primaryClass={cs.CV},
      url={https://arxiv.org/abs/2506.07045}, 
}

@misc{ji2026trace,
      title={Locate-Then-Examine: Grounded Region Reasoning Improves Detection of AI-Generated Images}, 
      author={Yikun Ji and Yan Hong and Bowen Deng and Jun Lan and Huijia Zhu and Weiqiang Wang and Liqing Zhang and Jianfu Zhang},
      year={2026},
      eprint={2510.04225},
      archivePrefix={arXiv},
      primaryClass={cs.CV},
      url={https://arxiv.org/abs/2510.04225}, 
}

@misc{huang2025sida,
      title={SIDA: Social Media Image Deepfake Detection, Localization and Explanation with Large Multimodal Model}, 
      author={Zhenglin Huang and Jinwei Hu and Xiangtai Li and Yiwei He and Xingyu Zhao and Bei Peng and Baoyuan Wu and Xiaowei Huang and Guangliang Cheng},
      year={2025},
      eprint={2412.04292},
      archivePrefix={arXiv},
      primaryClass={cs.CV},
      url={https://arxiv.org/abs/2412.04292}, 
}

@misc{zhou2025aigiholmes,
      title={AIGI-Holmes: Towards Explainable and Generalizable AI-Generated Image Detection via Multimodal Large Language Models}, 
      author={Ziyin Zhou and Yunpeng Luo and Yuanchen Wu and Ke Sun and Jiayi Ji and Ke Yan and Shouhong Ding and Xiaoshuai Sun and Yunsheng Wu and Rongrong Ji},
      year={2025},
      eprint={2507.02664},
      archivePrefix={arXiv},
      primaryClass={cs.CV},
      url={https://arxiv.org/abs/2507.02664}, 
}

@misc{wen2025spot,
      title={Spot the Fake: Large Multimodal Model-Based Synthetic Image Detection with Artifact Explanation}, 
      author={Siwei Wen and Junyan Ye and Peilin Feng and Hengrui Kang and Zichen Wen and Yize Chen and Jiang Wu and Wenjun Wu and Conghui He and Weijia Li},
      year={2025},
      eprint={2503.14905},
      archivePrefix={arXiv},
      primaryClass={cs.CV},
      url={https://arxiv.org/abs/2503.14905}, 
}

@misc{xu2025fakeshield,
      title={FakeShield: Explainable Image Forgery Detection and Localization via Multi-modal Large Language Models}, 
      author={Zhipei Xu and Xuanyu Zhang and Runyi Li and Zecheng Tang and Qing Huang and Jian Zhang},
      year={2025},
      eprint={2410.02761},
      archivePrefix={arXiv},
      primaryClass={cs.CV},
      url={https://arxiv.org/abs/2410.02761}, 
}

@article{rong2024human,
   title={Towards Human-Centered Explainable AI: A Survey of User Studies for Model Explanations},
   volume={46},
   ISSN={1939-3539},
   url={http://dx.doi.org/10.1109/TPAMI.2023.3331846},
   DOI={10.1109/tpami.2023.3331846},
   number={4},
   journal={IEEE Transactions on Pattern Analysis and Machine Intelligence},
   publisher={Institute of Electrical and Electronics Engineers (IEEE)},
   author={Rong, Yao and Leemann, Tobias and Nguyen, Thai-Trang and Fiedler, Lisa and Qian, Peizhu and Unhelkar, Vaibhav and Seidel, Tina and Kasneci, Gjergji and Kasneci, Enkelejda},
   year={2024},
   month=Apr, pages={2104–2122}
}

@misc{liao2022human,
      title={Human-Centered Explainable AI (XAI): From Algorithms to User Experiences}, 
      author={Q. Vera Liao and Kush R. Varshney},
      year={2022},
      eprint={2110.10790},
      archivePrefix={arXiv},
      primaryClass={cs.AI},
      url={https://arxiv.org/abs/2110.10790}, 
}

@article{hertz2022prompttoprompt,
  title={Prompt-to-prompt image editing with cross attention control},
  author={Hertz, Amir and Mokady, Ron and Tenenbaum, Jay and Aberman, Kfir and Pritch, Yael and Cohen-Or, Daniel},
  journal={arXiv preprint arXiv:2208.01626},
  year={2022}
}

@inproceedings{brooks2023instructpix2pix,
  title={Instructpix2pix: Learning to follow image editing instructions},
  author={Brooks, Tim and Holynski, Aleksander and Efros, Alexei A},
  booktitle={Proceedings of the IEEE/CVF conference on computer vision and pattern recognition},
  pages={18392--18402},
  year={2023}
}

@misc{zhang2019bertscore,
      title={BERTScore: Evaluating Text Generation with BERT}, 
      author={Tianyi Zhang and Varsha Kishore and Felix Wu and Kilian Q. Weinberger and Yoav Artzi},
      year={2020},
      eprint={1904.09675},
      archivePrefix={arXiv},
      primaryClass={cs.CL},
      url={https://arxiv.org/abs/1904.09675}, 
}

@misc{cripwell2023sle,
      title={Simplicity Level Estimate (SLE): A Learned Reference-Less Metric for Sentence Simplification}, 
      author={Liam Cripwell and Joël Legrand and Claire Gardent},
      year={2023},
      eprint={2310.08170},
      archivePrefix={arXiv},
      primaryClass={cs.CL},
      url={https://arxiv.org/abs/2310.08170}, 
}

@misc{huang2025xtransfer,
      title={X-Transfer Attacks: Towards Super Transferable Adversarial Attacks on CLIP}, 
      author={Hanxun Huang and Sarah Erfani and Yige Li and Xingjun Ma and James Bailey},
      year={2025},
      eprint={2505.05528},
      archivePrefix={arXiv},
      primaryClass={cs.CV},
      url={https://arxiv.org/abs/2505.05528}, 
}

@misc{jia2025foa,
      title={Adversarial Attacks against Closed-Source MLLMs via Feature Optimal Alignment}, 
      author={Xiaojun Jia and Sensen Gao and Simeng Qin and Tianyu Pang and Chao Du and Yihao Huang and Xinfeng Li and Yiming Li and Bo Li and Yang Liu},
      year={2025},
      eprint={2505.21494},
      archivePrefix={arXiv},
      primaryClass={cs.CV},
      url={https://arxiv.org/abs/2505.21494}, 
}

@misc{openai2024gpt4omini,
  title={GPT-4o mini: Advancing Cost-Efficient Intelligence},
  author={{OpenAI}},
  year={2024},
  howpublished={OpenAI Blog},
  url={https://openai.com/index/gpt-4o-mini-advancing-cost-efficient-intelligence/}
}

@misc{openai2026gpt5mini,
  title={GPT-5 mini Model Documentation},
  author={{OpenAI}},
  year={2026},
  howpublished={OpenAI API Documentation},
  url={https://developers.openai.com/api/docs/models/gpt-5-mini}
}

@misc{openai2025gptimage15,
  title={GPT Image 1.5 Model Documentation},
  author={{OpenAI}},
  year={2025},
  howpublished={OpenAI API Documentation},
  url={https://developers.openai.com/api/docs/models/gpt-image-1.5}
}

@misc{openai2025gptoss,
  title={Introducing gpt-oss},
  author={{OpenAI}},
  year={2025},
  howpublished={OpenAI Blog},
  url={https://openai.com/index/introducing-gpt-oss/}
}

@misc{google2025gemini3flash,
  title={Gemini 3 Flash: Frontier Intelligence Built for Speed},
  author={{Google}},
  year={2025},
  howpublished={Google Blog},
  url={https://blog.google/products-and-platforms/products/gemini/gemini-3-flash/}
}

@misc{google2025gemini25pro,
  title={Gemini 2.5 Pro},
  author={{Google}},
  year={2025},
  howpublished={Google AI for Developers Documentation},
  url={https://ai.google.dev/gemini-api/docs/models/gemini-2.5-pro}
}

@misc{google2025nanobananapro,
  title={Introducing Nano Banana Pro},
  author={{Google DeepMind}},
  year={2025},
  howpublished={Google Blog},
  url={https://blog.google/innovation-and-ai/products/nano-banana-pro/}
}

@misc{google2026nanobanana2,
  title={Build with Nano Banana 2},
  author={{Google}},
  year={2026},
  howpublished={Google Blog},
  url={https://blog.google/innovation-and-ai/technology/developers-tools/build-with-nano-banana-2/}
}

@misc{zai2025glm46v,
  title={GLM-4.6V: Open Source Multimodal Models with Native Tool Use},
  author={{Z.ai}},
  year={2025},
  howpublished={Z.ai Developer Documentation},
  url={https://docs.z.ai/guides/vlm/glm-4.6v}
}

@misc{qwen2026qwen35,
  title  = {{Qwen3.5}: Towards Native Multimodal Agents},
  author = {{Qwen Team}},
  month  = {February},
  year   = {2026},
  url    = {https://qwen.ai/blog?id=qwen3.5}
}

@misc{deepseek2025v32,
      title={DeepSeek-V3.2: Pushing the Frontier of Open Large Language Models}, 
      author={DeepSeek-AI and Aixin Liu and Aoxue Mei and Bangcai Lin and Bing Xue and Bingxuan Wang and Bingzheng Xu and Bochao Wu and Bowei Zhang and Chaofan Lin and Chen Dong and Chengda Lu and Chenggang Zhao and Chengqi Deng and Chenhao Xu and Chong Ruan and Damai Dai and Daya Guo and Dejian Yang and Deli Chen and Erhang Li and Fangqi Zhou and Fangyun Lin and Fucong Dai and Guangbo Hao and Guanting Chen and Guowei Li and H. Zhang and Hanwei Xu and Hao Li and Haofen Liang and Haoran Wei and Haowei Zhang and Haowen Luo and Haozhe Ji and Honghui Ding and Hongxuan Tang and Huanqi Cao and Huazuo Gao and Hui Qu and Hui Zeng and Jialiang Huang and Jiashi Li and Jiaxin Xu and Jiewen Hu and Jingchang Chen and Jingting Xiang and Jingyang Yuan and Jingyuan Cheng and Jinhua Zhu and Jun Ran and Junguang Jiang and Junjie Qiu and Junlong Li and Junxiao Song and Kai Dong and Kaige Gao and Kang Guan and Kexin Huang and Kexing Zhou and Kezhao Huang and Kuai Yu and Lean Wang and Lecong Zhang and Lei Wang and Liang Zhao and Liangsheng Yin and Lihua Guo and Lingxiao Luo and Linwang Ma and Litong Wang and Liyue Zhang and M. S. Di and M. Y Xu and Mingchuan Zhang and Minghua Zhang and Minghui Tang and Mingxu Zhou and Panpan Huang and Peixin Cong and Peiyi Wang and Qiancheng Wang and Qihao Zhu and Qingyang Li and Qinyu Chen and Qiushi Du and Ruiling Xu and Ruiqi Ge and Ruisong Zhang and Ruizhe Pan and Runji Wang and Runqiu Yin and Runxin Xu and Ruomeng Shen and Ruoyu Zhang and S. H. Liu and Shanghao Lu and Shangyan Zhou and Shanhuang Chen and Shaofei Cai and Shaoyuan Chen and Shengding Hu and Shengyu Liu and Shiqiang Hu and Shirong Ma and Shiyu Wang and Shuiping Yu and Shunfeng Zhou and Shuting Pan and Songyang Zhou and Tao Ni and Tao Yun and Tian Pei and Tian Ye and Tianyuan Yue and Wangding Zeng and Wen Liu and Wenfeng Liang and Wenjie Pang and Wenjing Luo and Wenjun Gao and Wentao Zhang and Xi Gao and Xiangwen Wang and Xiao Bi and Xiaodong Liu and Xiaohan Wang and Xiaokang Chen and Xiaokang Zhang and Xiaotao Nie and Xin Cheng and Xin Liu and Xin Xie and Xingchao Liu and Xingkai Yu and Xingyou Li and Xinyu Yang and Xinyuan Li and Xu Chen and Xuecheng Su and Xuehai Pan and Xuheng Lin and Xuwei Fu and Y. Q. Wang and Yang Zhang and Yanhong Xu and Yanru Ma and Yao Li and Yao Li and Yao Zhao and Yaofeng Sun and Yaohui Wang and Yi Qian and Yi Yu and Yichao Zhang and Yifan Ding and Yifan Shi and Yiliang Xiong and Ying He and Ying Zhou and Yinmin Zhong and Yishi Piao and Yisong Wang and Yixiao Chen and Yixuan Tan and Yixuan Wei and Yiyang Ma and Yiyuan Liu and Yonglun Yang and Yongqiang Guo and Yongtong Wu and Yu Wu and Yuan Cheng and Yuan Ou and Yuanfan Xu and Yuduan Wang and Yue Gong and Yuhan Wu and Yuheng Zou and Yukun Li and Yunfan Xiong and Yuxiang Luo and Yuxiang You and Yuxuan Liu and Yuyang Zhou and Z. F. Wu and Z. Z. Ren and Zehua Zhao and Zehui Ren and Zhangli Sha and Zhe Fu and Zhean Xu and Zhenda Xie and Zhengyan Zhang and Zhewen Hao and Zhibin Gou and Zhicheng Ma and Zhigang Yan and Zhihong Shao and Zhixian Huang and Zhiyu Wu and Zhuoshu Li and Zhuping Zhang and Zian Xu and Zihao Wang and Zihui Gu and Zijia Zhu and Zilin Li and Zipeng Zhang and Ziwei Xie and Ziyi Gao and Zizheng Pan and Zongqing Yao and Bei Feng and Hui Li and J. L. Cai and Jiaqi Ni and Lei Xu and Meng Li and Ning Tian and R. J. Chen and R. L. Jin and S. S. Li and Shuang Zhou and Tianyu Sun and X. Q. Li and Xiangyue Jin and Xiaojin Shen and Xiaosha Chen and Xinnan Song and Xinyi Zhou and Y. X. Zhu and Yanping Huang and Yaohui Li and Yi Zheng and Yuchen Zhu and Yunxian Ma and Zhen Huang and Zhipeng Xu and Zhongyu Zhang and Dongjie Ji and Jian Liang and Jianzhong Guo and Jin Chen and Leyi Xia and Miaojun Wang and Mingming Li and Peng Zhang and Ruyi Chen and Shangmian Sun and Shaoqing Wu and Shengfeng Ye and T. Wang and W. L. Xiao and Wei An and Xianzu Wang and Xiaowen Sun and Xiaoxiang Wang and Ying Tang and Yukun Zha and Zekai Zhang and Zhe Ju and Zhen Zhang and Zihua Qu},
      year={2025},
      eprint={2512.02556},
      archivePrefix={arXiv},
      primaryClass={cs.CL},
      url={https://arxiv.org/abs/2512.02556}, 
}

@misc{blackforestlabs2026flux2dev,
  title={FLUX.2-dev},
  author={{Black Forest Labs}},
  year={2026},
  howpublished={Hugging Face Model Card},
  url={https://huggingface.co/black-forest-labs/FLUX.2-dev}
}

@misc{hunyuan2025hunyuanimage,
      title={HunyuanImage 3.0 Technical Report}, 
      author={Siyu Cao and Hangting Chen and Peng Chen and Yiji Cheng and Yutao Cui and Xinchi Deng and Ying Dong and Kipper Gong and Tianpeng Gu and Xiusen Gu and Tiankai Hang and Duojun Huang and Jie Jiang and Zhengkai Jiang and Weijie Kong and Changlin Li and Donghao Li and Junzhe Li and Xin Li and Yang Li and Zhenxi Li and Zhimin Li and Jiaxin Lin and Linus and Lucaz Liu and Shu Liu and Songtao Liu and Yu Liu and Yuhong Liu and Yanxin Long and Fanbin Lu and Qinglin Lu and Yuyang Peng and Yuanbo Peng and Xiangwei Shen and Yixuan Shi and Jiale Tao and Yangyu Tao and Qi Tian and Pengfei Wan and Chunyu Wang and Kai Wang and Lei Wang and Linqing Wang and Lucas Wang and Qixun Wang and Weiyan Wang and Hao Wen and Bing Wu and Jianbing Wu and Yue Wu and Senhao Xie and Fang Yang and Miles Yang and Xiaofeng Yang and Xuan Yang and Zhantao Yang and Jingmiao Yu and Zheng Yuan and Chao Zhang and Jian-Wei Zhang and Peizhen Zhang and Shi-Xue Zhang and Tao Zhang and Weigang Zhang and Yepeng Zhang and Yingfang Zhang and Zihao Zhang and Zijian Zhang and Penghao Zhao and Zhiyuan Zhao and Xuefei Zhe and Jianchen Zhu and Zhao Zhong},
      year={2026},
      eprint={2509.23951},
      archivePrefix={arXiv},
      primaryClass={cs.CV},
      url={https://arxiv.org/abs/2509.23951}, 
}

@misc{meituan2025longcatimage,
      title={LongCat-Image Technical Report}, 
      author={Meituan LongCat Team and Hanghang Ma and Haoxian Tan and Jiale Huang and Junqiang Wu and Jun-Yan He and Lishuai Gao and Songlin Xiao and Xiaoming Wei and Xiaoqi Ma and Xunliang Cai and Yayong Guan and Jie Hu},
      year={2025},
      eprint={2512.07584},
      archivePrefix={arXiv},
      primaryClass={cs.CV},
      url={https://arxiv.org/abs/2512.07584}, 
}

@misc{wu2025qwenimage,
      title={Qwen-Image Technical Report}, 
      author={Chenfei Wu and Jiahao Li and Jingren Zhou and Junyang Lin and Kaiyuan Gao and Kun Yan and Sheng-ming Yin and Shuai Bai and Xiao Xu and Yilei Chen and Yuxiang Chen and Zecheng Tang and Zekai Zhang and Zhengyi Wang and An Yang and Bowen Yu and Chen Cheng and Dayiheng Liu and Deqing Li and Hang Zhang and Hao Meng and Hu Wei and Jingyuan Ni and Kai Chen and Kuan Cao and Liang Peng and Lin Qu and Minggang Wu and Peng Wang and Shuting Yu and Tingkun Wen and Wensen Feng and Xiaoxiao Xu and Yi Wang and Yichang Zhang and Yongqiang Zhu and Yujia Wu and Yuxuan Cai and Zenan Liu},
      year={2025},
      eprint={2508.02324},
      archivePrefix={arXiv},
      primaryClass={cs.CV},
      url={https://arxiv.org/abs/2508.02324}, 
}

@misc{bytedance2026seedream45,
  title={Seedream 4.5},
  author={{ByteDance Seed Team}},
  year={2026},
  howpublished={Official Model Page},
  url={https://seed.bytedance.com/en/seedream4_5}
}

@misc{alibaba2026wan26,
  title={Wan2.6 Image Generation and Editing},
  author={{Alibaba Cloud}},
  year={2026},
  howpublished={Alibaba Cloud Model Studio Documentation},
  url={https://www.alibabacloud.com/help/en/model-studio/wan-image-generation-api-reference}
}

@misc{liu2023llava,
      title={Visual Instruction Tuning}, 
      author={Haotian Liu and Chunyuan Li and Qingyang Wu and Yong Jae Lee},
      year={2023},
      eprint={2304.08485},
      archivePrefix={arXiv},
      primaryClass={cs.CV},
      url={https://arxiv.org/abs/2304.08485}, 
}

@misc{li2024llavanext,
  title={LLaVA-NeXT: Improved Reasoning, OCR, and World Knowledge},
  author={{LLaVA Team}},
  year={2024},
  howpublished={Project Blog},
  url={https://llava-vl.github.io/blog/2024-01-30-llava-next/}
}

@misc{wen2025busterxpp,
      title={BusterX++: Towards Unified Cross-Modal AI-Generated Content Detection and Explanation with MLLM}, 
      author={Haiquan Wen and Tianxiao Li and Zhenglin Huang and Yiwei He and Guangliang Cheng},
      year={2026},
      eprint={2507.14632},
      archivePrefix={arXiv},
      primaryClass={cs.CV},
      url={https://arxiv.org/abs/2507.14632}, 
}

@inproceedings{gupta2025multifakeverse,
author = {Gupta, Parul and Ghosh, Shreya and Gedeon, Tom and Do, Thanh-Toan and Dhall, Abhinav},
title = {Multiverse Through Deepfakes: The MultiFakeVerse Dataset of Person-Centric Visual and Conceptual Manipulations},
year = {2025},
isbn = {9798400720352},
publisher = {Association for Computing Machinery},
address = {New York, NY, USA},
url = {https://doi.org/10.1145/3746027.3758283},
doi = {10.1145/3746027.3758283},
booktitle = {Proceedings of the 33rd ACM International Conference on Multimedia},
pages = {13258–13265},
numpages = {8},
location = {Dublin, Ireland},
series = {MM '25}
}

@inproceedings{narang2025laylens,
author = {Narang, Abhijeet and Gupta, Parul and Su, Liuyijia and Dhall, Abhinav},
title = {LayLens: Improving Deepfake Understanding through Simplified Explanations},
year = {2025},
isbn = {9798400714993},
publisher = {Association for Computing Machinery},
address = {New York, NY, USA},
url = {https://doi.org/10.1145/3716553.3757093},
doi = {10.1145/3716553.3757093},
booktitle = {Proceedings of the 27th International Conference on Multimodal Interaction},
pages = {788–790},
numpages = {3},
location = {
},
series = {ICMI '25}
}

@article{miller2019explanation,
title = {Explanation in artificial intelligence: Insights from the social sciences},
journal = {Artificial Intelligence},
volume = {267},
pages = {1-38},
year = {2019},
issn = {0004-3702},
doi = {https://doi.org/10.1016/j.artint.2018.07.007},
url = {https://www.sciencedirect.com/science/article/pii/S0004370218305988},
author = {Tim Miller}
}

@inproceedings{ribera2019better,
author = {Ribera Turró, Mireia and Lapedriza, Agata},
year = {2019},
month = {03},
pages = {},
title = {Can we do better explanations? A proposal of User-Centered Explainable AI}
}

@inproceedings{liao2020questioning,
author = {Liao, Q. Vera and Gruen, Daniel and Miller, Sarah},
title = {Questioning the AI: Informing Design Practices for Explainable AI User Experiences},
year = {2020},
isbn = {9781450367080},
publisher = {Association for Computing Machinery},
address = {New York, NY, USA},
url = {https://doi.org/10.1145/3313831.3376590},
doi = {10.1145/3313831.3376590},
booktitle = {Proceedings of the 2020 CHI Conference on Human Factors in Computing Systems},
pages = {1–15},
numpages = {15},
location = {Honolulu, HI, USA},
series = {CHI '20}
}

@misc{ehsan2020human,
      title={Human-centered Explainable AI: Towards a Reflective Sociotechnical Approach}, 
      author={Upol Ehsan and Mark O. Riedl},
      year={2020},
      eprint={2002.01092},
      archivePrefix={arXiv},
      primaryClass={cs.HC},
      url={https://arxiv.org/abs/2002.01092}, 
}

@article{linder2021level,
author = {Linder, Rhema and Mohseni, Sina and Yang, Fan and Pentyala, Shiva K. and Ragan, Eric D. and Hu, Xia Ben},
title = {How level of explanation detail affects human performance in interpretable intelligent systems: A study on explainable fact checking},
journal = {Applied AI Letters},
volume = {2},
number = {4},
pages = {e49},
doi = {https://doi.org/10.1002/ail2.49},
url = {https://onlinelibrary.wiley.com/doi/abs/10.1002/ail2.49},
eprint = {https://onlinelibrary.wiley.com/doi/pdf/10.1002/ail2.49},
year = {2021}
}

@misc{zhang2025icedit,
      title={In-Context Edit: Enabling Instructional Image Editing with In-Context Generation in Large Scale Diffusion Transformer}, 
      author={Zechuan Zhang and Ji Xie and Yu Lu and Zongxin Yang and Yi Yang},
      year={2025},
      eprint={2504.20690},
      archivePrefix={arXiv},
      primaryClass={cs.CV},
      url={https://arxiv.org/abs/2504.20690}, 
}

@misc{openai2025gptimage1,
  title        = {GPT Image 1 Model},
  author       = {{OpenAI}},
  year         = {2025},
  howpublished = {OpenAI API Documentation},
  url          = {https://developers.openai.com/api/docs/models/gpt-image-1},
  note         = {Accessed: 2026-05-06}
}

@misc{kampf2025gemini20flash,
  title        = {Experiment with Gemini 2.0 Flash native image generation},
  author       = {Google},
  year         = {2025},
  howpublished = {Google Developers Blog},
  url          = {https://developers.googleblog.com/experiment-with-gemini-20-flash-native-image-generation/},
  note         = {Accessed: 2026-05-06}
}

@misc{bai2025qwen3vl,
      title={Qwen3-VL Technical Report}, 
      author={Shuai Bai and Yuxuan Cai and Ruizhe Chen and Keqin Chen and Xionghui Chen and Zesen Cheng and Lianghao Deng and Wei Ding and Chang Gao and Chunjiang Ge and Wenbin Ge and Zhifang Guo and Qidong Huang and Jie Huang and Fei Huang and Binyuan Hui and Shutong Jiang and Zhaohai Li and Mingsheng Li and Mei Li and Kaixin Li and Zicheng Lin and Junyang Lin and Xuejing Liu and Jiawei Liu and Chenglong Liu and Yang Liu and Dayiheng Liu and Shixuan Liu and Dunjie Lu and Ruilin Luo and Chenxu Lv and Rui Men and Lingchen Meng and Xuancheng Ren and Xingzhang Ren and Sibo Song and Yuchong Sun and Jun Tang and Jianhong Tu and Jianqiang Wan and Peng Wang and Pengfei Wang and Qiuyue Wang and Yuxuan Wang and Tianbao Xie and Yiheng Xu and Haiyang Xu and Jin Xu and Zhibo Yang and Mingkun Yang and Jianxin Yang and An Yang and Bowen Yu and Fei Zhang and Hang Zhang and Xi Zhang and Bo Zheng and Humen Zhong and Jingren Zhou and Fan Zhou and Jing Zhou and Yuanzhi Zhu and Ke Zhu},
      year={2025},
      eprint={2511.21631},
      archivePrefix={arXiv},
      primaryClass={cs.CV},
      url={https://arxiv.org/abs/2511.21631}, 
}

@inproceedings{wang2022m2tr,
author = {Wang, Junke and Wu, Zuxuan and Ouyang, Wenhao and Han, Xintong and Chen, Jingjing and Jiang, Yu-Gang and Li, Ser-Nam},
title = {M2TR: Multi-modal Multi-scale Transformers for Deepfake Detection},
year = {2022},
isbn = {9781450392389},
publisher = {Association for Computing Machinery},
address = {New York, NY, USA},
url = {https://doi.org/10.1145/3512527.3531415},
doi = {10.1145/3512527.3531415},
booktitle = {Proceedings of the 2022 International Conference on Multimedia Retrieval},
pages = {615–623},
numpages = {9},
location = {Newark, NJ, USA},
series = {ICMR '22}
}

@misc{wang2025internvl35,
      title={InternVL3.5: Advancing Open-Source Multimodal Models in Versatility, Reasoning, and Efficiency}, 
      author={Weiyun Wang and Zhangwei Gao and Lixin Gu and Hengjun Pu and Long Cui and Xingguang Wei and Zhaoyang Liu and Linglin Jing and Shenglong Ye and Jie Shao and Zhaokai Wang and Zhe Chen and Hongjie Zhang and Ganlin Yang and Haomin Wang and Qi Wei and Jinhui Yin and Wenhao Li and Erfei Cui and Guanzhou Chen and Zichen Ding and Changyao Tian and Zhenyu Wu and Jingjing Xie and Zehao Li and Bowen Yang and Yuchen Duan and Xuehui Wang and Zhi Hou and Haoran Hao and Tianyi Zhang and Songze Li and Xiangyu Zhao and Haodong Duan and Nianchen Deng and Bin Fu and Yinan He and Yi Wang and Conghui He and Botian Shi and Junjun He and Yingtong Xiong and Han Lv and Lijun Wu and Wenqi Shao and Kaipeng Zhang and Huipeng Deng and Biqing Qi and Jiaye Ge and Qipeng Guo and Wenwei Zhang and Songyang Zhang and Maosong Cao and Junyao Lin and Kexian Tang and Jianfei Gao and Haian Huang and Yuzhe Gu and Chengqi Lyu and Huanze Tang and Rui Wang and Haijun Lv and Wanli Ouyang and Limin Wang and Min Dou and Xizhou Zhu and Tong Lu and Dahua Lin and Jifeng Dai and Weijie Su and Bowen Zhou and Kai Chen and Yu Qiao and Wenhai Wang and Gen Luo},
      year={2025},
      eprint={2508.18265},
      archivePrefix={arXiv},
      primaryClass={cs.CV},
      url={https://arxiv.org/abs/2508.18265}, 
}
